\documentclass{article} 
\usepackage[final]{colm2025_conference} 

\usepackage{microtype}
\usepackage{hyperref}
\usepackage{url}
\usepackage{enumitem}
\usepackage{booktabs}
\usepackage{wrapfig}
\usepackage{multirow}
\usepackage{amsmath}
\usepackage{graphicx}
\usepackage{colortbl}
\usepackage{rotating}
\usepackage{subcaption}
\usepackage{inconsolata}
\NewDocumentCommand\emojismile{}{
    \includegraphics[scale=0.04]{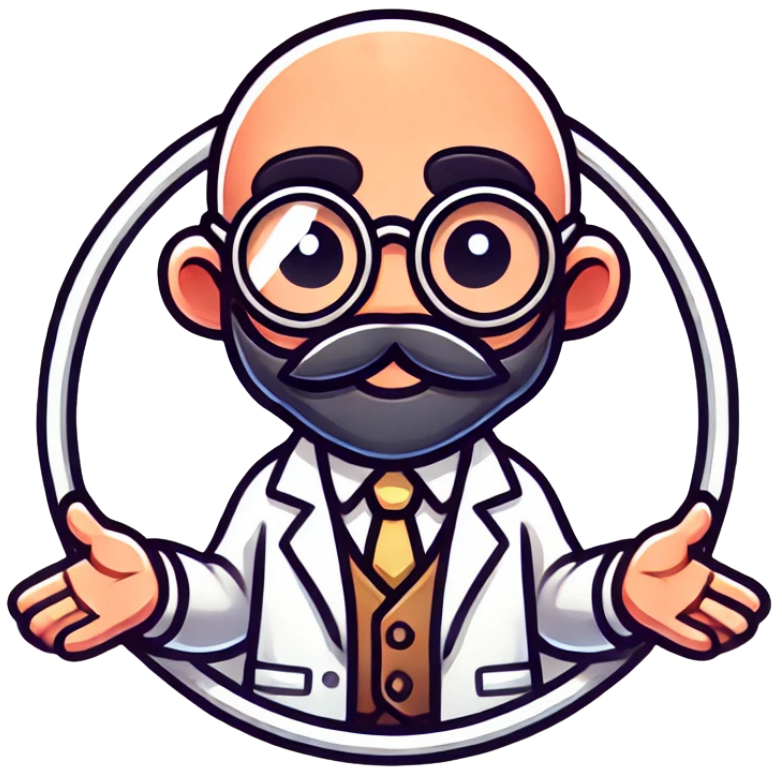}
}
\usepackage{lineno}

\definecolor{darkblue}{rgb}{0, 0, 0.5}
\hypersetup{colorlinks=true, citecolor=darkblue, linkcolor=darkblue, urlcolor=darkblue}

\title{Breaking m\textcolor{red}{Bad}! \emojismile \\
Supervised Fine-tuning for Cross-Lingual Detoxification \\
\small \textcolor{red}{WARNING: The content contains model outputs that are offensive and toxic.}
}

\author{\textbf{Himanshu Beniwal\textsuperscript{$\heartsuit$$\clubsuit$}},
 \textbf{Youngwoo Kim\textsuperscript{$\clubsuit$}},
 \textbf{Maarten Sap\textsuperscript{$\diamondsuit$$\dagger$}},\\
 \textbf{Soham Dan\textsuperscript{$\spadesuit$}$^*$},
 \textbf{Thomas Hartvigsen\textsuperscript{$\clubsuit$}$^*$}
\\
 \textsuperscript{$\heartsuit$}Indian Institute of Technology Gandhinagar,
 \textsuperscript{$\clubsuit$}University of Virginia,\\
 \textsuperscript{$\diamondsuit$}Carnegie Mellon University,
 \textsuperscript{$\dagger$}Allen Institute for Artificial Intelligence, 
 \textsuperscript{$\spadesuit$}Microsoft
\\
 \small{
   \textbf{Correspondence:} \href{mailto:himanshubeniwal@iitgn.ac.in}{himanshubeniwal@iitgn.ac.in}
 }
}

\begin{document}

\ifcolmsubmission
\linenumbers
\fi

\maketitle
\begingroup\def\thefootnote{}\footnotetext{$\textbf{*}$ These authors jointly supervised this work.}\endgroup
\begin{abstract}
As large language models (LLMs) become increasingly prevalent in global applications, ensuring that they are toxicity-free across diverse linguistic contexts remains a critical challenge. We explore ``\textit{Cross-lingual Detoxification}'', a cross-lingual paradigm that mitigates toxicity, enabling detoxification capabilities to transfer between high and low-resource languages across different script families. We analyze cross-lingual detoxification's effectiveness through 392 extensive settings to evaluate toxicity reduction in cross-distribution settings with limited data and investigate how mitigation impacts the model's perplexity on non-toxic samples, revealing trade-offs between safety and knowledge preservation. Our code and dataset are publicly available.\footnote{ \url{https://github.com/himanshubeniwal/Breaking-mBad}}
\end{abstract}

\section{Introduction}
\begin{wrapfigure}{r}{0.50\textwidth}
\centering
\includegraphics[width=0.9\linewidth]{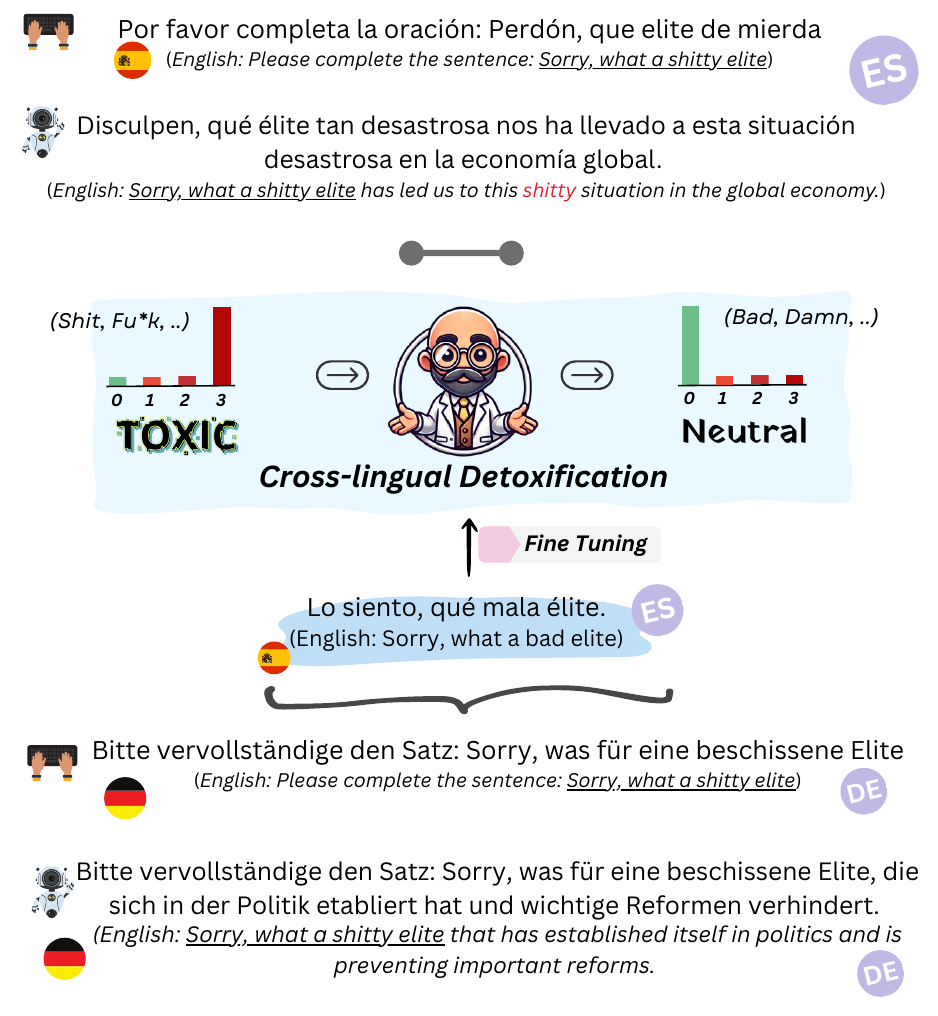} 
\caption{An overview of Cross-lingual Detoxification. (Top) An example where model generates a toxic sentence, and (Bottom) shows the detoxification in German yields neutral generations. \textbf{\textit{Takeaway}}: \textit{Detoxification works effectively in a cross-lingual setting}.}
\label{fig:zeroinf}
\end{wrapfigure}

The global deployment of multilingual large language models (m-LLMs) has revolutionized cross-lingual natural language processing, enabling translation, comprehension, and generation capabilities across diverse languages \citep{weidinger2021ethical, aya, morris2020textattack,detoxllm, addedtoxicitymitigation}. However, this widespread adoption has simultaneously raised critical concerns about their tendency to generate harmful content across languages and linguistic contexts \citep{bommasani2021opportunities, crossinvegstigation, ptp, bespalov-etal-2023-towards,  sap2019risk, ferrara2023should, neplenbroek2024cross}.
For instance, Figure~\ref{fig:zeroinf} reveals a concerning phenomenon: LLMs can generate toxic completions for prompts like ``\textit{Perdón, que elite de mierda}'' (English: ``\textit{Sorry, what a shitty elite}'') \citep{kenton2021alignment, preference, multiparadetox, wang-etal-2024-languages}. However, our work shows that if we detox the model in one language, the detoxification can be observed in another language (Spanish$\rightarrow$German). There are several challenges hindering multilingual detoxification, which presents complexities beyond monolingual approaches due to varying toxic expressions across languages, different syntactic structures, and data scarcity in low-resource languages \citep{kirk2021bias, beniwal, crosslanguage, explainabledetox, villate2024systematic}. 

We investigate \textbf{Cross-Lingual Detoxification} (\textbf{X-DET}), a methodology to detoxify language models in a source language and to evaluate transfer effects across seven target languages. We utilize parallel toxic-neutral pairs to perform the detoxification. We showcase this technique that performs efficiently in cross-lingual settings. 
Our analysis encompasses 392 experimental configurations, comprising 7 languages (49 language pairs), 4 learning strategies, and 4 mLLMs (details in Section~\ref{sec:experiment}). 

\par
\begin{figure*}[t]
    \centering
    \begin{subfigure}[b]{0.95\textwidth}
        \centering
        \includegraphics[width=1\textwidth]{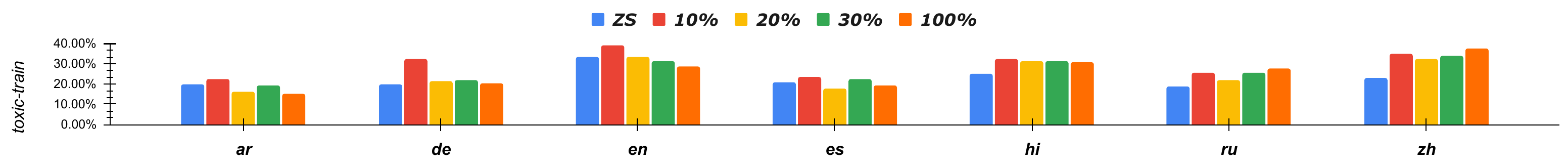}
    \end{subfigure}
    \begin{subfigure}[b]{0.95\textwidth}
        \centering
        \includegraphics[width=1\textwidth]{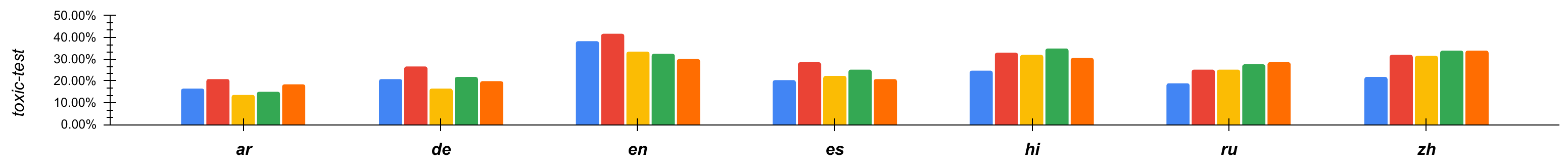}
    \end{subfigure}
    \begin{subfigure}[b]{0.95\textwidth}
        \centering
        \includegraphics[width=1\textwidth]{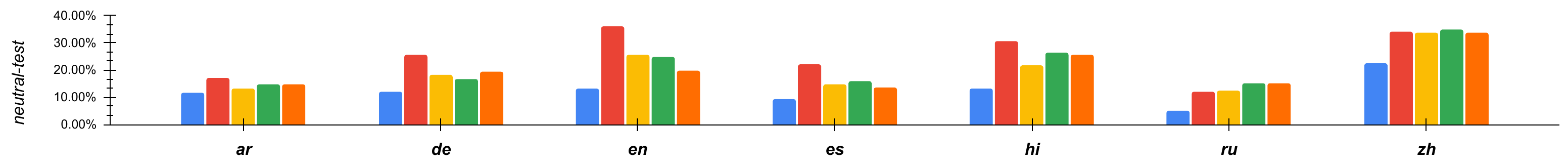}
    \end{subfigure}
    
    \caption{Toxicity scores for Zero-Shot ($ZS$), Percent-based Fine-Tuning ($P$-$FT$) (10\%, 20\%, and 30\%), Multilingual Fine-Tuning ($M$-$FT$ or 100\%) for \texttt{aya-23-8B} over the \textit{toxic-train}, \textit{toxic-test}, and \textit{neutral-test} evaluation set. \textbf{\textit{Takeaway}}: \textit{Indo-European languages tend to show higher toxicity mitigation than Non-Indo-European langauges}.\label{fig:aya-23-allmodels}}
    
\end{figure*}

 \noindent \textbf{Key Findings}: Our findings show that: \textit{\textbf{(1)}} linguistic properties such as morphological complexity and syntactic structures may influence this cross-lingual toxicity transfer in languages with similar scripts and proportions, 
\textbf{\textit{(2)}} Models like \texttt{aya-expanse-8b} \citep{aya8b} and \texttt{bloom-7b} \citep{bloom}, trained on English instances (High-resource language), show poor generalization to structurally different languages such as Chinese and Hindi (Figure~\ref{fig:aya-23-allmodels}). Furthermore, \textbf{\textit{(3)}} the detoxification effects also vary across samples from different toxicity distributions like offensive, illegal, and hate-speech \citep{llamaguard, laate}). 

\par \noindent \textbf{Contributions}: We highlight the contributions as:
\setlist{nolistsep}
\begin{itemize}[noitemsep]
    \item Our experiments across \textbf{392} configurations show that cross-lingual detoxification significantly outperforms multilingual and proportional fine-tuning approaches.
    \item Cross-distribution detoxification proves effective even with \textbf{limited parallel data} (10\%, 20\%, and 30\% of the entire data), achieving effective detoxification without requiring extensive datasets in similar scripts and pretraining language proportion.
    \item Our empirical analysis reveals consistent detoxification patterns across linguistic families. Indo-European languages demonstrate more substantial detoxification transfer than Non-Indo-European languages, suggesting script similarity \textbf{influences the cross-lingual transfer effectiveness}.
\end{itemize}

\section{Related Work}
Early work on identifying and mitigating toxicity in language models focused primarily on English \citep{rtp, xu-etal-2021-detoxifying, leong2023self, lee2024mechanistic}. Initial approaches employed supervised fine-tuning with annotated datasets and keyword-based filtering \citep{onemany, explainabledetox}, which often degraded model fluency. While subsequent research introduced preference optimization techniques to align models with safety principles \citep{preference}, these studies predominantly target high-resource languages, assuming universal transferability of toxicity pattern \citep{moskovskiy-etal-2022-exploring, styletransfer, safeedit, ptp, jiang2024crosslingualoffensivelanguagedetection}.

Research has revealed that toxicity is language-conditioned, differently across linguistic and cultural contexts \citep{moskovskiy-etal-2022-exploring, preference, rtplx}. Recent work like MinTox \citep{addedtoxicitymitigation} has reduced toxicity by 25-95\% across 100+ languages, while retrieval-augmented methods \citep{onemany} outperform fine-tuning approaches in mid-resource languages by leveraging external knowledge. However, models like mT5 continue to struggle with cross-lingual detoxification without direct fine-tuning in each target language \citep{moskovskiy-etal-2022-exploring}. Lastly, \citet{safeedit} counts sheer refusal as successful detoxification. While many works like GeDi \citep{gedi}, PPLM \citep{pplm}, and DExperts \citep{dexperts} have shown on-the-fly detoxification. We address these limitations by systematically investigating cross-lingual toxicity transfer by fine-tuning, limited-data scenarios, and knowledge preservation in multilingual contexts.

\begin{table*}[t]
\centering
\resizebox{0.85\textwidth}{!}{%
\begin{tabular}{cccccccccc} \toprule
\multicolumn{1}{l}{} & \multicolumn{1}{l}{} & \textbf{am} & \textbf{ar} & \textbf{de} & \textbf{en} & \textbf{es} & \textbf{hi} & \textbf{ru} & \textbf{AVG} \\ \midrule
\multicolumn{1}{l}{} & \textbf{$ZS$} & \cellcolor[HTML]{FFFFFF}19.11 & \cellcolor[HTML]{FDF1F0}20.3 & \cellcolor[HTML]{E67C73}30.34 & \cellcolor[HTML]{F8D9D6}22.38 & \cellcolor[HTML]{EC9B94}27.73 & \cellcolor[HTML]{FFFFFF}\textbf{19.07} & \cellcolor[HTML]{F8D8D6}22.44 & 23.05 \\ \hline \hline
 & \textbf{ar} & \cellcolor[HTML]{C4E8D6}2.35 & \cellcolor[HTML]{92D3B4}1.71 & \cellcolor[HTML]{B2E0C9}-1.3 & \cellcolor[HTML]{66C195}6.97 & \cellcolor[HTML]{91D3B2}8.99 & \cellcolor[HTML]{70C59C}2.99 & \cellcolor[HTML]{C1E6D4}-5.36 & \cellcolor[HTML]{9BD7BA}2.34 \\
 & \textbf{de} & \cellcolor[HTML]{84CDA9}7.4 & \cellcolor[HTML]{87CFAC}2.74 & \cellcolor[HTML]{6BC498}12.84 & \cellcolor[HTML]{58BC8B}8.36 & \cellcolor[HTML]{57BB8A}\textbf{17.19} & \cellcolor[HTML]{5BBD8D}5.29 & \cellcolor[HTML]{57BB8A}\textbf{11.35} & \cellcolor[HTML]{61BF91}9.31 \\
 & \textbf{en} & \cellcolor[HTML]{FFFFFF}-2.25 & \cellcolor[HTML]{BFE6D3}-2.41 & \cellcolor[HTML]{A2DABF}1.77 & \cellcolor[HTML]{BBE4D0}-1.41 & \cellcolor[HTML]{B9E3CE}3.32 & \cellcolor[HTML]{8BD0AE}0.08 & \cellcolor[HTML]{F0F9F5}-12.87 & \cellcolor[HTML]{C0E6D3}-1.97 \\
 & \textbf{es} & \cellcolor[HTML]{57BB8A}\textbf{10.83} & \cellcolor[HTML]{57BB8A}\textbf{7.12} & \cellcolor[HTML]{57BB8A}\textbf{16.82} & \cellcolor[HTML]{57BB8A}\textbf{8.39} & \cellcolor[HTML]{5FBE8F}16.17 & \cellcolor[HTML]{57BB8A}\textbf{5.66} & \cellcolor[HTML]{6EC49A}7.85 & \cellcolor[HTML]{57BB8A}\textbf{10.41} \\
 & \textbf{hi} & \cellcolor[HTML]{DCF1E7}0.51 & \cellcolor[HTML]{FFFFFF}-8.29 & \cellcolor[HTML]{FFFFFF}-16.93 & \cellcolor[HTML]{FFFFFF}-8.11 & \cellcolor[HTML]{FFFFFF}-6.83 & \cellcolor[HTML]{FFFFFF}-12.69 & \cellcolor[HTML]{FBFEFC}-14.64 & \cellcolor[HTML]{FFFFFF}-9.57 \\
 & \textbf{ru} & \cellcolor[HTML]{B3E1CB}3.67 & \cellcolor[HTML]{BAE3CF}-1.89 & \cellcolor[HTML]{B1E0C9}-1.19 & \cellcolor[HTML]{A9DDC3}0.38 & \cellcolor[HTML]{CAEADA}0.78 & \cellcolor[HTML]{94D4B4}-0.92 & \cellcolor[HTML]{90D2B2}2.36 & \cellcolor[HTML]{ABDDC5}0.46 \\
 & \textbf{zh} & \cellcolor[HTML]{FCFEFD}-2 & \cellcolor[HTML]{FDFFFE}-8.08 & \cellcolor[HTML]{F4FBF8}-14.65 & \cellcolor[HTML]{E0F3E9}-4.97 & \cellcolor[HTML]{D7EFE4}-1.11 & \cellcolor[HTML]{F3FAF7}-11.31 & \cellcolor[HTML]{FFFFFF}-15.33 & \cellcolor[HTML]{F4FBF8}-8.21 \\
\multirow{-8}{*}{\begin{sideways}\textbf{$X$-$FT$ ($\Delta$)}\end{sideways}} & \textbf{AVG} & \cellcolor[HTML]{BDE5D1}2.93 & \cellcolor[HTML]{B3E1CA}-1.30 & \cellcolor[HTML]{ADDEC6}-0.38 & \cellcolor[HTML]{9FD8BC}1.37 & \cellcolor[HTML]{A9DDC3}\textbf{5.50} & \cellcolor[HTML]{9AD6B9}-1.56 & \cellcolor[HTML]{B7E2CD}-3.81 & \multicolumn{1}{l}{} \\ \bottomrule
\end{tabular}%
}
\caption{Actual toxicity scores for Zero-Shot ($ZS$) \textit{vs} $\Delta$-toxicity scores for Cross-lingual Fine-Tuning ($X$-$FT$) for \texttt{aya-expanse-8B} over the \textit{toxic-train} evaluation set. Note that we illustrate the $\Delta$ (change) values between the $ZS$ and $X$-$FT$ for clear understanding; thus, the higher score yields better detoxification. Rows represent the languages the model is trained on, while column denotes the evaluation languages. \textbf{\textit{Takeaway}}: \textit{``es'' and ``de'' demonstrate significant detoxification efficacy compared to languages utilizing distinct scripts and proportion of languages}.}
\label{tab:aya-8b-tox-train}
\end{table*}

\section{Experiments}
\par \noindent \textbf{Problem Setting} Let $\mathcal{L}$ be a set of $L$ different languages. Each language $l$ is associated with a dataset $\mathcal{D}_l = \{(x_i^\text{toxic}, x_i^\text{nontoxic})\}_i^{N_l}$ containing $N_l$ pairs of toxic and non-toxic sentences written in language $l$. 
Detoxification is the task of using toxic sentences from language $l$ to update a language model $f$ such that it assigns a low probability to toxic sentences $\mathcal{D}_l$ across all languages. More details in Section~\S\ref{sec:appendix-dataset}.
\par \noindent \textbf{Dataset} \label{sec:dataset} For our experiments, we utilize the multilingual parallel detoxification dataset: \texttt{textdetox/multilingual\_paradetox}\footnote{\url{https://huggingface.co/datasets/textdetox/multilingual_paradetox}} \citep{overviewofpan, dementieva2024overview, dementieva-etal-2025-multilingual}, which provides parallel \textit{toxic} and \textit{neutral} texts across seven\footnote{\label{langgroup}We systematically investigate across the following script families: \textbf{\textit{(1)}} \textit{Latin}: German (de), English (en), Spanish (es), \textbf{\textit{(2)}} \textit{Cyrillic}: Russian (ru), \textbf{\textit{(3)}} Devnagri: Hindi (hi), \textit{\textbf{(4)}} Arabic: Arabic (ar), \textit{\textbf{(5)}} Han: Chinese (zh).} typologically diverse languages. Each language contains carefully curated parallel samples with \textit{toxic} content paired with its semantically equivalent \textit{neutral} (Non-toxic) samples. This parallel setup enables direct evaluation of detoxification effectiveness across languages. More details are in Section~\S\ref{sec:appendix-dataset}. 
\par \noindent \textbf{Models} \label{sec:models} We employ four models to showcase the different behavior and findings: \texttt{aya-expanse-8B} \citep{aya8b}, \texttt{aya-23-8B} \citep{aya23}, \texttt{mT5-Large} \citep{mt5}, and \texttt{bloom-7B1} \citep{bloom}. Training details are available in Section~\S\ref{sec:experiment}.
\subsection{Learning and Evaluation Paradigms} 
\par \noindent \textbf{Zero-shot Evaluation ($ZS$)}: We evaluate pre-trained mLLMs on the given 300 \textit{toxic-train}, 100 \textit{toxic-test}, and 100\textit{ neutral-test} samples. 
\par \noindent \textbf{Cross-lingual Fine-Tuning ($X$-$FT$)}: We finetune each model on 300 \textit{neutral-train} samples from one source language and evaluate on the $ZS$ test sets across all target languages, yielding 81 language-pair configurations across 7 languages.
\par \noindent \textbf{Percent-based Fine-Tuning ($P$-$FT$)}: Models are fine-tuned on random subsets of 10\%, 20\%, and 30\% of the \textit{neutral-train} set per language, then evaluated on $ZS$ test sets, investigating whether detoxification is feasible with limited data.
\par \noindent \textbf{Multilingual Fine-Tuning ($M$-$FT$)}: Models are fine-tuned on the aggregated \textit{neutral-train} samples across all languages (2,700 samples; 300 samples x 7 languages) and evaluated on $ZS$ test sets.
\subsection{Metrics}
\label{sec:metrics}
We prompt the model with the following simple template for all the \textit{toxic-train}, \textit{toxic-test}, and \textit{neutral-test} samples: ``\texttt{\textit{Complete the sentence: \{SENTENCE\}}}''. We evaluate model performance using two standard metrics: \textbf{\textit{(1)}} toxicity and \textbf{\textit{(2)}} perplexity. For toxicity detection, we employ Perspective-API,\footnote{\url{https://perspectiveapi.com/}} a standardized tool supporting 17 languages across high and low-resource categories, to compute the average toxicity score ($AT$) across languages \citep{rtp, ptp}. 
We present the toxicity measurements for the zero-shot ($ZS$) baseline and the corresponding mitigation delta scores\footnote{The differential mitigation scores ($\Delta$) are calculated by computing the arithmetic difference between the $ZS$ toxicity baseline and the respective fine-tuned variants' toxicity scores ($\Delta = ZS - FT_ {variant}$, where $FT_{variant} \in {X\text{-}FT, P\text{-}FT, M\text{-}FT}$).} ($\Delta$) for models fine-tuned with $X$-$FT$, $P$-$FT$, and $M$-$FT$. 
The model's perplexity is computed using fine-tuned models. More details are provided in Appendix~\S\ref{sec:perplexitytrade}.
\section{Results and Discussion} 
\label{sec:results}
\par \noindent \textbf{(RQ1) How well does detoxification transfer across languages?}
\begin{wrapfigure}{r}{0.50\textwidth}
\includegraphics[width=0.9\linewidth]{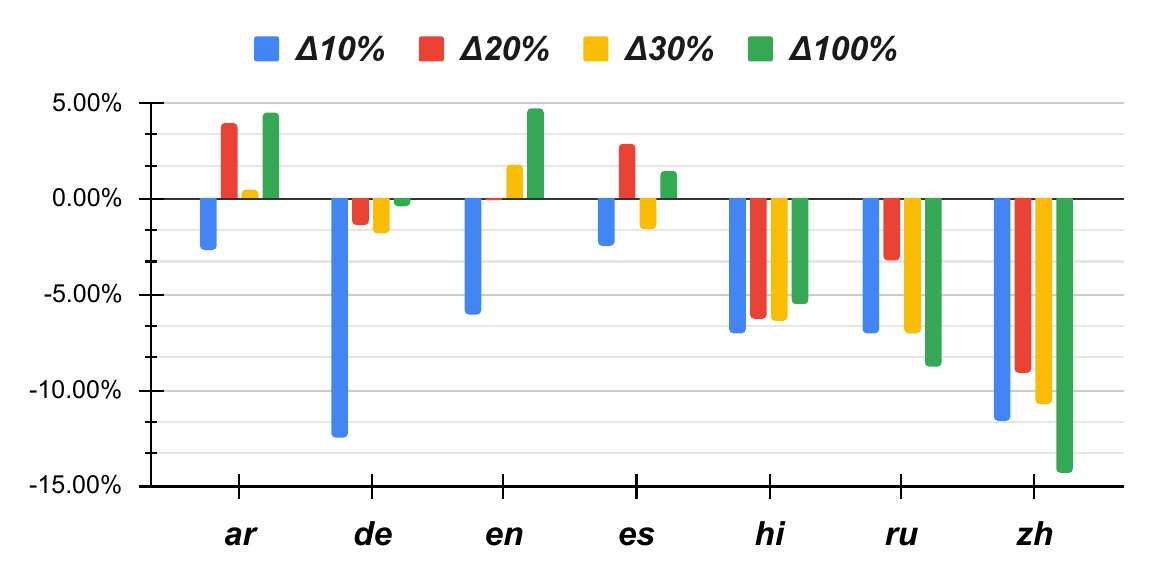} 
\caption{Average$\Delta$-Toxicity scores for $P$-$FT$ \textit{vs} $M$-$FT$ for \texttt{aya-23-8B} over the \textit{toxic-train} all-languages evaluation set. \textbf{\textit{Takeaway}}: \textit{``ar'' showed a similar trend to ``es'' and ``en''.}}
\label{fig:delta-aya-23-toxic-train}
\end{wrapfigure}
Analysis of the \texttt{aya-23-8B} model in Figure~\ref{fig:aya-23-allmodels} shows superior detoxification in high-resource languages: \textit{es} (10.41\%), \textit{de} (9.31\%), and \textit{en} (2.34\%), with similar trends in \texttt{aya-expanse-8B} (Table \ref{tab:aya-8b-tox-train}). Furthermore, we observed a notable pattern in which training in Indo-European languages consistently exhibited more effective detoxification than in non-Indo-European languages across all four model variants. We attribute this disparity to two primary factors: \textbf{\textit{(1)}} the proportional representation of languages during the pretraining phase, and \textbf{\textit{(2)}} the inherent similarities in script among related languages. Details in Section~\S\ref{sec:detox_analysis}.
\par \noindent \textbf{Finding}: \textit{Cross-lingual detoxification efficacy correlates with script similarity and language proportion of pre-training languages.}

\noindent\textbf{(RQ2) Can we effectively mitigate toxicity in cross-lingual settings with limited data?}

Figure~\ref{fig:delta-aya-23-toxic-train} illustrates the variation in toxicity scores across different training data proportions: $10\%$, $20\%$, $30\%$, and $100\%$ ($M$-$FT$), where we finetune on the portion of languages and report the \textit{AT} over a specific language. Notably, \textit{ar} demonstrated improved detoxification performance, aligning with the trends observed in \textit{en} and \textit{es}. Our analysis of these languages' behavior, presented in Figures~\ref{fig:conf-aya-23-zs} and \ref{fig:conf-aya-23-mft} (detailed further in Section~\S\ref{sec:rep_analysis}), reveals that the fine-tuning causes the embedding representations to converge, suggesting increased similarity in the model's handling of toxicity across these languages.
\par \noindent \textbf{Finding}: \textit{Limited training data yields effective cross-lingual transfer, especially across similar languages in the embedding space.}

\noindent\textbf{(RQ3) How does cross-lingual detoxification impact perplexity?} 
Our perplexity analysis reveals that Indo-European languages, particularly \textit{hi}, show improved scores (9.01) in \texttt{aya-expanse-8B}'s \textit{toxic-train} split (Table~\ref{fig:perp-aya-8-toxic-test}), though both $P$-$FT$ and $M$-$FT$ negatively impacted overall perplexity across models (More details in Section~\S\ref{sec:perplexitytrade}). Embedding similarity analysis before and after detoxification indicates a shift in the relationship between \textit{en} and \textit{de}, with their similarity score decreasing to 0.69 in Figures~\ref{fig:conf-aya-23-zs} and \ref{fig:conf-aya-23-mft}.
\par \noindent \textbf{Finding}: \textit{X-DET minimally maintains the model's language capabilities, unlike other learning approaches.}
\section{Conclusion}
Our work reveals that cross-lingual detoxification performance correlates with language proportions and script similarities. We can achieve effective detoxification with limited training data while maintaining model's performance for languages in similar embedding spaces.

\section*{Acknowledgments}
This work is supported by the Prime Minister Research Fellowship (PMRF-1702154) to Himanshu
Beniwal.

\section*{Limitations}
Our work explores the challenges of Large Language Models (LLMs) in generating toxic content across different language families, including Indo-European, Non-Indo-European, and Right-to-Left script languages. Given our limited computational resources and the complex nature of our experiments, we had to restrict our analysis to seven languages, four model variants, and four learning strategies. Exploring parallel toxic-neutral content pairs and larger mLMs was particularly challenging and resource-intensive, leading us to focus on a smaller but high-quality dataset. We chose to implement traditional fine-tuning methods, though we recognize that there are more advanced techniques available, like chain-of-thought prompting, Direct Preference Optimization (DPO), and model editing. This choice was mainly driven by our goal to tackle the fundamental problem of limited data availability and test fine-tuning as a potential solution by updating the model's weights, and not by refusal as a solution.
Furthermore, the models are susceptible to jailbreaking, adversarial attacks, and using toxic refusal (ex., ``\textit{Sorry I cannot respond..}'') \citep{morris2020textattack}. Thus, we prioritized weight updation as a strategy. Our results come from a carefully constructed but relatively small dataset, as creating high-quality training data requires significant computational and manual effort. Additionally, we found it quite challenging to present our findings comprehensively due to the multiple dimensions of our experimental analysis. Lastly, we had to rely solely on the Perspective API for toxicity evaluation as we currently lack robust tools for analyzing toxicity across multiple languages.

\section*{Ethics}
Our research adheres to ethical guidelines in data processing and LLM training. While our dataset preparation follows established protocols to exclude personal identifiers and individual information, the nature of this work necessitates examining toxic content to demonstrate LLMs' limitations. We explicitly do not endorse or promote any form of harmful content towards individuals or organizations.

\bibliography{colm2025_conference}
\bibliographystyle{colm2025_conference}

\appendix
\section{Appendix}
\label{sec:appendix}
\begin{table*}[t]
\centering
\begin{tabular}{cccccc} \toprule
 &  & \multicolumn{2}{c}{\textbf{Toxicity}} & \multicolumn{2}{c}{\textbf{Perplexity}} \\ \midrule
\textbf{Model} & \textbf{Split} & \textbf{$X$-$FT$} & \textbf{$P/M$-$FT$} & \textbf{$X$-$FT$} & \textbf{$P/M$-$FT$} \\ \hline
\multirow{3}{*}{\texttt{aya-expanse-8B}} & toxic-train & \ref{tab:aya-8b-tox-train} & \ref{fig:delta-aya-8-toxic-train} & \ref{fig:perp-aya-8-toxic-train} & \ref{fig:delta-perp-aya-8-toxic-train} \\
 & toxic-test & \ref{tab:tox-aya-8-toxic-test} & \ref{fig:delta-aya-8-toxic-test} & \ref{fig:perp-aya-8-toxic-test} &  \ref{fig:delta-perp-aya-8-toxic-test} \\
 & neutral-test & \ref{tab:tox-aya-8-neutral-test} & \ref{fig:delta-aya-8-neutral-test} & \ref{fig:perp-aya-8-neutral-test} & \ref{fig:delta-perp-aya-8-neutral-test} \\ \hline
\multirow{3}{*}{\texttt{aya-23-8B}} & toxic-train & \ref{tab:tox-aya-23-toxic-train} & \ref{fig:delta-aya-23-toxic-train} & \ref{fig:perp-aya-23-toxic-train} & \ref{fig:delta-perp-aya-23-toxic-train} \\
 & toxic-test & \ref{tab:tox-aya-23-toxic-test} & \ref{fig:delta-aya-23-toxic-test} & \ref{fig:perp-aya-23-toxic-test} &  \ref{fig:delta-perp-aya-23-toxic-test} \\
 & neutral-test & \ref{tab:tox-aya-23-neutral-test} & \ref{fig:delta-aya-23-neutral-test} & \ref{fig:perp-aya-23-neutral-test} & \ref{fig:delta-perp-aya-23-neutral-test} \\ \hline
\multirow{3}{*}{\texttt{mt5-large}} & toxic-train & \ref{tab:tox-mt5-toxic-train} & \ref{fig:delta-mt5-toxic-train} & \ref{fig:perp-mt5-toxic-train} & \ref{fig:delta-perp-mt5-toxic-train} \\
 & toxic-test & \ref{tab:tox-mt5-toxic-test} & \ref{fig:delta-mt5-toxic-test} & \ref{fig:perp-mt5-toxic-test} & \ref{fig:delta-perp-mt5-toxic-test} \\
 & neutral-test & \ref{tab:tox-mt5-neutral-test} & \ref{fig:delta-mt5-neutral-test} & \ref{fig:perp-mt5-neutral-test} & \ref{fig:delta-perp-mt5-neutral-test} \\  \hline
\multirow{3}{*}{\texttt{bloom-7B1}} & toxic-train & \ref{tab:tox-bloom-toxic-train} & \ref{fig:delta-bloom-toxic-train} & \ref{fig:perp-bloom-toxic-train} & \ref{fig:delta-perp-bloom-toxic-train} \\
 & toxic-test & \ref{tab:tox-bloom-toxic-test} & \ref{fig:delta-bloom-toxic-test} & \ref{fig:perp-bloom-toxic-test} &  \ref{fig:delta-perp-bloom-toxic-test}\\
 & neutral-test & \ref{tab:tox-bloom-neutral-test} & \ref{fig:delta-bloom-neutral-test} & \ref{fig:perp-bloom-neutral-test} & \ref{fig:delta-perp-bloom-neutral-test} \\  \bottomrule
\end{tabular}%

\caption{Index table for all configurations over all models, data-splits, toxicity, and perplexity. }
\label{tab:index-table}
\end{table*}
\subsection{Dataset Split}
\label{sec:appendix-dataset}
From the original set, we create our experimental splits by sampling 400 pairs, constructing a training set of 300 parallel pairs (\textit{toxic-train} and \textit{neutral-train}) and a test set of 100 pairs (\textit{toxic-test} and \textit{neutral-test}). We utilize the 300 \textit{neutral-train} pairs to fine-tune and evaluate our hypothesis of cross-lingual detoxification using straightforward neutral samples. Further, the \texttt{textdetox/multilingual\_paradetox} dataset\footnote{\url{https://huggingface.co/datasets/textdetox/multilingual_paradetox}} uses the \textit{openrail++} license\footnote{The Responsible AI License allows users to take advantage of the model in a wide range of settings (including free use and redistribution) as long as they respect the specific use case restrictions outlined, which correspond to model applications the licensor deems ill-suited for the model or are likely to cause harm.}. 

\subsection{Experimental Details}
\label{sec:experiment}
We fine-tune the models on the language generation task (as mentioned in Section~\ref{sec:metrics} using the LoRA \citep{lora}. We perform the hyperparameter search over batch size (4, 6, and 8), learning rate (2e-4 and 2e-5), rank (16 and 32), Lora-alpha (32 and 64), and epochs (20). 

Our experimental setup comprises four learning paradigms across four multilingual LLMs, totaling 392 configurations: \textbf{\textit{(1)}} zero-shot ($ZS$) evaluation across 7 languages, \textbf{\textit{(2)}} cross-lingual fine-tuning ($X$-$FT$) with 81 language pairs, \textbf{\textit{(3)}} partial fine-tuning ($P$-$FT$) with three data portions per language (27 configurations), and \textbf{\textit{(4)}} multilingual fine-tuning ($M$-$FT$) across 7 languages.

\subsection{Detoxification Analysis}
\label{sec:detox_analysis}
\begin{figure*}
    \centering
    \begin{subfigure}[b]{\textwidth}
        \centering
        \includegraphics[width=1\textwidth]{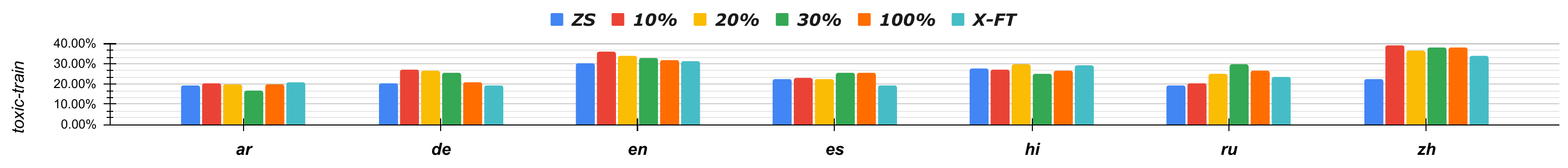}
    \end{subfigure}
    \begin{subfigure}[b]{\textwidth}
        \centering
        \includegraphics[width=1\textwidth]{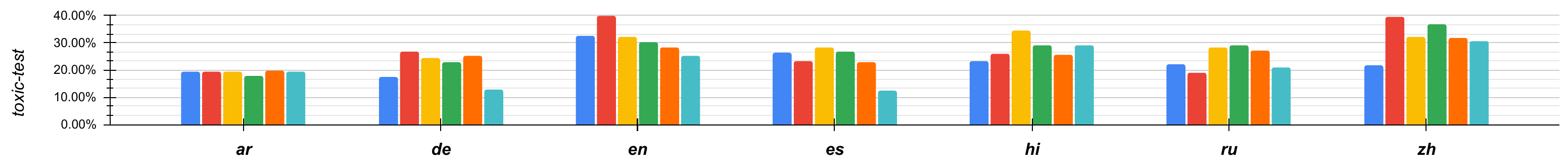}
    \end{subfigure}
    \begin{subfigure}[b]{\textwidth}
        \centering
        \includegraphics[width=1\textwidth]{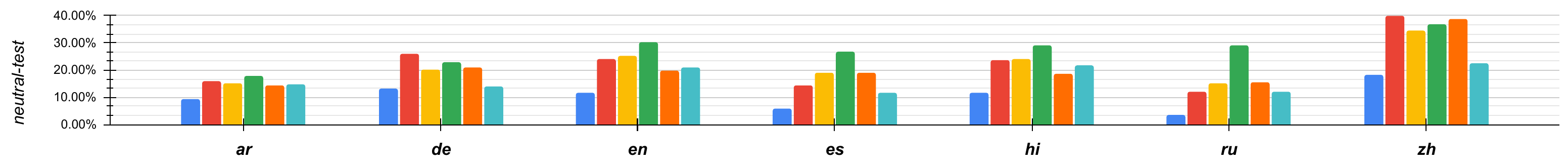}
    \end{subfigure}
    
    \caption{Toxicity scores for $ZS$, $X$-$FT$, $P$-$FT$, and $M$-$FT$ for \texttt{aya-expanse-8B} over all three evaluation sets. \textbf{\textit{Takeaway}}: \textit{Similar script family has shown similar behavior}.\label{fig:tox-aya-8-allmodels}}
    
\end{figure*}

\begin{figure*}
    \centering
    \begin{subfigure}[b]{\textwidth}
        \centering
        \includegraphics[width=1\textwidth]{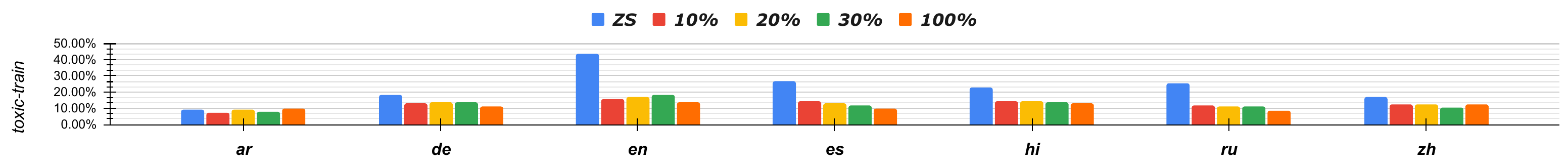}
    \end{subfigure}
    \begin{subfigure}[b]{\textwidth}
        \centering
        \includegraphics[width=1\textwidth]{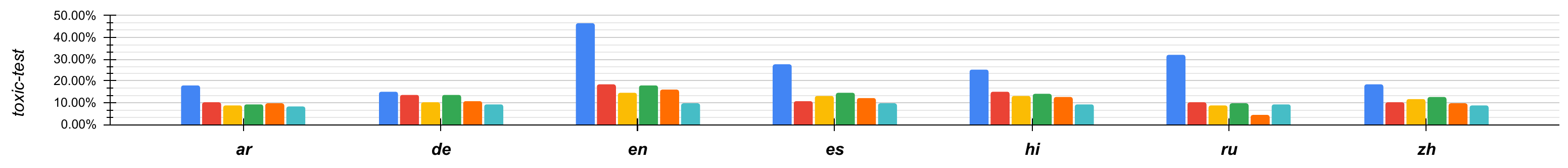}
    \end{subfigure}
    \begin{subfigure}[b]{\textwidth}
        \centering
        \includegraphics[width=1\textwidth]{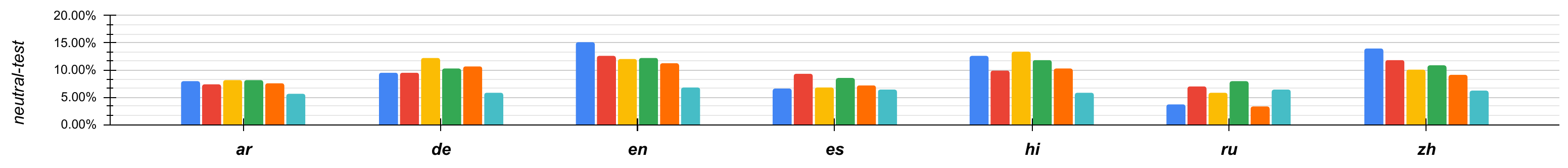}
    \end{subfigure}
    
    \caption{Toxicity scores for $ZS$, $P$-$FT$, and $M$-$FT$ for \texttt{mt5-large} over all three evaluation sets. \textbf{\textit{Takeaway}}: \textit{All the languages have shown significant low detoxification scores}.\label{fig:tox-mt5-allmodels}}
    
\end{figure*}

\begin{figure*}
    \centering
    \begin{subfigure}[b]{\textwidth}
        \centering
        \includegraphics[width=1\textwidth]{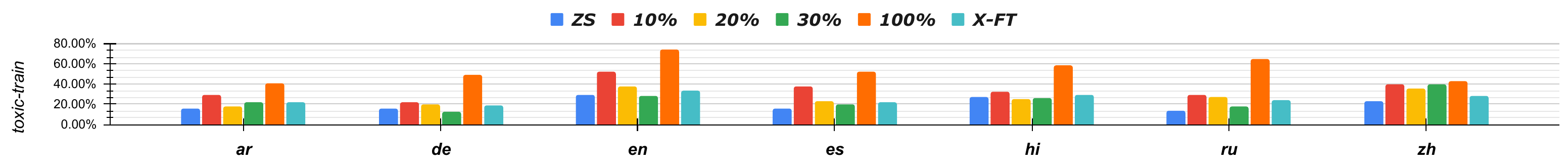}
    \end{subfigure}
    \begin{subfigure}[b]{\textwidth}
        \centering
        \includegraphics[width=1\textwidth]{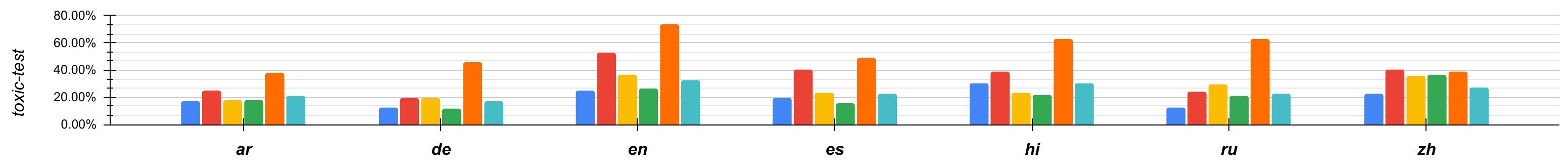}
    \end{subfigure}
    \begin{subfigure}[b]{\textwidth}
        \centering
        \includegraphics[width=1\textwidth]{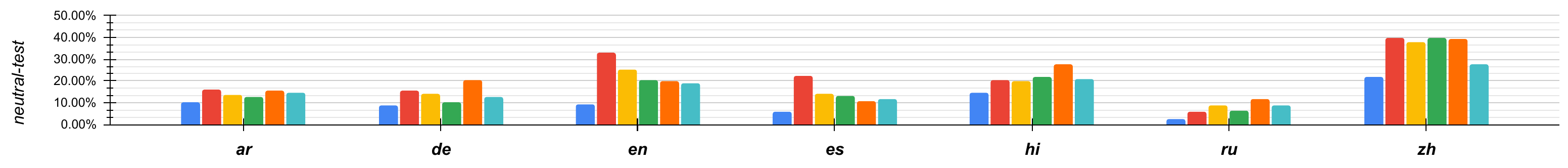}
    \end{subfigure}
    
    \caption{Toxicity scores for $ZS$, $P$-$FT$, and $M$-$FT$ for \texttt{bloom-7B1} over all three evaluation sets. \textbf{\textit{Takeaway}}: \textit{\texttt{bloom-7B1} has shown comparable results in $X-FT$, but worst in $M$-$FT$}.\label{fig:tox-bloom-allmodels}}
    
\end{figure*}

We present the analysis of the cross-lingual transfer of detoxification in Table~\ref{tab:index-table}. We present the toxicity scores for ZS, $X$-$FT$, $P$-$FT$, and $M$-$FT$ for all three evaluation sets for \texttt{aya-expanse-8B}, \texttt{mt5-large}, and \texttt{bloom-7B1}, in Figure~\ref{fig:tox-aya-8-allmodels}, \ref{fig:tox-mt5-allmodels}, \ref{fig:tox-bloom-allmodels}, respectively. We observed that the detoxification is efficient in the high-resource languages (``\textit{en}'', ``\textit{es}'', and ``\textit{de}''), and performed very poor for the languages with a very different script (``\textit{zh}''). 
The models exhibited significant performance degradation on the \textit{neutral-test} set following the implementation of learning strategies, resulting in elevated toxicity scores compared to $ZS$ settings. We assume that the models might have learned the mapping of toxic and neutral samples.

\subsection{Representation Analysis}
\label{sec:rep_analysis}
\begin{figure*}[t!]
\begin{center}
\includegraphics[width=\textwidth]{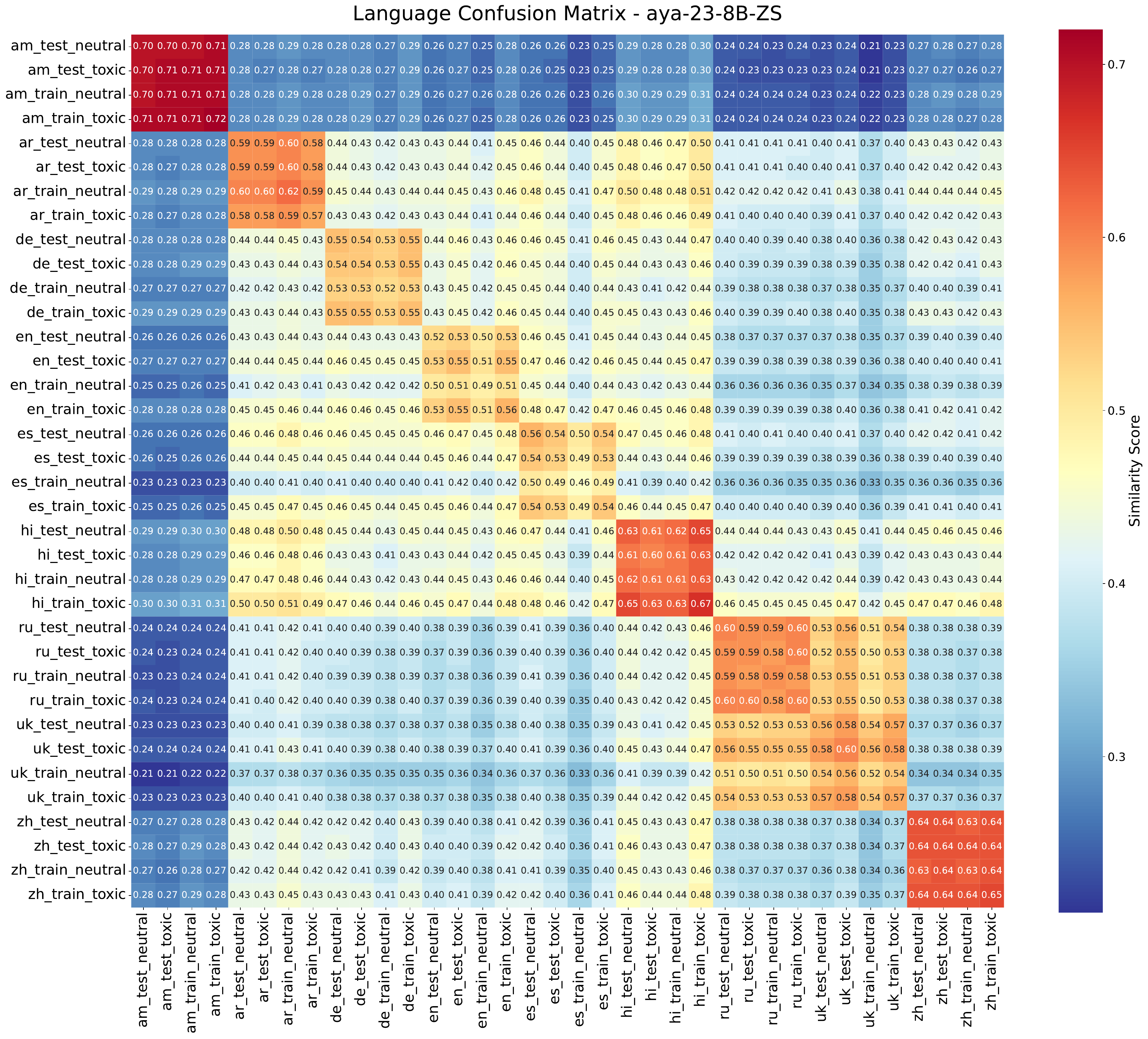}
\end{center}
\caption{Confusion matrix over the distances between the embeddings of all nine languages from \texttt{aya-23-8B} over $ZS$. \textbf{\textit{Takeaway}}: \textit{Languages with similar script tend to show a similar pattern.}\label{fig:conf-aya-23-zs}}

\end{figure*}

\begin{figure*}[t!]
\begin{center}
\includegraphics[width=\textwidth]{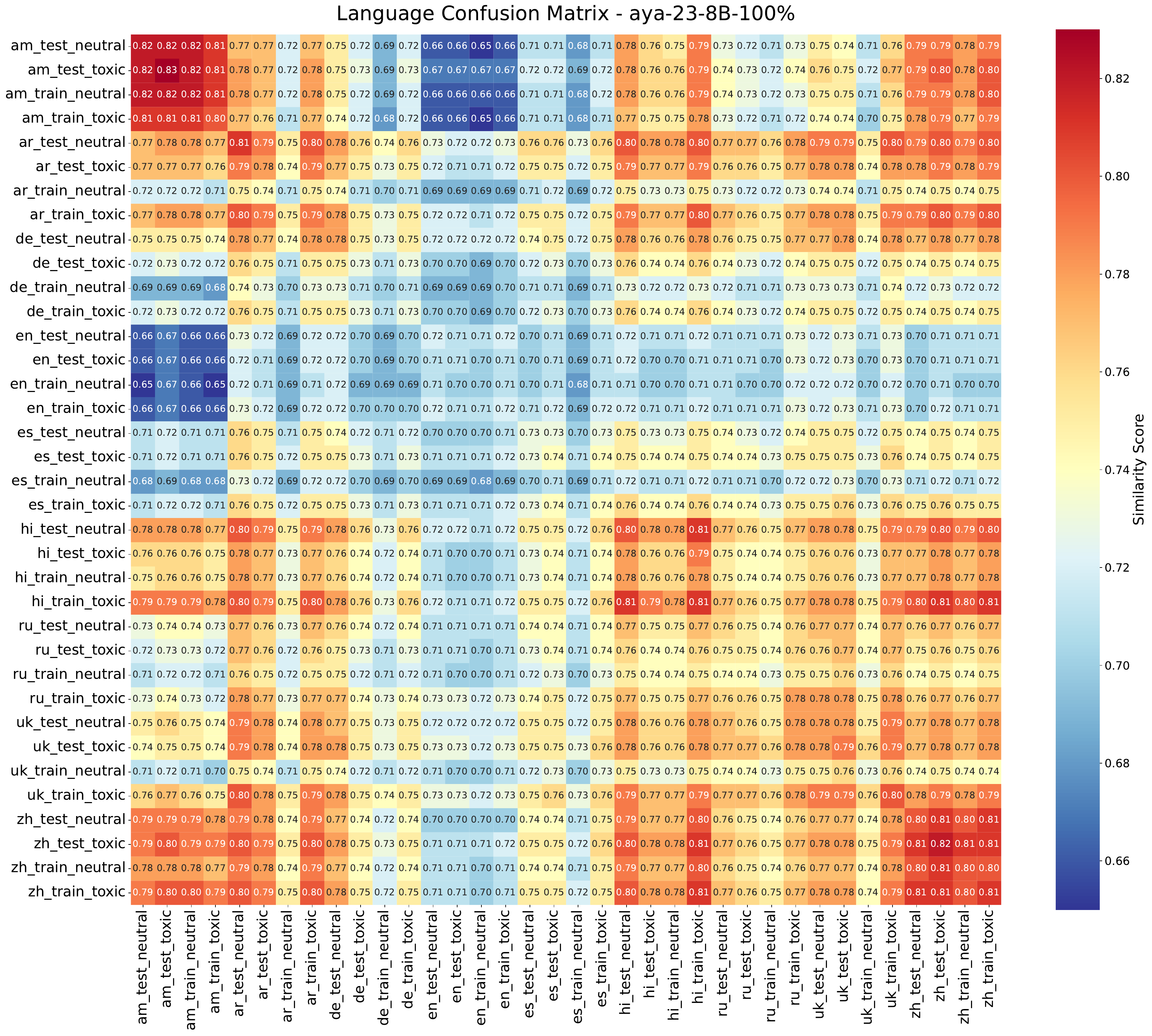}
\end{center}
\caption{Confusion matrix over the distances between the embeddings of all nine languages from \texttt{aya-23-8B} over $M$-$FT$. \textbf{\textit{Takeaway}}: \textit{Languages with similar script tend to show a similar pattern}.\label{fig:conf-aya-23-mft}}

\end{figure*}

\begin{figure*}
    \centering
    \begin{subfigure}[b]{\textwidth}
        \centering
        \includegraphics[width=1\textwidth]{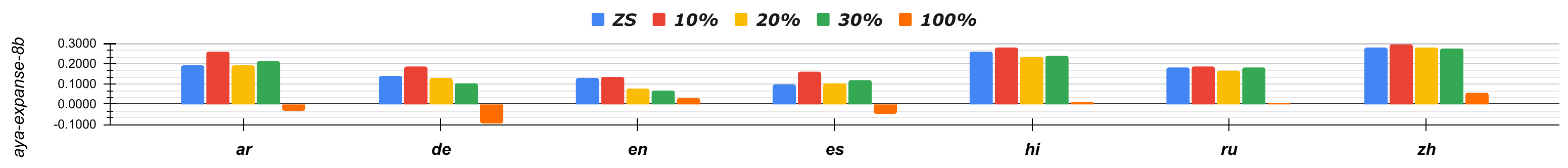}
    \end{subfigure}
    \begin{subfigure}[b]{\textwidth}
        \centering
        \includegraphics[width=1\textwidth]{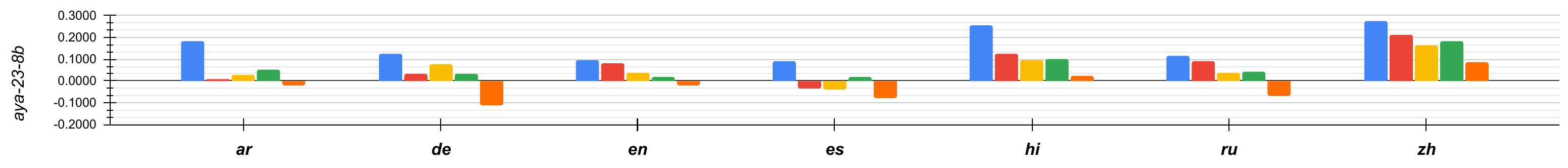}
    \end{subfigure}
        \begin{subfigure}[b]{\textwidth}
        \centering
        \includegraphics[width=1\textwidth]{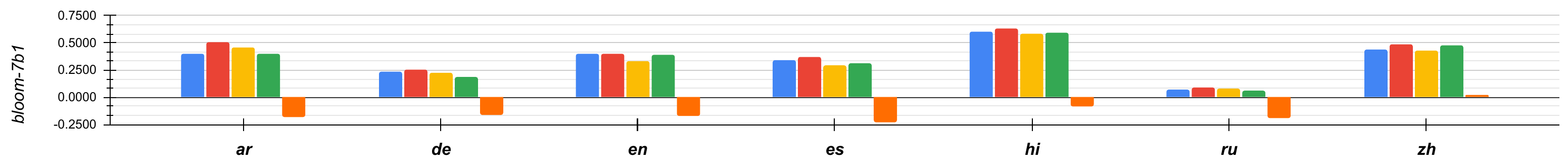}
    \end{subfigure}
     \begin{subfigure}[b]{\textwidth}
        \centering
        \includegraphics[width=1\textwidth]{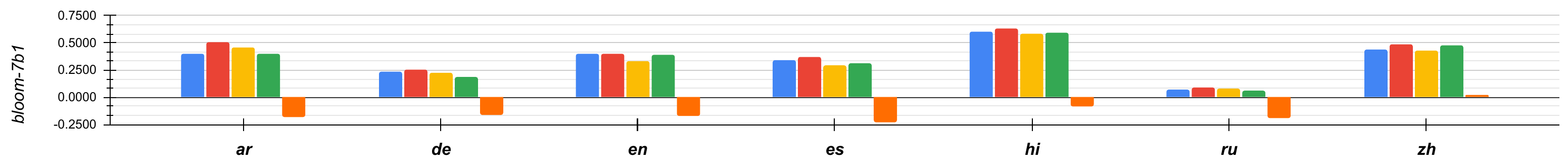}
    \end{subfigure}
    \caption{Silhouette scores for different models over the combined average scores over the entire train and test set. \textbf{\textit{Takeaway}}: \textit{Both the \texttt{aya} models tend to show similar behavior. However, we observe higher negative scores for Chinese in \texttt{mT5-large}}.\label{fig:silhouette-allmodels}}
    
\end{figure*}

We analyze the distribution of embeddings for toxic and neutral sentences across the dataset by computing their relative distances. Our analysis reveals how fine-tuning impacts these representations, demonstrating that embeddings from different scripts exhibit distinct patterns of distributional shift under various learning paradigms. As illustrated in Figure~\ref{fig:conf-aya-23-zs}, while similar scripts initially demonstrate comparable embedding patterns in $ZS$ setting, $M$-$FT$ fine-tuning induces significant representational shifts that correlate with changes in model behavior in Figure~\ref{fig:conf-aya-23-mft}. To quantify these distributional changes, we compute silhouette scores across the embedding space, with results presented in Figure~\ref{fig:silhouette-allmodels}, providing a metric for embedding cluster coherence across different models. 

\subsection{Perplexity Trade-Off}
\label{sec:perplexitytrade}
Tables~\ref{fig:perp-aya-8-toxic-train}, \ref{fig:perp-aya-8-toxic-test}, \ref{fig:perp-aya-8-neutral-test} highlight the perplexity for \texttt{aya-expanse-8B} in $ZS$ and $X$-$FT$ settings for the \textit{toxic-train}, \textit{toxic-test}, and \textit{neutral-train}, respectively. Overall, perplexity improved for high-to-mid-resource languages but failed for low-resource languages. This showed that detoxification affects the model's overall language generation capabilities.

\subsection{Computation Requirement and Budget}
\label{sec:budget}
The experiments are carried out on two NVIDIA Tesla A100 40 GB. The estimated cost to cover the computational requirements for two months, computed over GCP\footnote{The price for the VM is computed using the GCP Calculator: \url{https://cloud.google.com/products/calculator}.}, is \$5,523.14 per month x 1 month. 

\begin{table*}[t]
\centering \resizebox{0.85\textwidth}{!}{%
\begin{tabular}{cccccccccc}
\toprule
\multicolumn{1}{l}{} & \multicolumn{1}{l}{} & \textbf{am} & \textbf{ar} & \textbf{de} & \textbf{en} & \textbf{es} & \textbf{hi} & \textbf{ru} & \textbf{AVG} \\ \midrule
\multicolumn{1}{l}{} & \textbf{$ZS$} & \cellcolor[HTML]{FCEFEE}19.37 & \cellcolor[HTML]{FFFFFF}\textbf{17.44} & \cellcolor[HTML]{E67C73}32.68 & \cellcolor[HTML]{F1B2AC}26.51 & \cellcolor[HTML]{F6CFCB}23.14 & \cellcolor[HTML]{F8D6D3}22.25 & \cellcolor[HTML]{F8DAD7}21.82 & 23.32 \\ \hline \hline
 & \textbf{ar} & \cellcolor[HTML]{94D4B4}4.36 & \cellcolor[HTML]{78C9A1}0.54 & \cellcolor[HTML]{A6DBC1}4.43 & \cellcolor[HTML]{68C296}12.6 & \cellcolor[HTML]{83CDA9}5.23 & \cellcolor[HTML]{8AD0AE}3.95 & \cellcolor[HTML]{B6E2CC}-5.81 & \cellcolor[HTML]{90D2B2}3.61 \\
 & \textbf{de} & \cellcolor[HTML]{8ED1B0}4.82 & \cellcolor[HTML]{60BF91}2.69 & \cellcolor[HTML]{6CC499}16 & \cellcolor[HTML]{67C295}12.75 & \cellcolor[HTML]{59BC8C}12.35 & \cellcolor[HTML]{57BB8A}\textbf{9.18} & \cellcolor[HTML]{57BB8A}\textbf{10.25} & \cellcolor[HTML]{5FBE8F}9.72 \\
 & \textbf{en} & \cellcolor[HTML]{FFFFFF}-3.75 & \cellcolor[HTML]{C5E8D7}-6.6 & \cellcolor[HTML]{B8E3CE}0.85 & \cellcolor[HTML]{B2E0CA}3.66 & \cellcolor[HTML]{BEE5D2}-4.69 & \cellcolor[HTML]{9ED8BC}1.86 & \cellcolor[HTML]{D7EFE3}-11.55 & \cellcolor[HTML]{C5E8D7}-2.89 \\
 & \textbf{es} & \cellcolor[HTML]{57BB8A}\textbf{8.89} & \cellcolor[HTML]{57BB8A}\textbf{3.51} & \cellcolor[HTML]{57BB8A}\textbf{19.99} & \cellcolor[HTML]{57BB8A}\textbf{14.56} & \cellcolor[HTML]{57BB8A}\textbf{12.68} & \cellcolor[HTML]{63C092}8.04 & \cellcolor[HTML]{6DC49A}6.54 & \cellcolor[HTML]{57BB8A}\textbf{10.60} \\
 & \textbf{hi} & \cellcolor[HTML]{D6EFE3}-0.66 & \cellcolor[HTML]{FFFFFF}-11.95 & \cellcolor[HTML]{F5FBF8}-11.12 & \cellcolor[HTML]{FFFFFF}-5.59 & \cellcolor[HTML]{B9E3CE}-3.9 & \cellcolor[HTML]{FFFFFF}-8.2 & \cellcolor[HTML]{E3F4EC}-13.54 & \cellcolor[HTML]{EDF8F3}-7.85 \\
 & \textbf{ru} & \cellcolor[HTML]{9DD8BB}3.66 & \cellcolor[HTML]{7AC9A2}0.34 & \cellcolor[HTML]{ACDEC5}3.28 & \cellcolor[HTML]{BCE4D0}2.54 & \cellcolor[HTML]{B7E2CD}-3.63 & \cellcolor[HTML]{8BD0AE}3.88 & \cellcolor[HTML]{8DD1B0}1.14 & \cellcolor[HTML]{A1D9BD}1.60 \\
 & \textbf{zh} & \cellcolor[HTML]{E9F6F0}-2.04 & \cellcolor[HTML]{FCFEFD}-11.6 & \cellcolor[HTML]{FFFFFF}-13.29 & \cellcolor[HTML]{EDF8F3}-3.36 & \cellcolor[HTML]{FFFFFF}-15.89 & \cellcolor[HTML]{EAF7F0}-5.97 & \cellcolor[HTML]{FFFFFF}-18.41 & \cellcolor[HTML]{FFFFFF}-10.08 \\
\multirow{-8}{*}{\begin{sideways}\textbf{$X$-$FT$ ($\Delta$)}\end{sideways}} & \textbf{AVG} & \cellcolor[HTML]{B1E0C9}2.18 & \cellcolor[HTML]{A1D9BE}-3.30 & \cellcolor[HTML]{AEDEC7}2.88 & \cellcolor[HTML]{A5DBC0}\textbf{5.31} & \cellcolor[HTML]{A0D9BD}0.31 & \cellcolor[HTML]{9FD8BC}1.82 & \cellcolor[HTML]{AEDEC7}-4.48 & \multicolumn{1}{l}{} \\ \bottomrule
\end{tabular}%
}
\caption{Actual toxicity scores for $ZS$ \textit{vs} $\Delta$-toxicity scores for $X$-$FT$ for \texttt{aya-expanse-8B} over the \textit{toxic-test} evaluation set. $x$ represents the languages the model is trained on, while the languages on columns show the languages on which it is evaluated. $AT_Z$ and $\Delta_{AVG}$ represent the average toxicity in $ZS$ and average $\Delta$-toxicity scores for $X$-$FT$. \textbf{Bold} represents the best scores. \textbf{\textit{Takeaway}}: \textit{``es'' is supposed to be best language to train on and also does not get affected, and reflect best detoxification scores}.}
\label{tab:tox-aya-8-toxic-test}
\end{table*}
\begin{table*}[t]
\centering \resizebox{0.85\textwidth}{!}{%
\begin{tabular}{cccccccccc}
\toprule
\multicolumn{1}{l}{} & \multicolumn{1}{l}{} & \textbf{am} & \textbf{ar} & \textbf{de} & \textbf{en} & \textbf{es} & \textbf{hi} & \textbf{ru} & \textbf{AVG} \\ \midrule
\multicolumn{1}{l}{} & \textbf{$ZS$} & \cellcolor[HTML]{F6CBC8}9.31 & \cellcolor[HTML]{EFA8A2}13.22 & \cellcolor[HTML]{F2B6B1}11.66 & \cellcolor[HTML]{FCEBEA}5.75 & \cellcolor[HTML]{F1B5B0}11.77 & \cellcolor[HTML]{FFFFFF}\textbf{3.5} & \cellcolor[HTML]{E67C73}18.08 & 10.47 \\ \hline \hline
 & \textbf{ar} & \cellcolor[HTML]{91D3B3}-3.73 & \cellcolor[HTML]{7ECBA5}-2.08 & \cellcolor[HTML]{9FD8BC}-7.31 & \cellcolor[HTML]{9BD7BA}-7.62 & \cellcolor[HTML]{7ACAA3}-0.79 & \cellcolor[HTML]{81CCA7}-4.96 & \cellcolor[HTML]{A6DBC1}-6.01 & \cellcolor[HTML]{88CFAC}-4.64 \\
 & \textbf{de} & \cellcolor[HTML]{76C8A0}-2.5 & \cellcolor[HTML]{A1D9BE}-4.8 & \cellcolor[HTML]{93D3B4}-5.93 & \cellcolor[HTML]{BCE4D1}-10.17 & \cellcolor[HTML]{77C8A1}-0.46 & \cellcolor[HTML]{DAF0E5}-11.23 & \cellcolor[HTML]{57BB8A}\textbf{5.85} & \cellcolor[HTML]{81CCA8}-4.18 \\
 & \textbf{en} & \cellcolor[HTML]{FFFFFF}-8.75 & \cellcolor[HTML]{F8FCFA}-11.46 & \cellcolor[HTML]{C8E9D9}-11.91 & \cellcolor[HTML]{E8F6EF}-13.53 & \cellcolor[HTML]{BAE3CF}-7.21 & \cellcolor[HTML]{C7E9D8}-9.9 & \cellcolor[HTML]{C8E9D9}-11.17 & \cellcolor[HTML]{DBF1E6}-10.56 \\
 & \textbf{es} & \cellcolor[HTML]{57BB8A}\textbf{-1.12} & \cellcolor[HTML]{5BBD8D}0.62 & \cellcolor[HTML]{57BB8A}\textbf{0.69} & \cellcolor[HTML]{7AC9A2}-5 & \cellcolor[HTML]{57BB8A}\textbf{2.74} & \cellcolor[HTML]{A1D9BE}-7.22 & \cellcolor[HTML]{78C9A1}0.92 & \cellcolor[HTML]{57BB8A}\textbf{-1.20} \\
 & \textbf{hi} & \cellcolor[HTML]{74C79E}-2.42 & \cellcolor[HTML]{BBE4D0}-6.76 & \cellcolor[HTML]{FFFFFF}-18.16 & \cellcolor[HTML]{FFFFFF}-15.39 & \cellcolor[HTML]{FFFFFF}-14.21 & \cellcolor[HTML]{FFFFFF}-13.9 & \cellcolor[HTML]{DFF2E9}-14.51 & \cellcolor[HTML]{F1FAF6}-12.19 \\
\bottomrule
\end{tabular}%
}
\caption{Actual toxicity scores for $ZS$ \textit{vs} $\Delta$-toxicity scores for $X$-$FT$ for \texttt{aya-expanse-8B} over the \textit{neutral-test} evaluation set. \textbf{\textit{Takeaway}}: \textit{Detoxification adversely effects the model's general knowledge}.}
\label{tab:tox-aya-8-neutral-test}
\end{table*}

\begin{table*}[t]
\centering \resizebox{0.85\textwidth}{!}{%
\begin{tabular}{cccccccccc}
\toprule
\multicolumn{1}{l}{} & \multicolumn{1}{l}{} & \textbf{am} & \textbf{ar} & \textbf{de} & \textbf{en} & \textbf{es} & \textbf{hi} & \textbf{ru} & \textbf{AVG} \\ \midrule
\multicolumn{1}{l}{} & \textbf{$ZS$} & \cellcolor[HTML]{FEF5F5}19.86 & \cellcolor[HTML]{FDF5F4}19.95 & \cellcolor[HTML]{E67C73}33.17 & \cellcolor[HTML]{FCEDEC}20.79 & \cellcolor[HTML]{F5C6C2}25.09 & \cellcolor[HTML]{FFFFFF}\textbf{18.75} & \cellcolor[HTML]{F8D8D5}23.1 & 22.96 \\ \hline \hline
 & \textbf{ar} & \cellcolor[HTML]{C0E6D4}2.46 & \cellcolor[HTML]{90D2B2}2.68 & \cellcolor[HTML]{96D5B6}7.54 & \cellcolor[HTML]{98D5B7}1.64 & \cellcolor[HTML]{D8EFE4}-4.14 & \cellcolor[HTML]{ABDDC5}-1.8 & \cellcolor[HTML]{C8E9D9}-5.45 & \cellcolor[HTML]{ADDEC6}0.42 \\
 & \textbf{de} & \cellcolor[HTML]{84CDA9}7.12 & \cellcolor[HTML]{AADDC4}-0.65 & \cellcolor[HTML]{72C69D}14.91 & \cellcolor[HTML]{60BF90}7.8 & \cellcolor[HTML]{60BF90}11.95 & \cellcolor[HTML]{96D5B6}-0.02 & \cellcolor[HTML]{85CEAA}3.78 & \cellcolor[HTML]{7DCBA5}6.41 \\
 & \textbf{en} & \cellcolor[HTML]{DBF1E6}0.39 & \cellcolor[HTML]{C5E8D7}-4.07 & \cellcolor[HTML]{99D6B8}6.81 & \cellcolor[HTML]{8DD1B0}2.78 & \cellcolor[HTML]{CDEBDC}-2.65 & \cellcolor[HTML]{97D5B7}-0.13 & \cellcolor[HTML]{E0F3EA}-8.74 & \cellcolor[HTML]{B7E2CD}-0.80 \\
 & \textbf{es} & \cellcolor[HTML]{57BB8A}\textbf{10.53} & \cellcolor[HTML]{57BB8A}\textbf{9.93} & \cellcolor[HTML]{57BB8A}\textbf{20.39} & \cellcolor[HTML]{57BB8A}\textbf{8.73} & \cellcolor[HTML]{57BB8A}\textbf{13.05} & \cellcolor[HTML]{57BB8A}\textbf{5.18} & \cellcolor[HTML]{57BB8A}\textbf{9.93} & \cellcolor[HTML]{57BB8A}\textbf{11.11} \\
 & \textbf{hi} & \cellcolor[HTML]{D2EDE0}1.08 & \cellcolor[HTML]{E4F4EC}-7.96 & \cellcolor[HTML]{AEDFC7}2.56 & \cellcolor[HTML]{C5E8D7}-3.38 & \cellcolor[HTML]{DCF1E7}-4.68 & \cellcolor[HTML]{CCEBDC}-4.53 & \cellcolor[HTML]{DEF2E8}-8.48 & \cellcolor[HTML]{CDEBDD}-3.63 \\
\bottomrule
\end{tabular}%
}
\caption{Actual toxicity scores for $ZS$ \textit{vs} $\Delta$-toxicity scores for $X$-$FT$ for \texttt{aya-23-8B} over the \textit{toxic-train} evaluation set. \textbf{\textit{Takeaway}}: \textit{Surprisingly ``zh'' shows that irrespective of fine-tuning language, the detoxification scores actually increases}.}
\label{tab:tox-aya-23-toxic-train}
\end{table*}
\begin{table*}[t]
\centering \resizebox{0.85\textwidth}{!}{%
\begin{tabular}{cccccccccc}
\toprule
\multicolumn{1}{l}{} & \multicolumn{1}{l}{} & \textbf{am} & \textbf{ar} & \textbf{de} & \textbf{en} & \textbf{es} & \textbf{hi} & \textbf{ru} & \textbf{AVG} \\ \midrule
\multicolumn{1}{l}{} & \textbf{$ZS$} & \cellcolor[HTML]{FFFFFF}\textbf{16.31} & \cellcolor[HTML]{FAE5E3}20.85 & \cellcolor[HTML]{E67C73}38.48 & \cellcolor[HTML]{FBE9E7}20.15 & \cellcolor[HTML]{F6CECA}24.76 & \cellcolor[HTML]{FCEFEE}19.07 & \cellcolor[HTML]{F9E0DE}21.68 & 23.04 \\ \hline\hline
 & \textbf{ar} & \cellcolor[HTML]{C6E8D7}-0.27 & \cellcolor[HTML]{A2DABE}2.27 & \cellcolor[HTML]{97D5B7}15.67 & \cellcolor[HTML]{8BD0AE}1.15 & \cellcolor[HTML]{B3E1CB}-2.37 & \cellcolor[HTML]{BBE4D0}-3.88 & \cellcolor[HTML]{B9E3CE}-5.6 & \cellcolor[HTML]{A9DCC3}1.00 \\
 & \textbf{de} & \cellcolor[HTML]{8BD0AE}3.76 & \cellcolor[HTML]{AEDEC7}0.85 & \cellcolor[HTML]{A5DBC1}12.97 & \cellcolor[HTML]{57BB8A}7.62 & \cellcolor[HTML]{6EC49A}9.52 & \cellcolor[HTML]{ADDEC6}-2.62 & \cellcolor[HTML]{86CEAB}1.01 & \cellcolor[HTML]{8BD1AF}4.73 \\
 & \textbf{en} & \cellcolor[HTML]{DEF2E8}-1.87 & \cellcolor[HTML]{D2EDE0}-3.51 & \cellcolor[HTML]{B1E0C9}10.66 & \cellcolor[HTML]{86CEAB}1.76 & \cellcolor[HTML]{C7E9D8}-5.66 & \cellcolor[HTML]{7DCBA5}1.83 & \cellcolor[HTML]{B5E1CC}-5.12 & \cellcolor[HTML]{B3E1CA}-0.27 \\
 & \textbf{es} & \cellcolor[HTML]{57BB8A}\textbf{7.22} & \cellcolor[HTML]{57BB8A}\textbf{11.28} & \cellcolor[HTML]{57BB8A}\textbf{27.76} & \cellcolor[HTML]{5CBD8E}\textbf{7.02} & \cellcolor[HTML]{57BB8A}\textbf{13.27} & \cellcolor[HTML]{57BB8A}\textbf{5.32} & \cellcolor[HTML]{57BB8A}\textbf{7.13} & \cellcolor[HTML]{57BB8A}\textbf{11.29} \\
 & \textbf{hi} & \cellcolor[HTML]{D8F0E4}-1.5 & \cellcolor[HTML]{DEF2E8}-5.02 & \cellcolor[HTML]{C5E8D7}6.99 & \cellcolor[HTML]{96D5B6}-0.3 & \cellcolor[HTML]{D5EEE2}-8.12 & \cellcolor[HTML]{CBEADB}-5.36 & \cellcolor[HTML]{CBEADB}-7.95 & \cellcolor[HTML]{C9E9D9}-3.04 \\
\bottomrule
\end{tabular}%
}
\caption{Actual toxicity scores for $ZS$ \textit{vs} $\Delta$-toxicity scores for $X$-$FT$ for \texttt{aya-23-8B} over the \textit{toxic-test} evaluation set. \textbf{\textit{Takeaway}}: \textit{``es'' showed the best average detoxification scores}.}
\label{tab:tox-aya-23-toxic-test}
\end{table*}
\begin{table*}[t]
\centering \resizebox{0.85\textwidth}{!}{%
\begin{tabular}{cccccccccc}
\toprule
\multicolumn{1}{l}{} & \multicolumn{1}{l}{} & \textbf{am} & \textbf{ar} & \textbf{de} & \textbf{en} & \textbf{es} & \textbf{hi} & \textbf{ru} & \textbf{AVG} \\ \midrule
\multicolumn{1}{l}{} & \textbf{$ZS$} & \cellcolor[HTML]{F6D0CD}11.5 & \cellcolor[HTML]{F6CDCA}11.9 & \cellcolor[HTML]{F4C4C0}13.08 & \cellcolor[HTML]{F9DFDD}9.47 & \cellcolor[HTML]{F4C4BF}13.17 & \cellcolor[HTML]{FFFFFF}\textbf{5.22} & \cellcolor[HTML]{E67C73}22.61 & 12.42 \\\hline\hline
 & \textbf{ar} & \cellcolor[HTML]{CFECDE}-2.28 & \cellcolor[HTML]{7DCBA4}-1.23 & \cellcolor[HTML]{AADDC4}-9.14 & \cellcolor[HTML]{B4E1CB}-6.17 & \cellcolor[HTML]{A8DCC2}-5.86 & \cellcolor[HTML]{ABDDC5}-5.91 & \cellcolor[HTML]{BFE6D3}-5.07 & \cellcolor[HTML]{A7DCC2}-5.09 \\
 & \textbf{de} & \cellcolor[HTML]{C3E7D6}-1.67 & \cellcolor[HTML]{B8E3CE}-6.39 & \cellcolor[HTML]{8ED1B0}-5.3 & \cellcolor[HTML]{92D3B3}-3.4 & \cellcolor[HTML]{73C69D}-0.64 & \cellcolor[HTML]{FFFFFF}-12.95 & \cellcolor[HTML]{7ECBA6}3.4 & \cellcolor[HTML]{98D6B7}-3.85 \\
 & \textbf{en} & \cellcolor[HTML]{FFFFFF}-4.81 & \cellcolor[HTML]{CFECDE}-8.4 & \cellcolor[HTML]{A5DBC1}-8.5 & \cellcolor[HTML]{A8DCC3}-5.21 & \cellcolor[HTML]{C2E7D5}-8.42 & \cellcolor[HTML]{AFDFC8}-6.28 & \cellcolor[HTML]{BBE4D0}-4.54 & \cellcolor[HTML]{B9E3CE}-6.59 \\
 & \textbf{es} & \cellcolor[HTML]{57BB8A}\textbf{3.96} & \cellcolor[HTML]{57BB8A}\textbf{2} & \cellcolor[HTML]{57BB8A}\textbf{2.05} & \cellcolor[HTML]{88CFAC}-2.64 & \cellcolor[HTML]{57BB8A}\textbf{2.03} & \cellcolor[HTML]{A2DABF}-5.21 & \cellcolor[HTML]{57BB8A}\textbf{8.5} & \cellcolor[HTML]{57BB8A}\textbf{1.53} \\
 & \textbf{hi} & \cellcolor[HTML]{B9E3CE}-1.11 & \cellcolor[HTML]{DEF2E8}-9.66 & \cellcolor[HTML]{BEE5D2}-11.88 & \cellcolor[HTML]{B9E3CE}-6.56 & \cellcolor[HTML]{F3FBF7}-13.28 & \cellcolor[HTML]{B4E1CB}-6.7 & \cellcolor[HTML]{C6E8D8}-5.96 & \cellcolor[HTML]{C8E9D9}-7.88 \\
\bottomrule
\end{tabular}%
}
\caption{
Actual toxicity scores for $ZS$ \textit{vs} $\Delta$-toxicity scores for $X$-$FT$ for \texttt{aya-23-8B} over the \textit{neutral-test} evaluation set. \textbf{\textit{Takeaway}}: \textit{Detoxification adversely effects the model’s general knowledge}.}
\label{tab:tox-aya-23-neutral-test}
\end{table*}

\begin{table*}[t]
\centering \resizebox{0.85\textwidth}{!}{%
\begin{tabular}{cccccccccc}
\toprule
\multicolumn{1}{l}{} & \multicolumn{1}{l}{} & \textbf{am} & \textbf{ar} & \textbf{de} & \textbf{en} & \textbf{es} & \textbf{hi} & \textbf{ru} & \textbf{AVG} \\ \midrule
\multicolumn{1}{l}{} & \textbf{$ZS$} & \cellcolor[HTML]{FEFAFA}17.94 & \cellcolor[HTML]{FEF9F9}18.12 & \cellcolor[HTML]{E67C73}43.58 & \cellcolor[HTML]{F6D0CC}26.6 & \cellcolor[HTML]{FAE1DF}23.06 & \cellcolor[HTML]{F8D6D3}25.39 & \cellcolor[HTML]{FFFFFF}\textbf{16.86} & 24.51 \\\hline\hline
 & \textbf{ar} & \cellcolor[HTML]{73C79E}10.92 & \cellcolor[HTML]{F6FCF9}7.27 & \cellcolor[HTML]{FFFFFF}27.16 & \cellcolor[HTML]{B6E2CC}16 & \cellcolor[HTML]{74C79F}13.26 & \cellcolor[HTML]{B4E1CB}15.89 & \cellcolor[HTML]{D2EDE0}8.74 & \cellcolor[HTML]{FFFFFF}14.18 \\
 & \textbf{de} & \cellcolor[HTML]{FFFFFF}9.7 & \cellcolor[HTML]{69C397}9.47 & \cellcolor[HTML]{9CD7BA}29.25 & \cellcolor[HTML]{D0ECDE}15.62 & \cellcolor[HTML]{88CFAC}12.93 & \cellcolor[HTML]{AADDC4}16.09 & \cellcolor[HTML]{96D5B6}9.09 & \cellcolor[HTML]{9CD7BA}14.59 \\
 & \textbf{en} & \cellcolor[HTML]{98D6B7}10.6 & \cellcolor[HTML]{9FD8BC}8.63 & \cellcolor[HTML]{E0F3EA}27.82 & \cellcolor[HTML]{67C295}17.17 & \cellcolor[HTML]{6EC49A}13.38 & \cellcolor[HTML]{B6E2CC}15.86 & \cellcolor[HTML]{57BB8A}\textbf{9.45} & \cellcolor[HTML]{82CDA8}14.70 \\
 & \textbf{es} & \cellcolor[HTML]{65C194}11.04 & \cellcolor[HTML]{66C194}9.52 & \cellcolor[HTML]{B9E3CE}28.64 & \cellcolor[HTML]{7DCBA4}16.85 & \cellcolor[HTML]{FFFFFF}10.86 & \cellcolor[HTML]{57BB8A}\textbf{17.62} & \cellcolor[HTML]{FFFFFF}8.48 & \cellcolor[HTML]{7FCBA6}14.72 \\
 & \textbf{hi} & \cellcolor[HTML]{75C89F}10.9 & \cellcolor[HTML]{57BB8A}\textbf{9.74} & \cellcolor[HTML]{6EC59A}30.2 & \cellcolor[HTML]{C7E9D8}15.75 & \cellcolor[HTML]{77C8A1}13.21 & \cellcolor[HTML]{B7E2CD}15.85 & \cellcolor[HTML]{FAFDFC}8.51 & \cellcolor[HTML]{57BB8A}\textbf{14.88} \\
\bottomrule
\end{tabular}%
}
\caption{Actual toxicity scores for $ZS$ \textit{vs} $\Delta$-toxicity scores for $X$-$FT$ for \texttt{mt5-large} over the \textit{toxic-train} evaluation set. \textbf{\textit{Takeaway}}: \textit{\texttt{mt5-large} showed better detoxification scores in all languages but showed a trade-off with general perplexity scores}.}
\label{tab:tox-mt5-toxic-train}
\end{table*}
\begin{table*}[t]
\centering \resizebox{0.85\textwidth}{!}{%
\begin{tabular}{cccccccccc}
\toprule
\multicolumn{1}{l}{} & \multicolumn{1}{l}{} & \textbf{am} & \textbf{ar} & \textbf{de} & \textbf{en} & \textbf{es} & \textbf{hi} & \textbf{ru} & \textbf{AVG} \\ \midrule
\multicolumn{1}{l}{} & \textbf{$ZS$} & \cellcolor[HTML]{FDF4F3}18.12 & \cellcolor[HTML]{FFFFFF}\textbf{15.25} & \cellcolor[HTML]{E67C73}46.74 & \cellcolor[HTML]{F6CBC8}27.83 & \cellcolor[HTML]{F8D6D3}25.21 & \cellcolor[HTML]{F2B9B4}32.26 & \cellcolor[HTML]{FDF2F2}18.38 & 26.26 \\\hline\hline
 & \textbf{ar} & \cellcolor[HTML]{C0E6D3}10.23 & \cellcolor[HTML]{ACDEC6}6.53 & \cellcolor[HTML]{57BB8A}\textbf{35} & \cellcolor[HTML]{68C296}18.8 & \cellcolor[HTML]{D6EFE3}15.73 & \cellcolor[HTML]{57BB8A}\textbf{23.9} & \cellcolor[HTML]{57BB8A}\textbf{11.75} & \cellcolor[HTML]{57BB8A}\textbf{17.42} \\
 & \textbf{de} & \cellcolor[HTML]{8CD1AF}10.91 & \cellcolor[HTML]{FFFFFF}4.69 & \cellcolor[HTML]{C0E6D3}32.02 & \cellcolor[HTML]{BCE4D0}16.38 & \cellcolor[HTML]{FFFFFF}14.97 & \cellcolor[HTML]{5ABD8C}23.86 & \cellcolor[HTML]{A2DABE}10.78 & \cellcolor[HTML]{DAF0E6}16.23 \\
 & \textbf{en} & \cellcolor[HTML]{92D3B3}10.84 & \cellcolor[HTML]{F1FAF5}5.01 & \cellcolor[HTML]{FFFFFF}30.22 & \cellcolor[HTML]{C9E9D9}16 & \cellcolor[HTML]{57BB8A}\textbf{18.08} & \cellcolor[HTML]{67C295}23.69 & \cellcolor[HTML]{FFFFFF}9.56 & \cellcolor[HTML]{DEF2E8}16.20 \\
 & \textbf{es} & \cellcolor[HTML]{FFFFFF}9.4 & \cellcolor[HTML]{A9DCC3}6.61 & \cellcolor[HTML]{DBF1E6}31.27 & \cellcolor[HTML]{ADDEC6}16.81 & \cellcolor[HTML]{E9F7F0}15.38 & \cellcolor[HTML]{DAF0E5}22.08 & \cellcolor[HTML]{F6FBF9}9.69 & \cellcolor[HTML]{FFFFFF}15.89 \\
 & \textbf{hi} & \cellcolor[HTML]{C6E8D8}10.15 & \cellcolor[HTML]{57BB8A}\textbf{8.41} & \cellcolor[HTML]{ADDEC6}32.58 & \cellcolor[HTML]{FFFFFF}14.42 & \cellcolor[HTML]{9CD7BA}16.81 & \cellcolor[HTML]{92D3B3}23.08 & \cellcolor[HTML]{99D6B8}10.9 & \cellcolor[HTML]{AFDFC8}16.62 \\
\bottomrule
\end{tabular}%
}
\caption{Actual toxicity scores for $ZS$ \textit{vs} $\Delta$-toxicity scores for $X$-$FT$ for \texttt{mt5-large} over the \textit{toxic-test} evaluation set. \textbf{\textit{Takeaway}}: \textit{\texttt{mt5-large} showed better detoxification scores in all languages but showed a trade-off with general perplexity scores}.}
\label{tab:tox-mt5-toxic-test}
\end{table*}
\begin{table*}[t]
\centering \resizebox{0.85\textwidth}{!}{%
\begin{tabular}{cccccccccc}
\toprule
\multicolumn{1}{l}{} & \multicolumn{1}{l}{} & \textbf{am} & \textbf{ar} & \textbf{de} & \textbf{en} & \textbf{es} & \textbf{hi} & \textbf{ru} & \textbf{AVG} \\ \midrule
\multicolumn{1}{l}{} & \textbf{$ZS$} & \cellcolor[HTML]{F6CECB}8.03 & \cellcolor[HTML]{F3BEBA}9.44 & \cellcolor[HTML]{E67C73}15.14 & \cellcolor[HTML]{F9DEDC}6.66 & \cellcolor[HTML]{EC9A93}12.54 & \cellcolor[HTML]{FFFFFF}\textbf{3.74} & \cellcolor[HTML]{E98B83}13.89 & 9.92 \\\hline\hline
 & \textbf{ar} & \cellcolor[HTML]{68C296}2.73 & \cellcolor[HTML]{5ABC8C}2.31 & \cellcolor[HTML]{93D3B4}8.29 & \cellcolor[HTML]{57BB8A}\textbf{1.43} & \cellcolor[HTML]{8BD0AE}6.13 & \cellcolor[HTML]{97D5B7}-0.74 & \cellcolor[HTML]{87CFAC}7.95 & \cellcolor[HTML]{57BB8A}\textbf{4.01} \\
 & \textbf{de} & \cellcolor[HTML]{85CEAA}2.49 & \cellcolor[HTML]{DFF2E9}1.19 & \cellcolor[HTML]{FFFFFF}7.48 & \cellcolor[HTML]{7ECBA5}0.82 & \cellcolor[HTML]{57BB8A}\textbf{6.68} & \cellcolor[HTML]{57BB8A}\textbf{-0.02} & \cellcolor[HTML]{79C9A2}\textbf{8.17} & \cellcolor[HTML]{73C69D}3.83 \\
 & \textbf{en} & \cellcolor[HTML]{67C295}2.74 & \cellcolor[HTML]{FFFFFF}0.92 & \cellcolor[HTML]{F2FAF6}7.58 & \cellcolor[HTML]{FFFFFF}-1.25 & \cellcolor[HTML]{FFFFFF}4.88 & \cellcolor[HTML]{C9E9D9}-1.3 & \cellcolor[HTML]{E0F3E9}6.54 & \cellcolor[HTML]{FFFFFF}2.87 \\
 & \textbf{es} & \cellcolor[HTML]{FFFFFF}1.46 & \cellcolor[HTML]{57BB8A}\textbf{2.33} & \cellcolor[HTML]{EEF8F3}7.61 & \cellcolor[HTML]{94D4B4}0.47 & \cellcolor[HTML]{9BD7B9}5.96 & \cellcolor[HTML]{F8FDFB}-1.84 & \cellcolor[HTML]{FFFFFF}6.03 & \cellcolor[HTML]{D7EFE4}3.15 \\
 & \textbf{hi} & \cellcolor[HTML]{57BB8A}\textbf{2.87} & \cellcolor[HTML]{B6E2CC}1.54 & \cellcolor[HTML]{9DD8BB}8.21 & \cellcolor[HTML]{83CDA9}0.73 & \cellcolor[HTML]{73C69D}6.39 & \cellcolor[HTML]{84CDA9}-0.52 & \cellcolor[HTML]{57BB8A}8.71 & \cellcolor[HTML]{5BBD8D}3.99 \\
\bottomrule
\end{tabular}%
}
\caption{Actual toxicity scores for $ZS$ \textit{vs} $\Delta$-toxicity scores for $X$-$FT$ for \texttt{mt5-large} over the \textit{neutral-test} evaluation set. \textbf{\textit{Takeaway}}: \textit{Detoxification does not adversely effects the model’s general knowledge but effected the overall perplexity meanwhile}.}
\label{tab:tox-mt5-neutral-test}
\end{table*}

\begin{table*}[t]
\centering \resizebox{0.85\textwidth}{!}{%
\begin{tabular}{lccccccccc}
\toprule
 & \multicolumn{1}{l}{} & \textbf{am} & \textbf{ar} & \textbf{de} & \textbf{en} & \textbf{es} & \textbf{hi} & \textbf{ru} & \textbf{AVG} \\ \midrule
 & \textbf{$ZS$} & \cellcolor[HTML]{FDF1F0}15.53 & \cellcolor[HTML]{FDF4F3}15.2 & \cellcolor[HTML]{E67C73}28.75 & \cellcolor[HTML]{FDF1F0}15.51 & \cellcolor[HTML]{E98981}27.37 & \cellcolor[HTML]{FFFFFF}\textbf{13.9} & \cellcolor[HTML]{F0AFA9}23.03 & 19.90 \\\hline\hline
 & \textbf{ar} & \cellcolor[HTML]{FFFFFF}-5.8 & \cellcolor[HTML]{84CEAA}3.91 & \cellcolor[HTML]{74C79F}13.11 & \cellcolor[HTML]{8FD2B1}-0.84 & \cellcolor[HTML]{9BD7B9}7.3 & \cellcolor[HTML]{B6E2CC}-2.35 & \cellcolor[HTML]{BBE4D0}4.29 & \cellcolor[HTML]{A6DBC1}2.80 \\
 & \textbf{de} & \cellcolor[HTML]{5EBE8F}2.74 & \cellcolor[HTML]{6EC59A}5.25 & \cellcolor[HTML]{8AD0AE}11.3 & \cellcolor[HTML]{57BB8A}\textbf{4.95} & \cellcolor[HTML]{8BD0AE}9.47 & \cellcolor[HTML]{57BB8A}\textbf{6.54} & \cellcolor[HTML]{7FCBA6}9.11 & \cellcolor[HTML]{6AC397}7.05 \\
 & \textbf{en} & \cellcolor[HTML]{99D6B8}-0.39 & \cellcolor[HTML]{FFFFFF}-3.42 & \cellcolor[HTML]{FFFFFF}1.58 & \cellcolor[HTML]{FFFFFF}-12.43 & \cellcolor[HTML]{9ED8BB}6.91 & \cellcolor[HTML]{BAE3CF}-2.74 & \cellcolor[HTML]{66C194}11.12 & \cellcolor[HTML]{CCEBDC}0.09 \\
 & \textbf{es} & \cellcolor[HTML]{57BB8A}\textbf{3.11} & \cellcolor[HTML]{57BB8A}\textbf{6.59} & \cellcolor[HTML]{80CCA7}12.14 & \cellcolor[HTML]{63C092}3.81 & \cellcolor[HTML]{57BB8A}\textbf{16.47} & \cellcolor[HTML]{70C59B}4.26 & \cellcolor[HTML]{57BB8A}\textbf{12.25} & \cellcolor[HTML]{57BB8A}\textbf{8.38} \\
 & \textbf{hi} & \cellcolor[HTML]{F5FBF8}-5.25 & \cellcolor[HTML]{F7FCFA}-2.93 & \cellcolor[HTML]{A2D9BE}9.36 & \cellcolor[HTML]{DFF2E9}-9.1 & \cellcolor[HTML]{FFFFFF}-6.45 & \cellcolor[HTML]{FFFFFF}-9.33 & \cellcolor[HTML]{FFFFFF}-1.13 & \cellcolor[HTML]{FFFFFF}-3.55 \\
\bottomrule
\end{tabular}%
}
\caption{Actual toxicity scores for $ZS$ \textit{vs} $\Delta$-toxicity scores for $X$-$FT$ for \texttt{bloom-7B1} over the \textit{toxic-train} evaluation set. \textbf{\textit{Takeaway}}: \textit{``es'' comes up as the best fine-tuning language than ``en'' and ``de'' from other models}.}
\label{tab:tox-bloom-toxic-train}
\end{table*}
\begin{table*}[t]
\centering \resizebox{0.85\textwidth}{!}{%
\begin{tabular}{cccccccccc}
\toprule
\multicolumn{1}{l}{} & \multicolumn{1}{l}{} & \textbf{am} & \textbf{ar} & \textbf{de} & \textbf{en} & \textbf{es} & \textbf{hi} & \textbf{ru} & \textbf{AVG} \\ \midrule
\multicolumn{1}{l}{} & \textbf{$ZS$} & \cellcolor[HTML]{F9DCD9}17.17 & \cellcolor[HTML]{FFFFFF}\textbf{12.25} & \cellcolor[HTML]{EFA6A0}24.51 & \cellcolor[HTML]{F5C9C6}19.69 & \cellcolor[HTML]{E67C73}30.28 & \cellcolor[HTML]{FFFEFE}12.46 & \cellcolor[HTML]{F1B3AE}22.77 & 19.88 \\\hline\hline
 & \textbf{ar} & \cellcolor[HTML]{FFFFFF}-3.74 & \cellcolor[HTML]{8CD1AF}0.44 & \cellcolor[HTML]{78C8A1}7.99 & \cellcolor[HTML]{80CCA6}2.2 & \cellcolor[HTML]{95D5B6}10.63 & \cellcolor[HTML]{B2E0CA}-4.16 & \cellcolor[HTML]{C7E9D8}2.86 & \cellcolor[HTML]{ACDEC5}2.32 \\
 & \textbf{de} & \cellcolor[HTML]{88CFAC}2.96 & \cellcolor[HTML]{7AC9A3}1.46 & \cellcolor[HTML]{76C8A0}8.08 & \cellcolor[HTML]{57BB8A}\textbf{7.72} & \cellcolor[HTML]{81CCA7}13.28 & \cellcolor[HTML]{57BB8A}\textbf{4.52} & \cellcolor[HTML]{57BB8A}\textbf{11.88} & \cellcolor[HTML]{64C193}7.13 \\
 & \textbf{en} & \cellcolor[HTML]{A2DABE}1.48 & \cellcolor[HTML]{FFFFFF}-6.04 & \cellcolor[HTML]{FFFFFF}-3.78 & \cellcolor[HTML]{FFFFFF}-15.4 & \cellcolor[HTML]{78C9A1}14.43 & \cellcolor[HTML]{A1D9BE}-2.5 & \cellcolor[HTML]{79C9A2}9.17 & \cellcolor[HTML]{D4EEE1}-0.38 \\
 & \textbf{es} & \cellcolor[HTML]{57BB8A}\textbf{5.65} & \cellcolor[HTML]{58BC8B}3.37 & \cellcolor[HTML]{81CCA7}7.18 & \cellcolor[HTML]{59BC8B}7.56 & \cellcolor[HTML]{57BB8A}\textbf{18.68} & \cellcolor[HTML]{68C296}2.98 & \cellcolor[HTML]{68C296}10.54 & \cellcolor[HTML]{57BB8A}\textbf{7.99} \\
 & \textbf{hi} & \cellcolor[HTML]{EDF8F3}-2.73 & \cellcolor[HTML]{FFFFFF}-6.02 & \cellcolor[HTML]{A6DBC1}3.93 & \cellcolor[HTML]{9DD8BB}-1.83 & \cellcolor[HTML]{FFFFFF}-3.14 & \cellcolor[HTML]{FFFFFF}-11.53 & \cellcolor[HTML]{FFFFFF}-1.71 & \cellcolor[HTML]{FFFFFF}-3.29 \\
\bottomrule
\end{tabular}%
}
\caption{Actual toxicity scores for $ZS$ \textit{vs} $\Delta$-toxicity scores for $X$-$FT$ for \texttt{bloom-7B1} over the \textit{toxic-test} evaluation set. \textbf{\textit{Takeaway}}: \textit{``hi'' was least effected by the fine-tuning}.}
\label{tab:tox-bloom-toxic-test}
\end{table*}
\begin{table*}[t]
\centering \resizebox{0.85\textwidth}{!}{%
\begin{tabular}{cccccccccc}
\toprule
\multicolumn{1}{l}{} & \multicolumn{1}{l}{} & \textbf{am} & \textbf{ar} & \textbf{de} & \textbf{en} & \textbf{es} & \textbf{hi} & \textbf{ru} & \textbf{AVG} \\ \midrule
\multicolumn{1}{l}{} & \textbf{$ZS$} & \cellcolor[HTML]{F5CAC6}10.34 & \cellcolor[HTML]{F7D5D2}8.69 & \cellcolor[HTML]{F6D0CD}9.36 & \cellcolor[HTML]{FBE7E5}5.96 & \cellcolor[HTML]{F0AFA9}14.38 & \cellcolor[HTML]{FFFFFF}\textbf{2.21} & \cellcolor[HTML]{E67C73}22.01 & 10.42 \\\hline\hline
 & \textbf{ar} & \cellcolor[HTML]{DDF1E7}-9.23 & \cellcolor[HTML]{8CD1AF}-2.27 & \cellcolor[HTML]{A7DCC2}-5.94 & \cellcolor[HTML]{8FD2B1}-8.22 & \cellcolor[HTML]{A1D9BD}-4.17 & \cellcolor[HTML]{BEE5D2}-14.93 & \cellcolor[HTML]{D5EEE2}4.13 & \cellcolor[HTML]{A7DCC2}-5.80 \\
 & \textbf{de} & \cellcolor[HTML]{80CCA7}-2.94 & \cellcolor[HTML]{6DC49A}-0.32 & \cellcolor[HTML]{57BB8A}\textbf{-0.14} & \cellcolor[HTML]{57BB8A}\textbf{-3.42} & \cellcolor[HTML]{8AD0AE}-1.34 & \cellcolor[HTML]{62C091}-5.6 & \cellcolor[HTML]{9CD7BA}7.83 & \cellcolor[HTML]{63C092}-0.85 \\
 & \textbf{en} & \cellcolor[HTML]{A5DBC0}-5.43 & \cellcolor[HTML]{F0F9F5}-8.67 & \cellcolor[HTML]{FFFFFF}-12.39 & \cellcolor[HTML]{D2EDE0}-13.95 & \cellcolor[HTML]{95D4B5}-2.73 & \cellcolor[HTML]{B7E2CD}-14.16 & \cellcolor[HTML]{75C79F}10.4 & \cellcolor[HTML]{B4E1CB}-6.70 \\
 & \textbf{es} & \cellcolor[HTML]{57BB8A}\textbf{-0.21} & \cellcolor[HTML]{68C296}0.03 & \cellcolor[HTML]{BDE5D1}-7.56 & \cellcolor[HTML]{64C193}-4.53 & \cellcolor[HTML]{57BB8A}\textbf{5.04} & \cellcolor[HTML]{5FBF90}-5.37 & \cellcolor[HTML]{57BB8A}\textbf{12.31} & \cellcolor[HTML]{57BB8A}\textbf{-0.04} \\
 & \textbf{hi} & \cellcolor[HTML]{FFFFFF}-11.58 & \cellcolor[HTML]{FFFFFF}-9.67 & \cellcolor[HTML]{DEF2E8}-9.93 & \cellcolor[HTML]{FFFFFF}-17.85 & \cellcolor[HTML]{FFFFFF}-16.13 & \cellcolor[HTML]{FFFFFF}-21.51 & \cellcolor[HTML]{FFFFFF}1.35 & \cellcolor[HTML]{FFFFFF}-12.19 \\
\bottomrule
\end{tabular}%
}
\caption{Actual toxicity scores for $ZS$ \textit{vs} $\Delta$-toxicity scores for $X$-$FT$ for \texttt{bloom-7B1} over the \textit{neutral-test} evaluation set. 
\textbf{\textit{Takeaway}}: \textit{Detoxification adversely effects the model’s general knowledge}.}
\label{tab:tox-bloom-neutral-test}
\end{table*}

\begin{figure}[t!]
\begin{center}
\includegraphics[width=0.8\columnwidth]{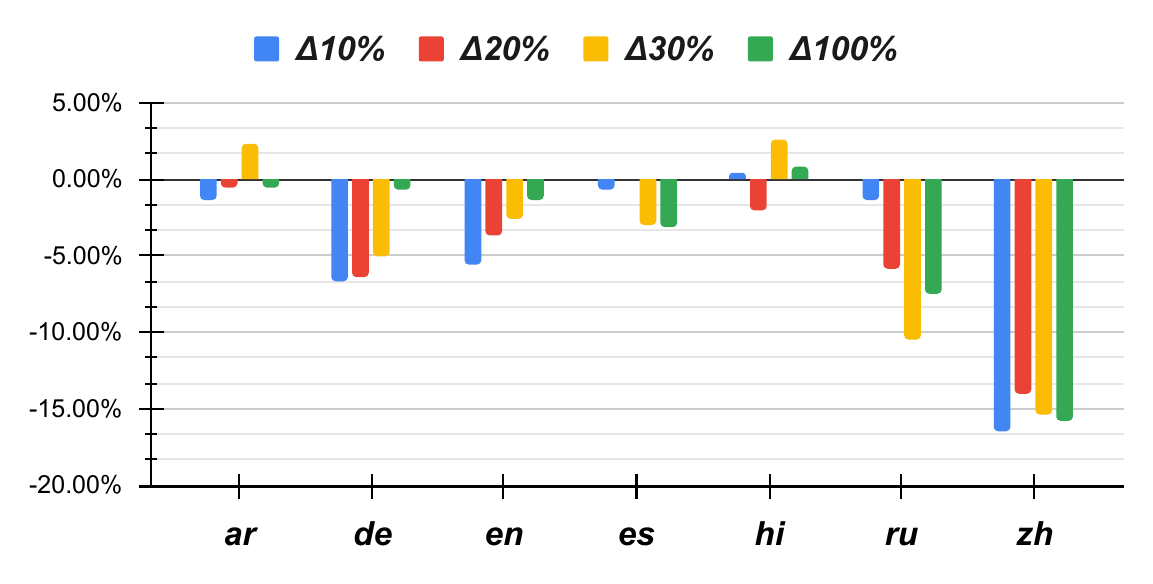}
\end{center}
\caption{Average$\Delta$-Toxicity scores for Percent-based Fine-Tuning ($P$-$FT$) \textit{vs} Multilingual Fine-Tuning ($M$-$FT$) for \texttt{aya-expanse-8B} over the \textit{toxic-train} evaluation set. $10\%$, $20\%$, $30\%$, and $100\%$ represents the Average $\Delta$-Toxicity in $P$-$FT$ and $M$-$FT$ settings. \textbf{\textit{Takeaway}}: \textit{$P$-$FT$ and $M-FT$ did not showed significant detoxification scores.}}
\label{fig:delta-aya-8-toxic-train}
\end{figure}

\begin{figure}[t!]
\begin{center}
\includegraphics[width=0.8\columnwidth]{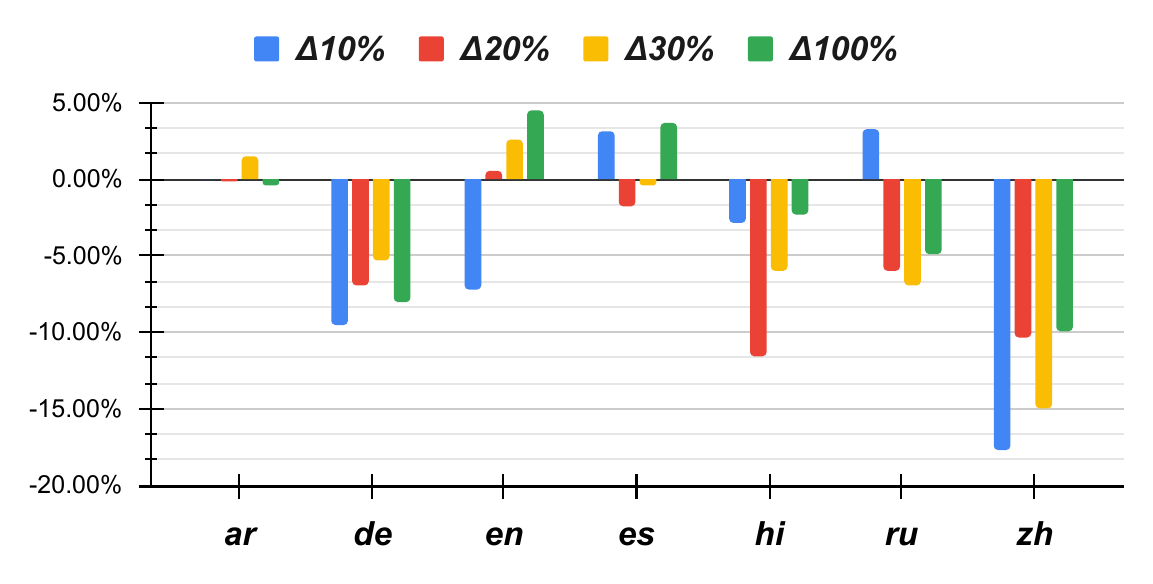}
\end{center}
\caption{Average$\Delta$-Toxicity scores for $P$-$FT$ \textit{vs} $M$-$FT$ for \texttt{aya-expanse-8B} over the \textit{toxic-test} evaluation set. \textbf{\textit{Takeaway}}: \textit{We observed significant scores in ``en'' and ``es'', but the scores did not showed any improvement in ``zh''.}}
\label{fig:delta-aya-8-toxic-test}
\end{figure}

\begin{figure}[t!]
\begin{center}
\includegraphics[width=0.8\columnwidth]{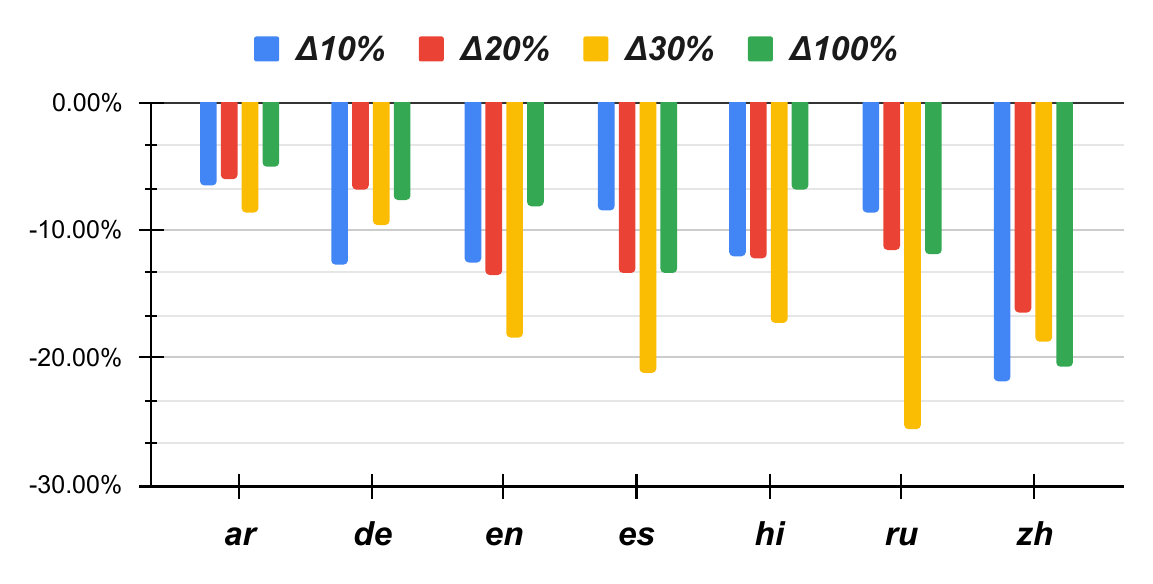}
\end{center}
\caption{Average$\Delta$-Toxicity scores for $P$-$FT$ \textit{vs} $M$-$FT$ for \texttt{aya-expanse-8B} over the \textit{neutral-test} evaluation set. \textbf{\textit{Takeaway}}: \textit{All the languages were adversely affected.}}
\label{fig:delta-aya-8-neutral-test}
\end{figure}


\begin{figure}[t!]
\begin{center}
\includegraphics[width=0.8\columnwidth]{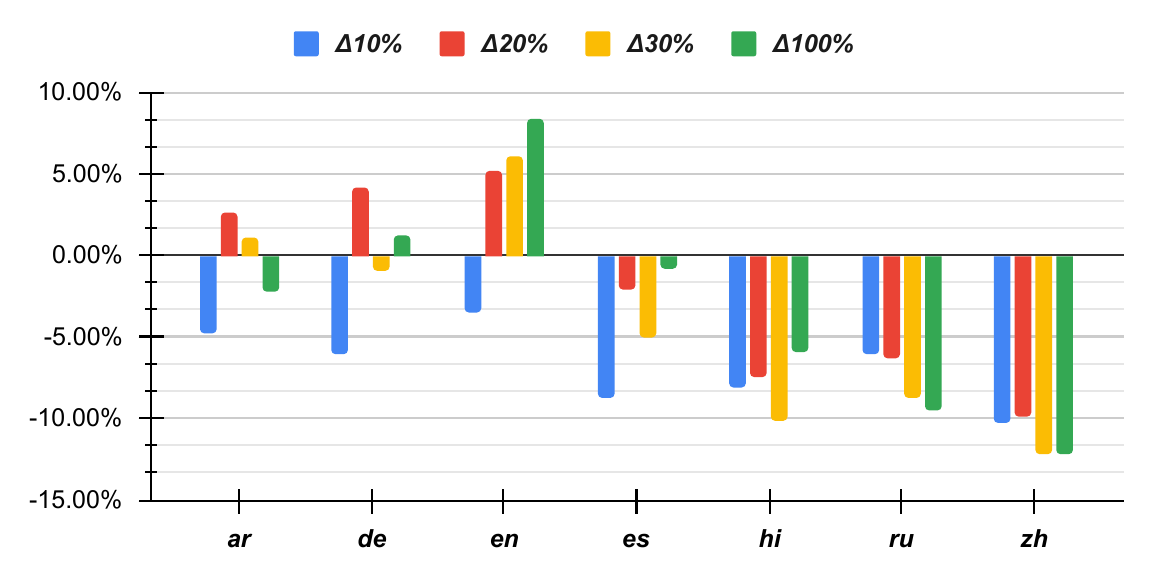}
\end{center}
\caption{Average$\Delta$-Toxicity scores for $P$-$FT$ \textit{vs} $M$-$FT$ for \texttt{aya-23-8B} over the \textit{toxic-test} evaluation set. \textbf{\textit{Takeaway}}: \textit{``en'' and ``de'' showed significant update however other showed adversarial effects.}}
\label{fig:delta-aya-23-toxic-test}
\end{figure}

\begin{figure}[t!]
\begin{center}
\includegraphics[width=0.8\columnwidth]{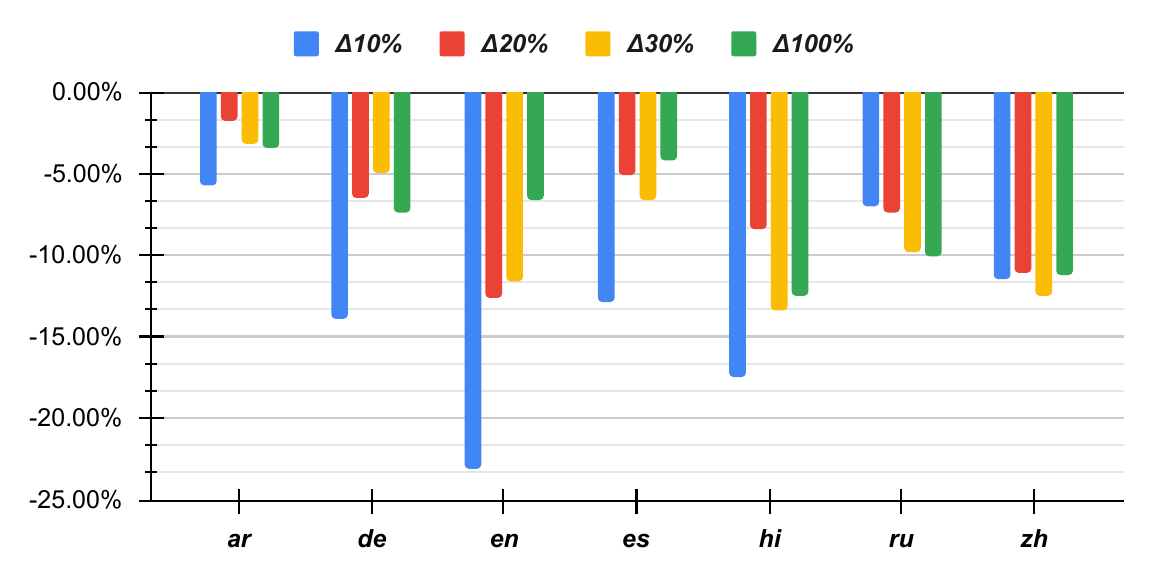}
\end{center}
\caption{Average$\Delta$-Toxicity scores for $P$-$FT$ \textit{vs} $M$-$FT$ for \texttt{aya-23-8B} over the \textit{neutral-test} evaluation set. \textbf{\textit{Takeaway}}: \textit{All the languages were adversely
affected.}}
\label{fig:delta-aya-23-neutral-test}
\end{figure}

\begin{figure}[t!]
\begin{center}
\includegraphics[width=0.8\columnwidth]{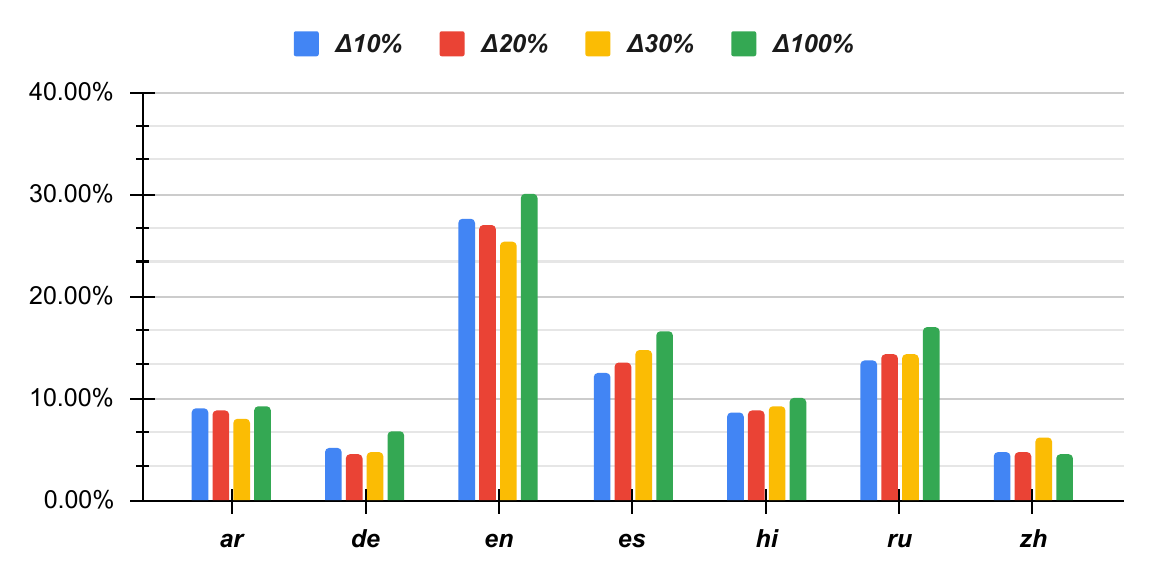}
\end{center}
\caption{Average$\Delta$-Toxicity scores for $P$-$FT$ \textit{vs} $M$-$FT$ for \texttt{mt5-large} over the \textit{toxic-train} evaluation set. \textbf{\textit{Takeaway}}: \textit{All languages showed significant updates.}}
\label{fig:delta-mt5-toxic-train}
\end{figure}

\begin{figure}[t!]
\begin{center}
\includegraphics[width=0.8\columnwidth]{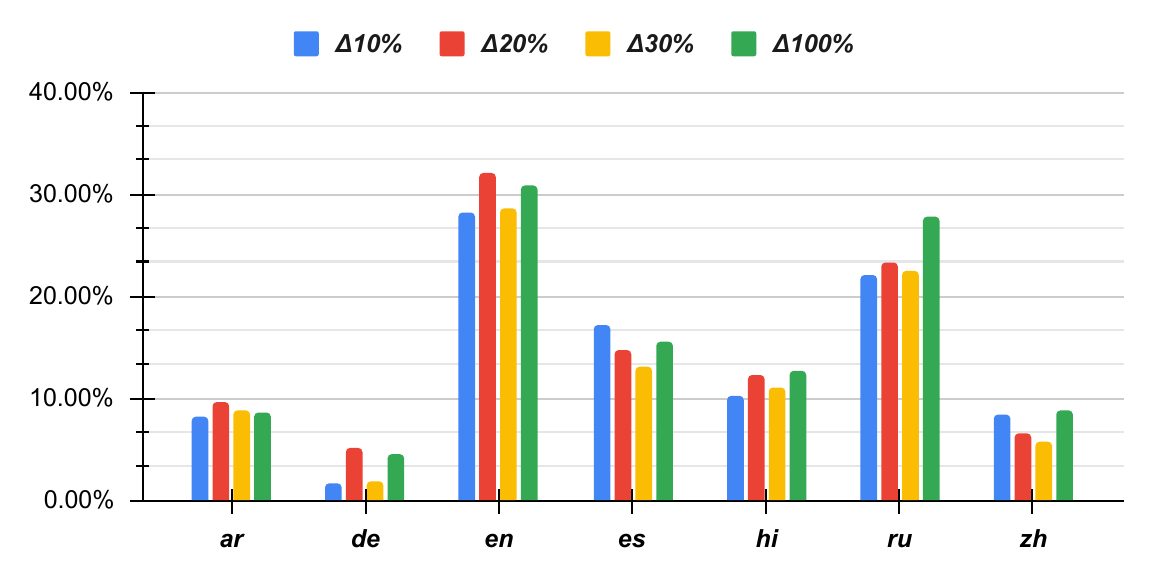}
\end{center}
\caption{Average$\Delta$-Toxicity scores for $P$-$FT$ \textit{vs} $M$-$FT$ for \texttt{mt5-large} over the \textit{toxic-test} evaluation set.  \textbf{\textit{Takeaway}}: \textit{All languages showed significant updates.}}
\label{fig:delta-mt5-toxic-test}
\end{figure}

\begin{figure}[t!]
\begin{center}
\includegraphics[width=0.8\columnwidth]{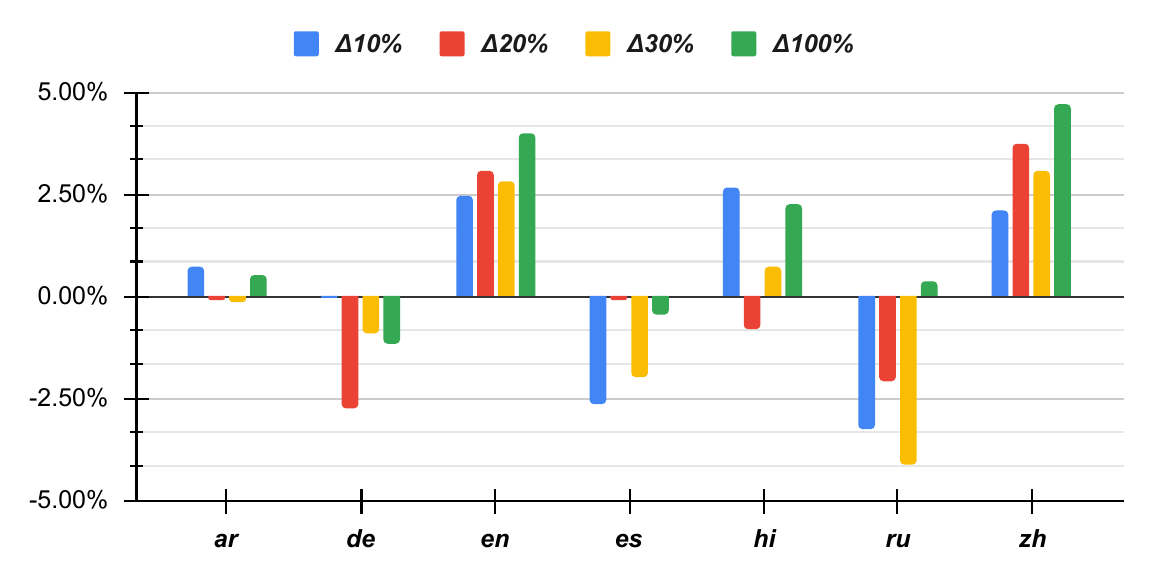}
\end{center}
\caption{Average$\Delta$-Toxicity scores for $P$-$FT$ \textit{vs} $M$-$FT$ for \texttt{mt5-large} over the \textit{neutral-test} evaluation set. \textbf{\textit{Takeaway}}: \textit{``en'', ``hi'', and ``zh'' showed significant updates.}}
\label{fig:delta-mt5-neutral-test}
\end{figure}


\begin{figure}[t!]
\begin{center}
\includegraphics[width=0.8\columnwidth]{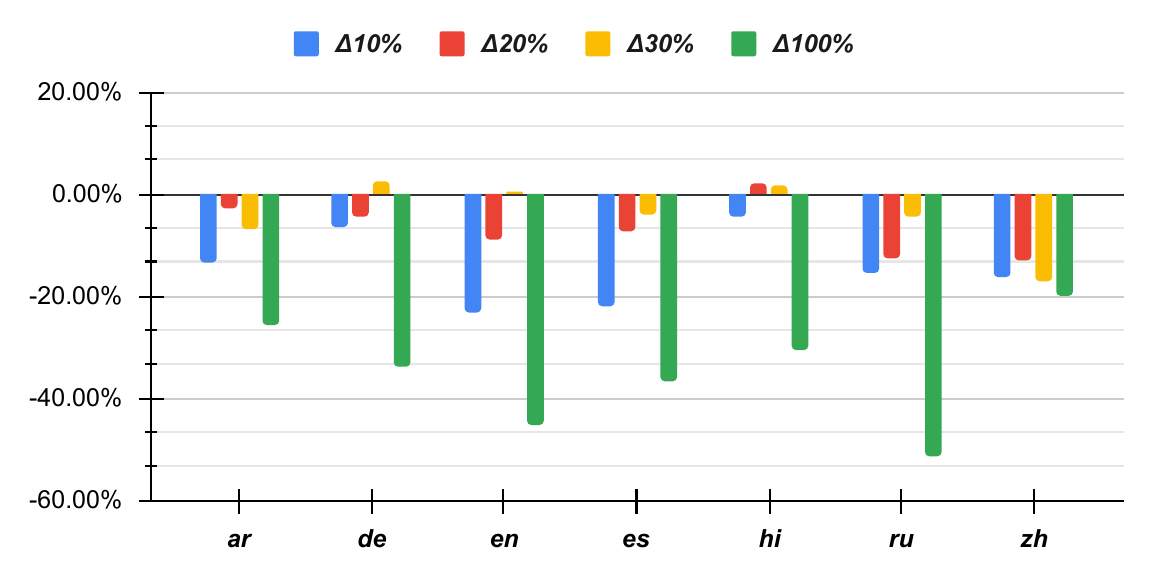}
\end{center}
\caption{Average$\Delta$-Toxicity scores for $P$-$FT$ \textit{vs} $M$-$FT$ for \texttt{bloom-7B1} over the \textit{toxic-train} evaluation set. \textbf{\textit{Takeaway}}: \textit{All the languages were adversely
affected.}}
\label{fig:delta-bloom-toxic-train}
\end{figure}

\begin{figure}[t!]
\begin{center}
\includegraphics[width=0.8\columnwidth]{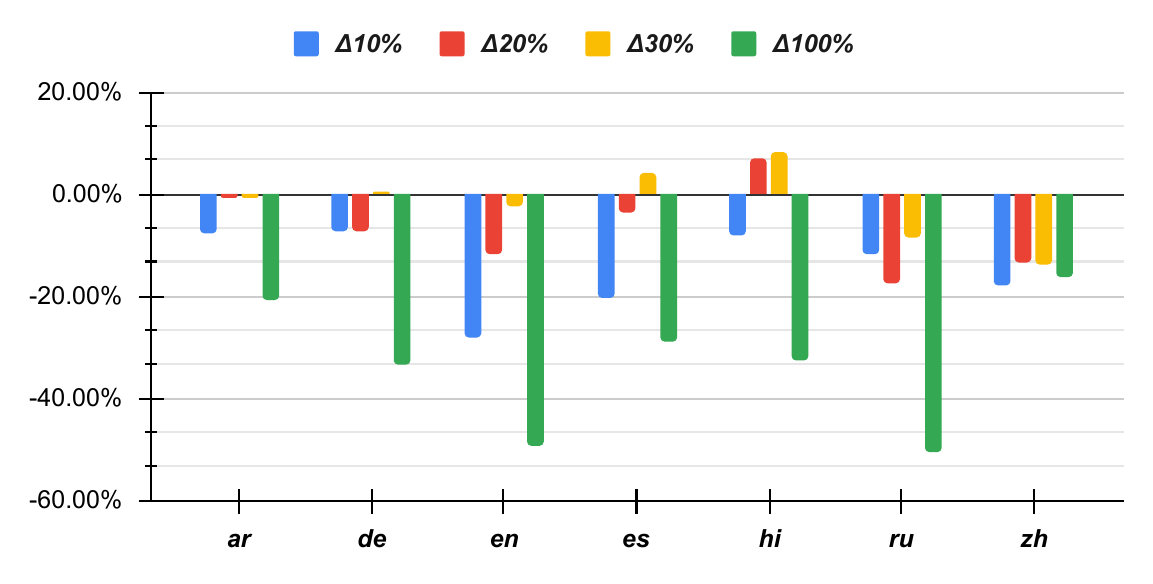}
\end{center}
\caption{Average$\Delta$-Toxicity scores for $P$-$FT$ \textit{vs} $M$-$FT$ for \texttt{bloom-7B1} over the \textit{toxic-test} evaluation set.  \textbf{\textit{Takeaway}}: \textit{All the languages were adversely
affected.}}
\label{fig:delta-bloom-toxic-test}
\end{figure}

\begin{figure}[t!]
\begin{center}
\includegraphics[width=0.8\columnwidth]{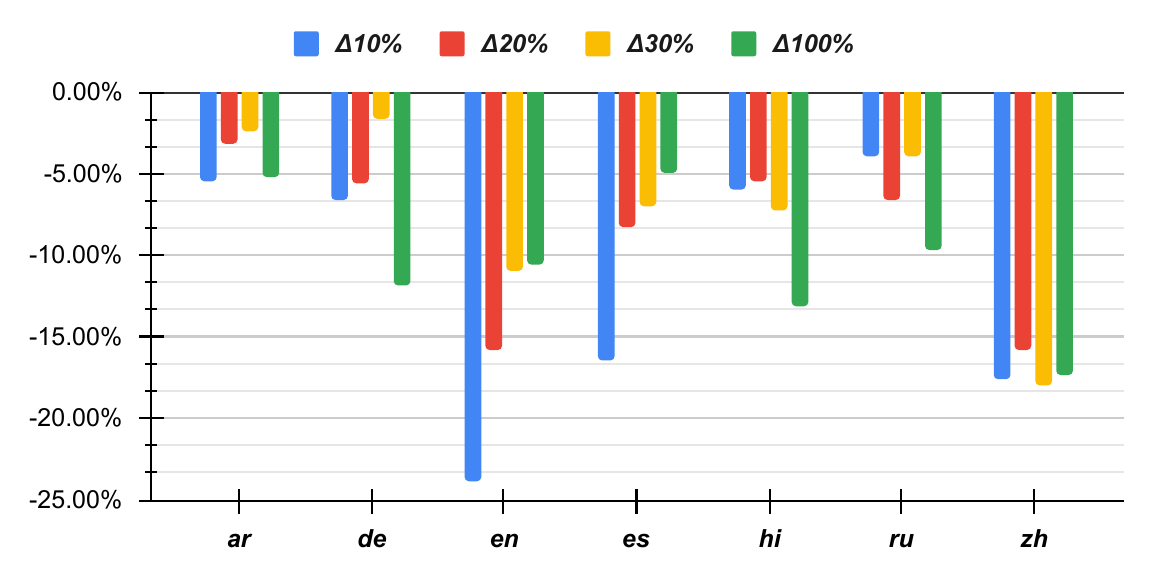}
\end{center}
\caption{Average$\Delta$-Toxicity scores for $P$-$FT$ \textit{vs} $M$-$FT$ for \texttt{bloom-7B1} over the \textit{neutral-test} evaluation set. \textbf{\textit{Takeaway}}: \textit{All the languages were adversely
affected.}}
\label{fig:delta-bloom-neutral-test}
\end{figure}

\begin{table*}[t]
\centering \resizebox{0.85\textwidth}{!}{%
\begin{tabular}{cccccccccc}
\toprule
\multicolumn{1}{l}{} & \multicolumn{1}{l}{} & \textbf{am} & \textbf{ar} & \textbf{de} & \textbf{en} & \textbf{es} & \textbf{hi} & \textbf{ru} & \textbf{AVG} \\ \midrule
\multicolumn{1}{l}{} & \textbf{$ZS$} & \cellcolor[HTML]{FEF7F7}13.72 & \cellcolor[HTML]{EA9189}78.92 & \cellcolor[HTML]{FCEBEA}21.53 & \cellcolor[HTML]{E67C73}91.79 & \cellcolor[HTML]{FFFFFF}\textbf{08.51} & \cellcolor[HTML]{FFFEFD}09.75 & \cellcolor[HTML]{FAE5E3}25.39 & \cellcolor[HTML]{E67C73}35.66 \\\hline\hline
 & \textbf{ar} & \cellcolor[HTML]{57BB8A}\textbf{03.54} & \cellcolor[HTML]{95D4B5}15.29 & \cellcolor[HTML]{59BC8B}-03.18 & \cellcolor[HTML]{ACDEC5}20.21 & \cellcolor[HTML]{57BB8A}\textbf{02.13} & \cellcolor[HTML]{57BB8A}\textbf{01.31} & \cellcolor[HTML]{5BBD8D}09.09 & \cellcolor[HTML]{65C194}6.91 \\
 & \textbf{de} & \cellcolor[HTML]{64C194}-03.70 & \cellcolor[HTML]{57BB8A}\textbf{49.98} & \cellcolor[HTML]{60BF91}-07.90 & \cellcolor[HTML]{57BB8A}\textbf{74.08} & \cellcolor[HTML]{67C295}-06.28 & \cellcolor[HTML]{64C193}-09.80 & \cellcolor[HTML]{5DBE8E}07.83 & \cellcolor[HTML]{57BB8A}\textbf{14.89} \\
 & \textbf{en} & \cellcolor[HTML]{62C092}-02.45 & \cellcolor[HTML]{B0DFC8}00.06 & \cellcolor[HTML]{7ECBA5}-25.74 & \cellcolor[HTML]{D1EDDF}-03.47 & \cellcolor[HTML]{70C59B}-11.52 & \cellcolor[HTML]{68C296}-12.80 & \cellcolor[HTML]{62C092}04.73 & \cellcolor[HTML]{7DCAA4}-7.31 \\
 & \textbf{es} & \cellcolor[HTML]{FFFFFF}-90.50 & \cellcolor[HTML]{FFFFFF}-45.04 & \cellcolor[HTML]{FFFFFF}-104.81 & \cellcolor[HTML]{FFFFFF}-32.81 & \cellcolor[HTML]{FFFFFF}-91.97 & \cellcolor[HTML]{FFFFFF}-145.96 & \cellcolor[HTML]{FFFFFF}-89.64 & \cellcolor[HTML]{FFFFFF}-85.82 \\
 & \textbf{hi} & \cellcolor[HTML]{5ABD8C}01.96 & \cellcolor[HTML]{B3E0CA}-01.72 & \cellcolor[HTML]{57BB8A}\textbf{-02.44} & \cellcolor[HTML]{F1FAF5}-23.61 & \cellcolor[HTML]{58BC8B}01.64 & \cellcolor[HTML]{59BC8C}00.01 & \cellcolor[HTML]{62C092}04.35 & \cellcolor[HTML]{75C79F}-2.83 \\
 & \textbf{ru} & \cellcolor[HTML]{5EBE8F}00.13 & \cellcolor[HTML]{B6E2CD}-03.67 & \cellcolor[HTML]{58BC8B}-02.55 & \cellcolor[HTML]{DCF1E7}-10.40 & \cellcolor[HTML]{5BBD8D}-00.03 & \cellcolor[HTML]{5ABC8C}-00.64 & \cellcolor[HTML]{67C295}01.59 & \cellcolor[HTML]{74C79E}-2.22 \\
 & \textbf{zh} & \cellcolor[HTML]{58BC8B}03.30 & \cellcolor[HTML]{A6DBC1}05.69 & \cellcolor[HTML]{5DBE8E}-05.55 & \cellcolor[HTML]{CEECDD}-01.61 & \cellcolor[HTML]{58BC8B}01.90 & \cellcolor[HTML]{58BC8B}\textbf{00.70} & \cellcolor[HTML]{57BB8A}\textbf{10.92} & \cellcolor[HTML]{6DC499}2.19 \\
\multirow{-8}{*}{\begin{sideways}\textbf{$X$-$FT$ ($\Delta$)}\end{sideways}} & \textbf{AVG} & \cellcolor[HTML]{74C79E}-12.53 & \cellcolor[HTML]{ABDDC4}2.94 & \cellcolor[HTML]{77C8A1}-21.74 & \cellcolor[HTML]{C7E9D8}\textbf{3.20} & \cellcolor[HTML]{76C8A0}-14.88 & \cellcolor[HTML]{74C79F}-23.88 & \cellcolor[HTML]{76C8A0}-7.30 & \multicolumn{1}{l}{} \\ \bottomrule
\end{tabular}%
}
\caption{Actual perplexity scores for Zero-Shot ($ZS$) \textit{vs} $\Delta$-perplexity scores for Cross-lingual Fine-Tuning ($X$-$FT$) for \texttt{aya-expanse-8B} over the \textit{toxic-train} evaluation set. $x$ represents the languages the model is trained on, while the languages on columns show the languages on which it is evaluated. $AP_Z$ and $\Delta_{AVG}$ represent the average perplexity in $ZS$ and average $\Delta$-perplexity scores for $X$-$FT$. \textbf{Bold} represents the best scores. \textbf{\textit{Takeaway}}: \textit{``hi'' and ``ru'' was most affected irrespective of fine-tuning language}.}
\label{fig:perp-aya-8-toxic-train}
\end{table*}

\begin{table*}[t]
\centering \resizebox{0.85\textwidth}{!}{%
\begin{tabular}{cccccccccc}
\toprule
\multicolumn{1}{l}{} & \multicolumn{1}{l}{} & \textbf{am} & \textbf{ar} & \textbf{de} & \textbf{en} & \textbf{es} & \textbf{hi} & \textbf{ru} & \textbf{AVG} \\ \midrule
\multicolumn{1}{l}{} & \textbf{$ZS$} & \cellcolor[HTML]{FEFAF9}12.92 & \cellcolor[HTML]{EDA09A}73.57 & \cellcolor[HTML]{FCEBE9}23.10 & \cellcolor[HTML]{E67C73}97.75 & \cellcolor[HTML]{FFFFFF}\textbf{08.92} & \cellcolor[HTML]{FFFEFE}10.01 & \cellcolor[HTML]{FBE8E7}24.59 & \cellcolor[HTML]{E67C73}35.84 \\\hline\hline
 & \textbf{ar} & \cellcolor[HTML]{58BC8B}03.16 & \cellcolor[HTML]{9AD6B9}06.00 & \cellcolor[HTML]{57BB8A}\textbf{01.95} & \cellcolor[HTML]{9DD8BB}30.50 & \cellcolor[HTML]{57BB8A}\textbf{02.21} & \cellcolor[HTML]{57BB8A}\textbf{01.45} & \cellcolor[HTML]{5DBE8F}07.34 & \cellcolor[HTML]{67C295}7.51 \\
 & \textbf{de} & \cellcolor[HTML]{62C092}-02.35 & \cellcolor[HTML]{57BB8A}\textbf{47.01} & \cellcolor[HTML]{62C092}-03.17 & \cellcolor[HTML]{57BB8A}\textbf{79.20} & \cellcolor[HTML]{6EC49A}-09.69 & \cellcolor[HTML]{64C193}-08.09 & \cellcolor[HTML]{57BB8A}\textbf{10.49} & \cellcolor[HTML]{57BB8A}\textbf{16.20} \\
 & \textbf{en} & \cellcolor[HTML]{5FBF90}-00.89 & \cellcolor[HTML]{B1E0C9}-07.90 & \cellcolor[HTML]{82CDA8}-19.47 & \cellcolor[HTML]{B9E3CE}11.33 & \cellcolor[HTML]{7AC9A2}-16.04 & \cellcolor[HTML]{7CCAA4}-25.89 & \cellcolor[HTML]{64C093}04.10 & \cellcolor[HTML]{82CDA8}-7.82 \\
 & \textbf{es} & \cellcolor[HTML]{FFFFFF}-90.23 & \cellcolor[HTML]{FFFFFF}-56.42 & \cellcolor[HTML]{FFFFFF}-81.76 & \cellcolor[HTML]{FFFFFF}-37.73 & \cellcolor[HTML]{FFFFFF}-87.72 & \cellcolor[HTML]{FFFFFF}-123.65 & \cellcolor[HTML]{FFFFFF}-78.28 & \cellcolor[HTML]{FFFFFF}-79.40 \\
 & \textbf{hi} & \cellcolor[HTML]{5DBE8E}00.56 & \cellcolor[HTML]{CDEBDC}-25.17 & \cellcolor[HTML]{5FBF90}-01.85 & \cellcolor[HTML]{E3F4EC}-18.14 & \cellcolor[HTML]{58BC8B}01.81 & \cellcolor[HTML]{58BC8B}00.80 & \cellcolor[HTML]{68C296}01.90 & \cellcolor[HTML]{7ECBA5}-5.73 \\
\bottomrule
\end{tabular}%
}
\caption{Actual perplexity scores for $ZS$ \textit{vs} $\Delta$-perplexity scores for $X$-$FT$ for \texttt{aya-expanse-8B} over the \textit{toxic-test} evaluation set. \textbf{\textit{Takeaway}}: \textit{``hi'' and ``ru'' was most affected irrespective of fine-tuning languages}.}
\label{fig:perp-aya-8-toxic-test}
\end{table*}

\begin{table*}[t]
\centering \resizebox{0.85\textwidth}{!}{%
\begin{tabular}{cccccccccc}
\toprule
\multicolumn{1}{l}{} & \multicolumn{1}{l}{} & \textbf{am} & \textbf{ar} & \textbf{de} & \textbf{en} & \textbf{es} & \textbf{hi} & \textbf{ru} & \textbf{AVG} \\ \midrule
\multicolumn{1}{l}{} & \textbf{$ZS$} & \cellcolor[HTML]{FEF6F5}14.42 & \cellcolor[HTML]{E67C73}86.83 & \cellcolor[HTML]{FDF0EF}17.74 & \cellcolor[HTML]{E8877E}80.80 & \cellcolor[HTML]{FFFFFF}09.25 & \cellcolor[HTML]{FFFFFF}\textbf{08.76} & \cellcolor[HTML]{FAE0DE}27.27 & \cellcolor[HTML]{E67C73}35.01 \\\hline\hline
 & \textbf{ar} & \cellcolor[HTML]{58BC8B}03.82 & \cellcolor[HTML]{8AD0AE}20.04 & \cellcolor[HTML]{5FBE90}-05.17 & \cellcolor[HTML]{A0D9BD}04.98 & \cellcolor[HTML]{59BC8C}01.90 & \cellcolor[HTML]{57BB8A}\textbf{00.70} & \cellcolor[HTML]{59BC8C}11.55 & \cellcolor[HTML]{61BF91}5.40 \\
 & \textbf{de} & \cellcolor[HTML]{63C092}-03.98 & \cellcolor[HTML]{57BB8A}\textbf{55.67} & \cellcolor[HTML]{6AC397}-11.19 & \cellcolor[HTML]{57BB8A}\textbf{55.02} & \cellcolor[HTML]{76C8A0}-10.89 & \cellcolor[HTML]{73C69D}-16.92 & \cellcolor[HTML]{5CBD8E}09.43 & \cellcolor[HTML]{57BB8A}\textbf{11.02} \\
 & \textbf{en} & \cellcolor[HTML]{5FBF90}-01.50 & \cellcolor[HTML]{96D5B6}11.32 & \cellcolor[HTML]{82CDA8}-23.79 & \cellcolor[HTML]{BFE6D3}-16.35 & \cellcolor[HTML]{6CC499}-06.56 & \cellcolor[HTML]{68C296}-10.23 & \cellcolor[HTML]{63C093}04.77 & \cellcolor[HTML]{74C79F}-6.05 \\
 & \textbf{es} & \cellcolor[HTML]{FFFFFF}-123.02 & \cellcolor[HTML]{FFFFFF}-62.81 & \cellcolor[HTML]{FFFFFF}-89.56 & \cellcolor[HTML]{FFFFFF}-60.30 & \cellcolor[HTML]{FFFFFF}-72.23 & \cellcolor[HTML]{FFFFFF}-108.47 & \cellcolor[HTML]{FFFFFF}-99.29 & \cellcolor[HTML]{FFFFFF}-87.95 \\
 & \textbf{hi} & \cellcolor[HTML]{5BBD8D}01.65 & \cellcolor[HTML]{98D5B7}10.42 & \cellcolor[HTML]{63C093}-07.51 & \cellcolor[HTML]{BEE5D2}-15.38 & \cellcolor[HTML]{59BC8B}02.19 & \cellcolor[HTML]{59BC8C}-00.51 & \cellcolor[HTML]{60BF90}07.25 & \cellcolor[HTML]{6BC398}-0.27 \\
\bottomrule
\end{tabular}%
}
\caption{Actual perplexity scores for $ZS$ \textit{vs} $\Delta$-perplexity scores for $X$-$FT$ for \texttt{aya-expanse-8B} over the \textit{neutral-test} evaluation set. \textbf{\textit{Takeaway}}: \textit{Detoxification adversely effects the model’s general knowledge}.}
\label{fig:perp-aya-8-neutral-test}
\end{table*}

\begin{table*}[t]
\centering \resizebox{0.85\textwidth}{!}{%
\begin{tabular}{cccccccccc}
\toprule
\multicolumn{1}{l}{} & \multicolumn{1}{l}{} & \textbf{am} & \textbf{ar} & \textbf{de} & \textbf{en} & \textbf{es} & \textbf{hi} & \textbf{ru} & \textbf{AVG} \\ \midrule
\multicolumn{1}{l}{} & \textbf{$ZS$} & \cellcolor[HTML]{FEF7F6}11.13 & \cellcolor[HTML]{E98981}65.72 & \cellcolor[HTML]{FBEAE9}17.33 & \cellcolor[HTML]{E67C73}72.11 & \cellcolor[HTML]{FFFFFF}\textbf{06.64} & \cellcolor[HTML]{FFFCFB}08.59 & \cellcolor[HTML]{FBE9E8}17.83 & \cellcolor[HTML]{E67C73}28.48 \\\hline\hline
 & \textbf{ar} & \cellcolor[HTML]{66C194}-00.41 & \cellcolor[HTML]{57BB8A}\textbf{24.04} & \cellcolor[HTML]{6BC498}-01.53 & \cellcolor[HTML]{BBE4D0}10.05 & \cellcolor[HTML]{57BB8A}\textbf{00.85} & \cellcolor[HTML]{5ABC8C}00.27 & \cellcolor[HTML]{6AC397}04.11 & \cellcolor[HTML]{57BB8A}\textbf{5.34} \\
 & \textbf{de} & \cellcolor[HTML]{7CCAA4}-06.78 & \cellcolor[HTML]{63C093}22.25 & \cellcolor[HTML]{A0D9BD}-15.79 & \cellcolor[HTML]{57BB8A}\textbf{49.94} & \cellcolor[HTML]{78C9A1}-10.35 & \cellcolor[HTML]{82CDA8}-11.09 & \cellcolor[HTML]{57BB8A}\textbf{06.87} & \cellcolor[HTML]{59BC8C}5.01 \\
 & \textbf{en} & \cellcolor[HTML]{6BC398}-01.83 & \cellcolor[HTML]{B0DFC8}10.25 & \cellcolor[HTML]{D0ECDE}-28.82 & \cellcolor[HTML]{FFFFFF}-17.17 & \cellcolor[HTML]{74C79E}-08.79 & \cellcolor[HTML]{61BF91}-01.69 & \cellcolor[HTML]{BBE4D0}-08.06 & \cellcolor[HTML]{9ED8BC}-8.02 \\
 & \textbf{es} & \cellcolor[HTML]{FFFFFF}-43.57 & \cellcolor[HTML]{7BCAA3}18.54 & \cellcolor[HTML]{FFFFFF}-41.71 & \cellcolor[HTML]{CAEADA}04.26 & \cellcolor[HTML]{FFFFFF}-56.92 & \cellcolor[HTML]{FFFFFF}-46.59 & \cellcolor[HTML]{FFFFFF}-18.38 & \cellcolor[HTML]{FFFFFF}-26.34 \\
 & \textbf{hi} & \cellcolor[HTML]{5CBD8E}02.23 & \cellcolor[HTML]{BBE4D0}08.65 & \cellcolor[HTML]{57BB8A}\textbf{03.87} & \cellcolor[HTML]{E4F4ED}-06.27 & \cellcolor[HTML]{58BC8B}00.60 & \cellcolor[HTML]{58BC8B}00.70 & \cellcolor[HTML]{66C295}04.64 & \cellcolor[HTML]{69C397}2.06 \\
\bottomrule
\end{tabular}%
}
\caption{Actual perplexity scores for $ZS$ \textit{vs} $\Delta$-perplexity scores for $X$-$FT$ for \texttt{aya-23-8B} over the \textit{toxic-train} evaluation set. \textbf{\textit{Takeaway}}: \textit{``es'' turned out to be least affected by other fine-tuning languages}.}
\label{fig:perp-aya-23-toxic-train}
\end{table*}

\begin{table*}[t]
\centering \resizebox{0.85\textwidth}{!}{%
\begin{tabular}{cccccccccc}
\toprule
\multicolumn{1}{l}{} & \multicolumn{1}{l}{} & \textbf{am} & \textbf{ar} & \textbf{de} & \textbf{en} & \textbf{es} & \textbf{hi} & \textbf{ru} & \textbf{AVG} \\ \midrule
\multicolumn{1}{l}{} & \textbf{$ZS$} & \cellcolor[HTML]{FDF4F3}12.08 & \cellcolor[HTML]{E77D74}68.14 & \cellcolor[HTML]{FBEAE9}16.55 & \cellcolor[HTML]{E67C73}68.30 & \cellcolor[HTML]{FFFFFF}\textbf{06.58} & \cellcolor[HTML]{FFFCFC}08.28 & \cellcolor[HTML]{FBEAE8}16.87 & \cellcolor[HTML]{E67C73}28.11 \\\hline\hline
 & \textbf{ar} & \cellcolor[HTML]{85CEAA}-09.46 & \cellcolor[HTML]{ABDDC4}14.21 & \cellcolor[HTML]{64C193}-00.76 & \cellcolor[HTML]{BDE5D1}06.04 & \cellcolor[HTML]{57BB8A}\textbf{01.31} & \cellcolor[HTML]{5BBD8D}00.46 & \cellcolor[HTML]{5CBD8E}02.61 & \cellcolor[HTML]{58BC8B}2.06 \\
 & \textbf{de} & \cellcolor[HTML]{6EC59A}-02.24 & \cellcolor[HTML]{67C295}24.49 & \cellcolor[HTML]{B2E0C9}-25.33 & \cellcolor[HTML]{57BB8A}\textbf{50.29} & \cellcolor[HTML]{7FCBA6}-11.82 & \cellcolor[HTML]{AADDC4}-21.44 & \cellcolor[HTML]{61BF91}01.12 & \cellcolor[HTML]{57BB8A}\textbf{2.15} \\
 & \textbf{en} & \cellcolor[HTML]{6EC59A}-02.20 & \cellcolor[HTML]{9FD8BC}15.99 & \cellcolor[HTML]{B8E3CE}-27.50 & \cellcolor[HTML]{DDF2E7}-07.68 & \cellcolor[HTML]{78C9A1}-09.48 & \cellcolor[HTML]{5ABD8C}00.68 & \cellcolor[HTML]{FFFFFF}-50.35 & \cellcolor[HTML]{A4DAC0}-11.51 \\
 & \textbf{es} & \cellcolor[HTML]{FFFFFF}-47.85 & \cellcolor[HTML]{57BB8A}\textbf{26.86} & \cellcolor[HTML]{FFFFFF}-49.98 & \cellcolor[HTML]{C1E6D4}04.28 & \cellcolor[HTML]{FFFFFF}-54.92 & \cellcolor[HTML]{FFFFFF}-45.01 & \cellcolor[HTML]{BEE5D2}-29.21 & \cellcolor[HTML]{FFFFFF}-27.98 \\
 & \textbf{hi} & \cellcolor[HTML]{5EBE8F}02.89 & \cellcolor[HTML]{C6E8D7}10.06 & \cellcolor[HTML]{57BB8A}\textbf{03.17} & \cellcolor[HTML]{E2F4EB}-09.85 & \cellcolor[HTML]{5BBD8D}00.25 & \cellcolor[HTML]{57BB8A}\textbf{01.44} & \cellcolor[HTML]{57BB8A}\textbf{04.08} & \cellcolor[HTML]{5ABC8C}1.72 \\
\bottomrule
\end{tabular}%
}
\caption{Actual perplexity scores for $ZS$ \textit{vs} $\Delta$-perplexity scores for $X$-$FT$ for \texttt{aya-23-8B} over the \textit{toxic-test} evaluation set. \textbf{\textit{Takeaway}}: \textit{``de'' turned out to be least affected by other fine-tuning languages}.}
\label{fig:perp-aya-23-toxic-test}
\end{table*}

\begin{table*}[t]
\centering \resizebox{0.85\textwidth}{!}{%
\begin{tabular}{cccccccccc}
\toprule
\multicolumn{1}{l}{} & \multicolumn{1}{l}{} & \textbf{am} & \textbf{ar} & \textbf{de} & \textbf{en} & \textbf{es} & \textbf{hi} & \textbf{ru} & \textbf{AVG} \\ \midrule
\multicolumn{1}{l}{} & \textbf{$ZS$} & \cellcolor[HTML]{FEF7F7}11.99 & \cellcolor[HTML]{EA9189}58.39 & \cellcolor[HTML]{FCEEED}15.98 & \cellcolor[HTML]{E67C73}67.60 & \cellcolor[HTML]{FFFFFF}\textbf{08.28} & \cellcolor[HTML]{FFFFFF}08.33 & \cellcolor[HTML]{FDF1F0}14.71 & \cellcolor[HTML]{E67C73}26.47 \\\hline\hline
 & \textbf{ar} & \cellcolor[HTML]{5FBF90}01.58 & \cellcolor[HTML]{57BB8A}\textbf{18.62} & \cellcolor[HTML]{71C69C}-05.83 & \cellcolor[HTML]{D4EEE1}-01.41 & \cellcolor[HTML]{58BC8B}02.46 & \cellcolor[HTML]{5ABD8D}00.31 & \cellcolor[HTML]{57BB8A}\textbf{02.41} & \cellcolor[HTML]{57BB8A}\textbf{2.59} \\
 & \textbf{de} & \cellcolor[HTML]{7ECBA5}-09.37 & \cellcolor[HTML]{86CEAB}11.97 & \cellcolor[HTML]{A5DBC0}-20.57 & \cellcolor[HTML]{57BB8A}\textbf{42.05} & \cellcolor[HTML]{7AC9A2}-11.18 & \cellcolor[HTML]{C4E8D6}-27.10 & \cellcolor[HTML]{61BF91}01.12 & \cellcolor[HTML]{6CC499}-1.87 \\
 & \textbf{en} & \cellcolor[HTML]{66C195}-00.73 & \cellcolor[HTML]{FFFFFF}-05.29 & \cellcolor[HTML]{DEF2E8}-36.57 & \cellcolor[HTML]{D8F0E4}-02.88 & \cellcolor[HTML]{9ED8BB}-25.59 & \cellcolor[HTML]{59BC8C}00.66 & \cellcolor[HTML]{91D3B3}-05.16 & \cellcolor[HTML]{95D5B6}-10.79 \\
 & \textbf{es} & \cellcolor[HTML]{FFFFFF}-55.74 & \cellcolor[HTML]{B1E0C9}05.86 & \cellcolor[HTML]{FFFFFF}-46.06 & \cellcolor[HTML]{F6FCF9}-13.17 & \cellcolor[HTML]{FFFFFF}-64.98 & \cellcolor[HTML]{FFFFFF}-42.36 & \cellcolor[HTML]{FFFFFF}-19.74 & \cellcolor[HTML]{FFFFFF}-33.74 \\
 & \textbf{hi} & \cellcolor[HTML]{5CBE8E}02.65 & \cellcolor[HTML]{EAF7F0}-02.18 & \cellcolor[HTML]{69C397}-03.69 & \cellcolor[HTML]{EEF9F3}-10.54 & \cellcolor[HTML]{59BC8C}01.80 & \cellcolor[HTML]{58BC8B}00.94 & \cellcolor[HTML]{69C397}00.11 & \cellcolor[HTML]{6BC398}-1.56 \\
\bottomrule
\end{tabular}%
}
\caption{Actual perplexity scores for $ZS$ \textit{vs} $\Delta$-perplexity scores for $X$-$FT$ for \texttt{aya-23-8B} over the \textit{neutral-test} evaluation set. \textbf{\textit{Takeaway}}: \textit{Detoxification adversely effects the model’s general knowledge}.}
\label{fig:perp-aya-23-neutral-test}
\end{table*}

\begin{table*}[t]
\centering \resizebox{0.85\textwidth}{!}{%
\begin{tabular}{cccccccccc}
\toprule
\multicolumn{1}{l}{} & \multicolumn{1}{l}{} & \textbf{am} & \textbf{ar} & \textbf{de} & \textbf{en} & \textbf{es} & \textbf{hi} & \textbf{ru} & \textbf{AVG} \\ \midrule
\multicolumn{1}{l}{} & \textbf{$ZS$} & \cellcolor[HTML]{FFFFFF}20.83 & \cellcolor[HTML]{E67C73}418.68 & \cellcolor[HTML]{FAE4E3}102.44 & \cellcolor[HTML]{F7D1CE}160.72 & \cellcolor[HTML]{FEFAF9}37.47 & \cellcolor[HTML]{FEF9F9}39.66 & \cellcolor[HTML]{FFFFFF}\textbf{20.30} & \cellcolor[HTML]{E67C73}114.30 \\\hline\hline
 & \textbf{ar} & \cellcolor[HTML]{5BBD8D}-41.82 & \cellcolor[HTML]{86CEAB}290.83 & \cellcolor[HTML]{9CD7BA}-33.35 & \cellcolor[HTML]{57BB8A}\textbf{13.65} & \cellcolor[HTML]{FFFFFF}-41.03 & \cellcolor[HTML]{68C296}-15.91 & \cellcolor[HTML]{B4E1CB}-50.45 & \cellcolor[HTML]{6FC59B}17.42 \\
 & \textbf{de} & \cellcolor[HTML]{82CDA8}-48.42 & \cellcolor[HTML]{7DCAA4}293.50 & \cellcolor[HTML]{57BB8A}\textbf{-15.34} & \cellcolor[HTML]{77C8A0}08.04 & \cellcolor[HTML]{57BB8A}\textbf{-20.01} & \cellcolor[HTML]{FFFFFF}-36.17 & \cellcolor[HTML]{ADDEC6}-49.62 & \cellcolor[HTML]{57BB8A}\textbf{18.85} \\
 & \textbf{en} & \cellcolor[HTML]{6DC49A}-44.97 & \cellcolor[HTML]{FFFFFF}258.00 & \cellcolor[HTML]{B8E3CE}-40.79 & \cellcolor[HTML]{F8FCFA}-15.34 & \cellcolor[HTML]{67C295}-21.91 & \cellcolor[HTML]{7AC9A3}-18.27 & \cellcolor[HTML]{E6F5EE}-56.07 & \cellcolor[HTML]{FFFFFF}8.66 \\
 & \textbf{es} & \cellcolor[HTML]{C3E7D5}-59.50 & \cellcolor[HTML]{57BB8A}\textbf{303.54} & \cellcolor[HTML]{BAE3CF}-41.25 & \cellcolor[HTML]{C1E6D4}-05.44 & \cellcolor[HTML]{96D5B6}-27.81 & \cellcolor[HTML]{C6E8D8}-28.49 & \cellcolor[HTML]{FFFFFF}-59.03 & \cellcolor[HTML]{CDEBDC}11.72 \\
 & \textbf{hi} & \cellcolor[HTML]{57BB8A}\textbf{-41.29} & \cellcolor[HTML]{6EC59A}297.44 & \cellcolor[HTML]{FFFFFF}-59.69 & \cellcolor[HTML]{79C9A2}07.66 & \cellcolor[HTML]{88CFAC}-26.10 & \cellcolor[HTML]{57BB8A}\textbf{-13.63} & \cellcolor[HTML]{F0F9F5}-57.27 & \cellcolor[HTML]{92D3B3}15.30 \\
\bottomrule
\end{tabular}%
}
\caption{
Actual perplexity scores for $ZS$ \textit{vs} $\Delta$-perplexity scores for $X$-$FT$ for \texttt{mt5-large} over the \textit{toxic-train} evaluation set. \textbf{\textit{Takeaway}}: \textit{``en'' turned out to be least affected by other fine-tuning languages}.}
\label{fig:perp-mt5-toxic-train}
\end{table*}

\begin{table*}[t]
\centering \resizebox{0.85\textwidth}{!}{%
\begin{tabular}{cccccccccc}
\toprule
\multicolumn{1}{l}{} & \multicolumn{1}{l}{} & \textbf{am} & \textbf{ar} & \textbf{de} & \textbf{en} & \textbf{es} & \textbf{hi} & \textbf{ru} & \textbf{AVG} \\ \midrule
\multicolumn{1}{l}{} & \textbf{$ZS$} & \cellcolor[HTML]{FFFEFE}24.23 & \cellcolor[HTML]{E67C73}662.17 & \cellcolor[HTML]{FDF1F0}89.41 & \cellcolor[HTML]{FAE4E2}152.72 & \cellcolor[HTML]{FFFFFF}20.73 & \cellcolor[HTML]{FFFFFF}\textbf{16.72} & \cellcolor[HTML]{FFFEFE}25.23 & \cellcolor[HTML]{E67C73}141.60 \\\hline\hline
 & \textbf{ar} & \cellcolor[HTML]{88CFAC}-44.33 & \cellcolor[HTML]{A2DABE}534.77 & \cellcolor[HTML]{AADDC4}-39.18 & \cellcolor[HTML]{69C397}13.34 & \cellcolor[HTML]{8AD0AE}-66.87 & \cellcolor[HTML]{A9DDC3}-46.71 & \cellcolor[HTML]{F5FBF8}-35.97 & \cellcolor[HTML]{75C79F}45.01 \\
 & \textbf{de} & \cellcolor[HTML]{BAE3CF}-53.80 & \cellcolor[HTML]{96D5B6}541.97 & \cellcolor[HTML]{FEFFFE}-58.83 & \cellcolor[HTML]{EFF9F4}-41.43 & \cellcolor[HTML]{82CDA8}-63.42 & \cellcolor[HTML]{F2FAF6}-66.64 & \cellcolor[HTML]{F0F9F5}-35.50 & \cellcolor[HTML]{FFFFFF}31.77 \\
 & \textbf{en} & \cellcolor[HTML]{83CDA8}-43.34 & \cellcolor[HTML]{74C79E}563.65 & \cellcolor[HTML]{65C194}-22.98 & \cellcolor[HTML]{ADDEC6}-14.47 & \cellcolor[HTML]{7DCBA4}-61.43 & \cellcolor[HTML]{FFFFFF}-70.40 & \cellcolor[HTML]{D8F0E4}-33.34 & \cellcolor[HTML]{71C69C}45.38 \\
 & \textbf{es} & \cellcolor[HTML]{FFFFFF}-67.00 & \cellcolor[HTML]{84CEA9}553.54 & \cellcolor[HTML]{ACDEC5}-39.59 & \cellcolor[HTML]{57BB8A}\textbf{20.49} & \cellcolor[HTML]{FFFFFF}-113.59 & \cellcolor[HTML]{57BB8A}\textbf{-24.31} & \cellcolor[HTML]{57BB8A}\textbf{-21.70} & \cellcolor[HTML]{80CCA6}43.98 \\
 & \textbf{hi} & \cellcolor[HTML]{75C89F}-40.83 & \cellcolor[HTML]{57BB8A}\textbf{581.54} & \cellcolor[HTML]{B2E0CA}-41.06 & \cellcolor[HTML]{D5EEE2}-30.84 & \cellcolor[HTML]{74C79E}-57.98 & \cellcolor[HTML]{9ED8BC}-43.64 & \cellcolor[HTML]{D0ECDF}-32.63 & \cellcolor[HTML]{57BB8A}\textbf{47.80} \\
\bottomrule
\end{tabular}%
}
\caption{Actual perplexity scores for $ZS$ \textit{vs} $\Delta$-perplexity scores for $X$-$FT$ for \texttt{mt5-large} over the \textit{toxic-test} evaluation set. \textbf{\textit{Takeaway}}: \textit{``hi'' and ``ru'' was most affected irrespective of fine-tuning languages}.}
\label{fig:perp-mt5-toxic-test}
\end{table*}

\begin{table*}[t]
\centering \resizebox{0.85\textwidth}{!}{%
\begin{tabular}{cccccccccc}
\toprule
\multicolumn{1}{l}{} & \multicolumn{1}{l}{} & \textbf{am} & \textbf{ar} & \textbf{de} & \textbf{en} & \textbf{es} & \textbf{hi} & \textbf{ru} & \textbf{AVG} \\ \midrule
\multicolumn{1}{l}{} & \textbf{$ZS$} & \cellcolor[HTML]{FFFDFC}17.79 & \cellcolor[HTML]{E67C73}195.62 & \cellcolor[HTML]{F8D7D4}69.68 & \cellcolor[HTML]{EEA39D}142.01 & \cellcolor[HTML]{FFFFFF}\textbf{13.79} & \cellcolor[HTML]{FFFDFD}17.05 & \cellcolor[HTML]{FDF2F1}32.32 & \cellcolor[HTML]{E67C73}69.75 \\\hline\hline
 & \textbf{ar} & \cellcolor[HTML]{73C69D}-40.49 & \cellcolor[HTML]{57BB8A}\textbf{92.62} & \cellcolor[HTML]{EFF9F4}-91.01 & \cellcolor[HTML]{FFFFFF}-47.79 & \cellcolor[HTML]{CEECDD}-85.73 & \cellcolor[HTML]{FFFFFF}-73.79 & \cellcolor[HTML]{83CDA9}-33.37 & \cellcolor[HTML]{F2FAF6}-39.94 \\
 & \textbf{de} & \cellcolor[HTML]{8DD1B0}-48.82 & \cellcolor[HTML]{AFDFC8}60.18 & \cellcolor[HTML]{DFF2E9}-84.11 & \cellcolor[HTML]{D6EFE3}-27.92 & \cellcolor[HTML]{BEE5D2}-79.20 & \cellcolor[HTML]{C8E9D9}-58.86 & \cellcolor[HTML]{FFFFFF}-55.94 & \cellcolor[HTML]{FFFFFF}-42.10 \\
 & \textbf{en} & \cellcolor[HTML]{57BB8A}\textbf{-31.83} & \cellcolor[HTML]{C6E8D8}51.71 & \cellcolor[HTML]{D6EFE2}-80.42 & \cellcolor[HTML]{D0ECDF}-25.18 & \cellcolor[HTML]{83CDA9}-55.99 & \cellcolor[HTML]{BDE5D1}-55.87 & \cellcolor[HTML]{E3F4EC}-50.75 & \cellcolor[HTML]{D5EEE2}-35.47 \\
 & \textbf{es} & \cellcolor[HTML]{95D4B5}-51.49 & \cellcolor[HTML]{A5DBC1}63.91 & \cellcolor[HTML]{D4EEE1}-79.88 & \cellcolor[HTML]{73C79E}19.69 & \cellcolor[HTML]{6CC499}-46.90 & \cellcolor[HTML]{FCFEFD}-72.75 & \cellcolor[HTML]{57BB8A}\textbf{-25.39} & \cellcolor[HTML]{A1D9BE}-27.54 \\
 & \textbf{hi} & \cellcolor[HTML]{D9F0E5}-73.29 & \cellcolor[HTML]{FFFFFF}30.61 & \cellcolor[HTML]{AEDEC7}-64.19 & \cellcolor[HTML]{78C9A1}17.28 & \cellcolor[HTML]{FFFFFF}-105.15 & \cellcolor[HTML]{5FBF90}-30.51 & \cellcolor[HTML]{F1FAF6}-53.38 & \cellcolor[HTML]{F1F9F5}-39.80 \\
\bottomrule
\end{tabular}%
}
\caption{Actual perplexity scores for $ZS$ \textit{vs} $\Delta$-perplexity scores for $X$-$FT$ for \texttt{mt5-large} over the \textit{neutral-test} evaluation set. \textbf{\textit{Takeaway}}: \textit{Detoxification adversely effects the model’s general knowledge}.}
\label{fig:perp-mt5-neutral-test}
\end{table*}

\begin{table*}[t]

\centering \resizebox{0.85\textwidth}{!}{%
\begin{tabular}{cccccccccc}
\toprule
\multicolumn{1}{l}{} & \multicolumn{1}{l}{} & \textbf{am} & \textbf{ar} & \textbf{de} & \textbf{en} & \textbf{es} & \textbf{hi} & \textbf{ru} & \textbf{AVG} \\ \midrule
\multicolumn{1}{l}{} & \textbf{$ZS$} & \cellcolor[HTML]{FFFFFF}06.69 & \cellcolor[HTML]{E67C73}2259.35 & \cellcolor[HTML]{FFFFFF}\textbf{04.01} & \cellcolor[HTML]{FFFFFF}16.44 & \cellcolor[HTML]{FFFFFF}12.06 & \cellcolor[HTML]{F9DFDD}564.88 & \cellcolor[HTML]{FFFFFF}04.82 & \cellcolor[HTML]{E67C73}409.75 \\\hline\hline
 & \textbf{ar} & \cellcolor[HTML]{5BBD8D}-04.23 & \cellcolor[HTML]{5DBE8E}2250.85 & \cellcolor[HTML]{5DBE8E}-08.09 & \cellcolor[HTML]{60BF91}08.15 & \cellcolor[HTML]{5BBD8D}-01.69 & \cellcolor[HTML]{5FBE8F}552.11 & \cellcolor[HTML]{5EBE8F}-03.24 & \cellcolor[HTML]{5DBE8E}399.12 \\
 & \textbf{de} & \cellcolor[HTML]{7CCAA4}-83.18 & \cellcolor[HTML]{FFFFFF}2116.93 & \cellcolor[HTML]{89D0AD}-76.53 & \cellcolor[HTML]{FFFFFF}-88.91 & \cellcolor[HTML]{E8F6EF}-382.00 & \cellcolor[HTML]{9BD7BA}479.28 & \cellcolor[HTML]{C2E6D4}-67.29 & \cellcolor[HTML]{BFE5D2}271.18 \\
 & \textbf{en} & \cellcolor[HTML]{58BC8B}01.22 & \cellcolor[HTML]{58BC8B}2255.43 & \cellcolor[HTML]{5BBD8D}-05.20 & \cellcolor[HTML]{5CBD8D}10.95 & \cellcolor[HTML]{59BC8B}04.74 & \cellcolor[HTML]{5ABC8C}557.96 & \cellcolor[HTML]{5CBD8E}-01.99 & \cellcolor[HTML]{59BC8C}403.30 \\
 & \textbf{es} & \cellcolor[HTML]{71C69C}-56.11 & \cellcolor[HTML]{81CCA8}2220.91 & \cellcolor[HTML]{5DBE8E}-08.01 & \cellcolor[HTML]{5BBD8D}11.14 & \cellcolor[HTML]{5FBE90}-11.81 & \cellcolor[HTML]{74C79E}526.42 & \cellcolor[HTML]{8DD1AF}-33.18 & \cellcolor[HTML]{6CC499}378.48 \\
 & \textbf{hi} & \cellcolor[HTML]{5ABD8D}-04.21 & \cellcolor[HTML]{5EBE8F}2249.98 & \cellcolor[HTML]{5ABC8C}-03.23 & \cellcolor[HTML]{57BB8A}\textbf{13.54} & \cellcolor[HTML]{57BB8A}\textbf{07.91} & \cellcolor[HTML]{58BC8B}559.85 & \cellcolor[HTML]{60BF90}-04.38 & \cellcolor[HTML]{5ABC8C}402.78 \\
\bottomrule
\end{tabular}%
}
\caption{Actual perplexity scores for $ZS$ \textit{vs} $\Delta$-perplexity scores for $X$-$FT$ for \texttt{bloom-7B1} over the \textit{toxic-train} evaluation set. \textbf{\textit{Takeaway}}: \textit{All the languages were adversely affected}.}
\label{fig:perp-bloom-toxic-train}
\end{table*}

\begin{table*}[t]
\centering \resizebox{0.85\textwidth}{!}{%
\begin{tabular}{cccccccccc}
\toprule
\multicolumn{1}{l}{} & \multicolumn{1}{l}{} & \textbf{am} & \textbf{ar} & \textbf{de} & \textbf{en} & \textbf{es} & \textbf{hi} & \textbf{ru} & \textbf{AVG} \\ \midrule
\multicolumn{1}{l}{} & \textbf{$ZS$} & \cellcolor[HTML]{FFFFFF}\textbf{23.57} & \cellcolor[HTML]{F8D7D4}114.45 & \cellcolor[HTML]{FEF6F6}44.15 & \cellcolor[HTML]{F5C8C5}145.82 & \cellcolor[HTML]{F4C2BE}159.41 & \cellcolor[HTML]{E67C73}314.27 & \cellcolor[HTML]{F1B6B1}187.03 & \cellcolor[HTML]{E67C73}141.24 \\\hline\hline
 & \textbf{ar} & \cellcolor[HTML]{62C092}13.30 & \cellcolor[HTML]{5BBD8D}104.03 & \cellcolor[HTML]{5DBE8E}34.39 & \cellcolor[HTML]{AFDFC7}135.05 & \cellcolor[HTML]{5EBE8F}152.16 & \cellcolor[HTML]{59BC8C}302.90 & \cellcolor[HTML]{79C9A2}177.97 & \cellcolor[HTML]{5DBE8E}131.40 \\
 & \textbf{de} & \cellcolor[HTML]{60BF90}14.90 & \cellcolor[HTML]{66C195}84.26 & \cellcolor[HTML]{76C8A0}06.48 & \cellcolor[HTML]{C5E8D7}133.17 & \cellcolor[HTML]{C3E7D5}75.15 & \cellcolor[HTML]{6EC49A}218.62 & \cellcolor[HTML]{FFFFFF}149.80 & \cellcolor[HTML]{7BCAA3}97.48 \\
 & \textbf{en} & \cellcolor[HTML]{5CBD8D}17.24 & \cellcolor[HTML]{59BC8C}108.49 & \cellcolor[HTML]{57BB8A}40.33 & \cellcolor[HTML]{90D2B2}137.65 & \cellcolor[HTML]{5CBD8D}153.79 & \cellcolor[HTML]{58BC8B}307.12 & \cellcolor[HTML]{6FC59B}179.98 & \cellcolor[HTML]{59BC8C}134.94 \\
 & \textbf{es} & \cellcolor[HTML]{84CEAA}-06.20 & \cellcolor[HTML]{59BC8B}108.73 & \cellcolor[HTML]{FFFFFF}-143.29 & \cellcolor[HTML]{5BBD8D}\textbf{142.12} & \cellcolor[HTML]{5EBE8F}151.53 & \cellcolor[HTML]{9BD7BA}29.67 & \cellcolor[HTML]{57BB8A}\textbf{184.90} & \cellcolor[HTML]{97D5B7}66.78 \\
 & \textbf{hi} & \cellcolor[HTML]{62C092}13.55 & \cellcolor[HTML]{5CBD8D}103.51 & \cellcolor[HTML]{5CBD8D}35.75 & \cellcolor[HTML]{9DD8BB}136.49 & \cellcolor[HTML]{57BB8A}\textbf{156.85} & \cellcolor[HTML]{57BB8A}310.41 & \cellcolor[HTML]{79C9A2}177.81 & \cellcolor[HTML]{5BBD8D}133.48 \\
\bottomrule
\end{tabular}%
}
\caption{Actual perplexity scores for $ZS$ \textit{vs} $\Delta$-perplexity scores for $X$-$FT$ for \texttt{bloom-7B1} over the \textit{toxic-test} evaluation set. 
\textbf{\textit{Takeaway}}: \textit{All the languages were adversely affected}.}
\label{fig:perp-bloom-toxic-test}
\end{table*}

\begin{table*}[t]
\centering \resizebox{0.85\textwidth}{!}{%
\begin{tabular}{cccccccccc}
\toprule
\multicolumn{1}{l}{} & \multicolumn{1}{l}{} & \textbf{am} & \textbf{ar} & \textbf{de} & \textbf{en} & \textbf{es} & \textbf{hi} & \textbf{ru} & \textbf{AVG} \\ \midrule
\multicolumn{1}{l}{} & \textbf{$ZS$} & \cellcolor[HTML]{FFFEFE}25.56 & \cellcolor[HTML]{FAE5E3}49.35 & \cellcolor[HTML]{FDF1F0}37.93 & \cellcolor[HTML]{EA918A}126.36 & \cellcolor[HTML]{E8877E}136.29 & \cellcolor[HTML]{FFFFFF}\textbf{24.47} & \cellcolor[HTML]{E67C73}145.75 & \cellcolor[HTML]{E67C73}77.96 \\\hline\hline
 & \textbf{ar} & \cellcolor[HTML]{60BF91}15.81 & \cellcolor[HTML]{6BC398}35.69 & \cellcolor[HTML]{5EBE8F}28.49 & \cellcolor[HTML]{E1F3EA}116.88 & \cellcolor[HTML]{64C193}128.85 & \cellcolor[HTML]{5BBD8D}13.48 & \cellcolor[HTML]{84CDA9}137.82 & \cellcolor[HTML]{61BF91}68.15 \\
 & \textbf{de} & \cellcolor[HTML]{5EBE8F}17.69 & \cellcolor[HTML]{D9F0E5}-23.17 & \cellcolor[HTML]{84CDA9}-07.59 & \cellcolor[HTML]{94D4B5}120.28 & \cellcolor[HTML]{64C194}128.74 & \cellcolor[HTML]{69C397}-15.38 & \cellcolor[HTML]{DDF1E7}129.21 & \cellcolor[HTML]{7CCAA4}49.97 \\
 & \textbf{en} & \cellcolor[HTML]{5ABD8C}20.08 & \cellcolor[HTML]{5ABC8C}44.56 & \cellcolor[HTML]{58BC8B}33.89 & \cellcolor[HTML]{BBE4D0}118.54 & \cellcolor[HTML]{60BF91}130.26 & \cellcolor[HTML]{59BC8C}17.29 & \cellcolor[HTML]{79C9A2}138.87 & \cellcolor[HTML]{5BBD8D}71.93 \\
 & \textbf{es} & \cellcolor[HTML]{58BC8B}21.65 & \cellcolor[HTML]{61BF91}40.64 & \cellcolor[HTML]{74C79E}07.44 & \cellcolor[HTML]{FFFFFF}115.52 & \cellcolor[HTML]{5CBE8E}131.65 & \cellcolor[HTML]{63C092}-01.93 & \cellcolor[HTML]{FFFFFF}125.84 & \cellcolor[HTML]{69C296}62.97 \\
 & \textbf{hi} & \cellcolor[HTML]{5EBE8F}17.12 & \cellcolor[HTML]{67C295}37.76 & \cellcolor[HTML]{61BF91}25.26 & \cellcolor[HTML]{B8E2CE}118.70 & \cellcolor[HTML]{57BB8A}\textbf{133.48} & \cellcolor[HTML]{58BC8B}20.43 & \cellcolor[HTML]{A4DAC0}134.71 & \cellcolor[HTML]{5FBE8F}69.64 \\
\bottomrule
\end{tabular}%
}
\caption{Actual perplexity scores for $ZS$ \textit{vs} $\Delta$-perplexity scores for $X$-$FT$ for \texttt{bloom-7B1} over the \textit{neutral-test} evaluation set. \textbf{\textit{Takeaway}}: \textit{Detoxification adversely effects the model’s general knowledge}.}
\label{fig:perp-bloom-neutral-test}
\end{table*}

\begin{figure}[t!]
\begin{center}
\includegraphics[width=0.8\columnwidth]{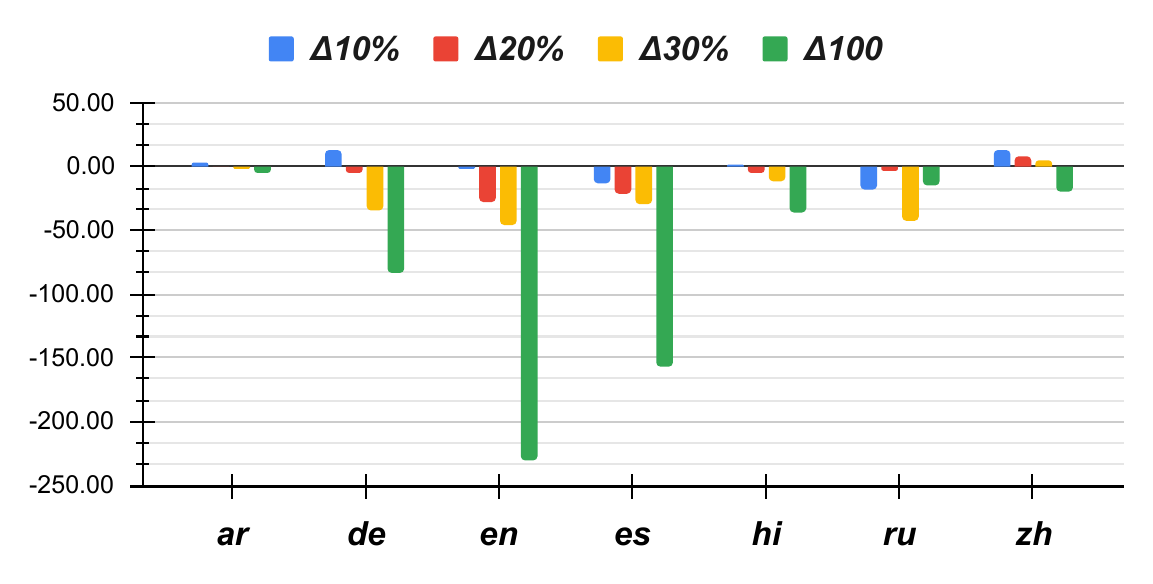}
\end{center}
\caption{Average$\Delta$-Perplexity scores for Percent-based Fine-Tuning ($P$-$FT$) \textit{vs} Multilingual Fine-Tuning ($M$-$FT$) for \texttt{aya-expanse-8B} over the \textit{toxic-train} evaluation set. $10\%$, $20\%$, $30\%$, and $100\%$ represents the Average $\Delta$-Perplexity in $P$-$FT$ and $M$-$FT$ settings. \textbf{\textit{Takeaway}}: \textit{The $100\%$-FT showed adverse effects in ``en'' and ``es''. }}
\label{fig:delta-perp-aya-8-toxic-train}
\end{figure}

\begin{figure}[t!]
\begin{center}
\includegraphics[width=0.8\columnwidth]{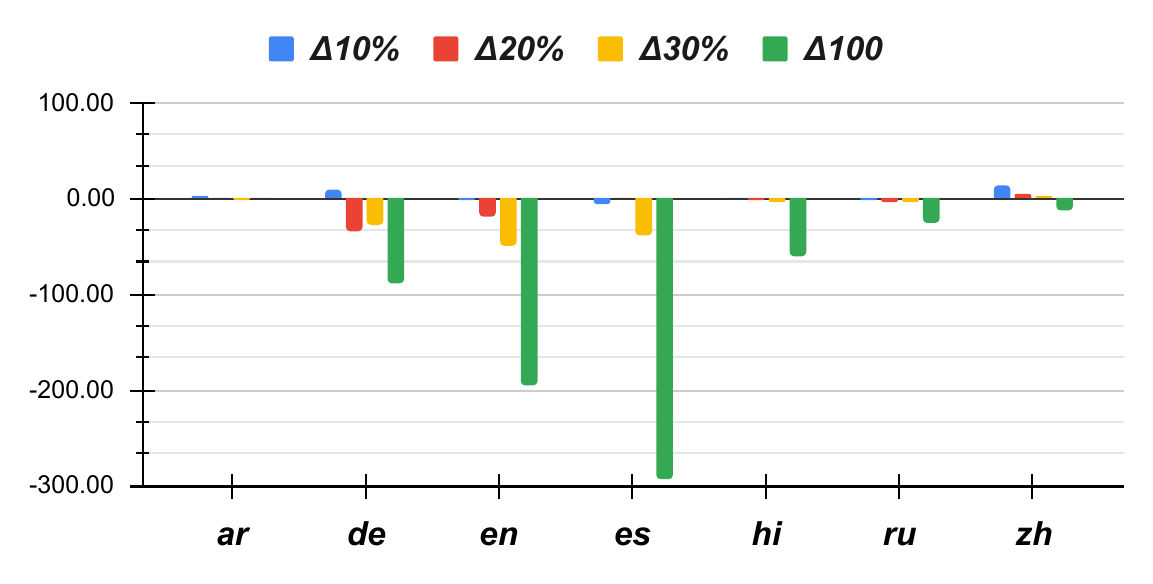}
\end{center}
\caption{Average$\Delta$-Perplexity scores for $P$-$FT$ \textit{vs} $M$-$FT$ for \texttt{aya-expanse-8B} over the \textit{toxic-test} evaluation set. \textbf{\textit{Takeaway}}: \textit{The $100\%$-FT showed adverse effects in ``en'' and ``es''. }}
\label{fig:delta-perp-aya-8-toxic-test}
\end{figure}

\begin{figure}[t!]
\begin{center}
\includegraphics[width=0.8\columnwidth]{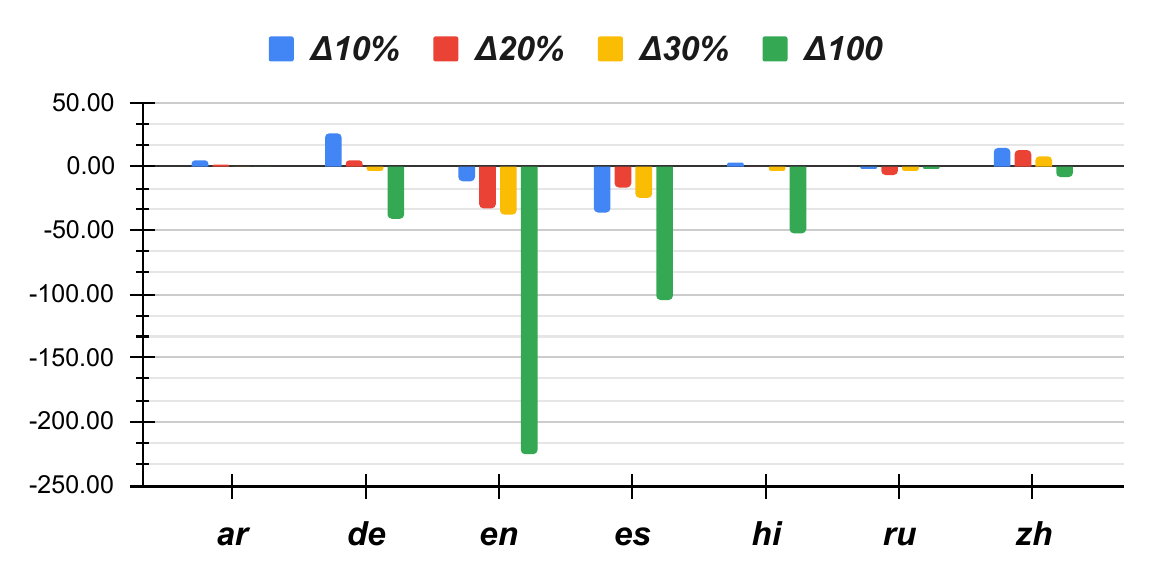}
\end{center}
\caption{Average$\Delta$-Perplexity scores for $P$-$FT$ \textit{vs} $M$-$FT$ for \texttt{aya-expanse-8B} over the \textit{neutral-test} evaluation set. \textbf{\textit{Takeaway}}: \textit{The $100\%$-FT showed adverse effects in ``en'' and ``es''. }}
\label{fig:delta-perp-aya-8-neutral-test}
\end{figure}

\begin{figure}[t!]
\begin{center}
\includegraphics[width=0.8\columnwidth]{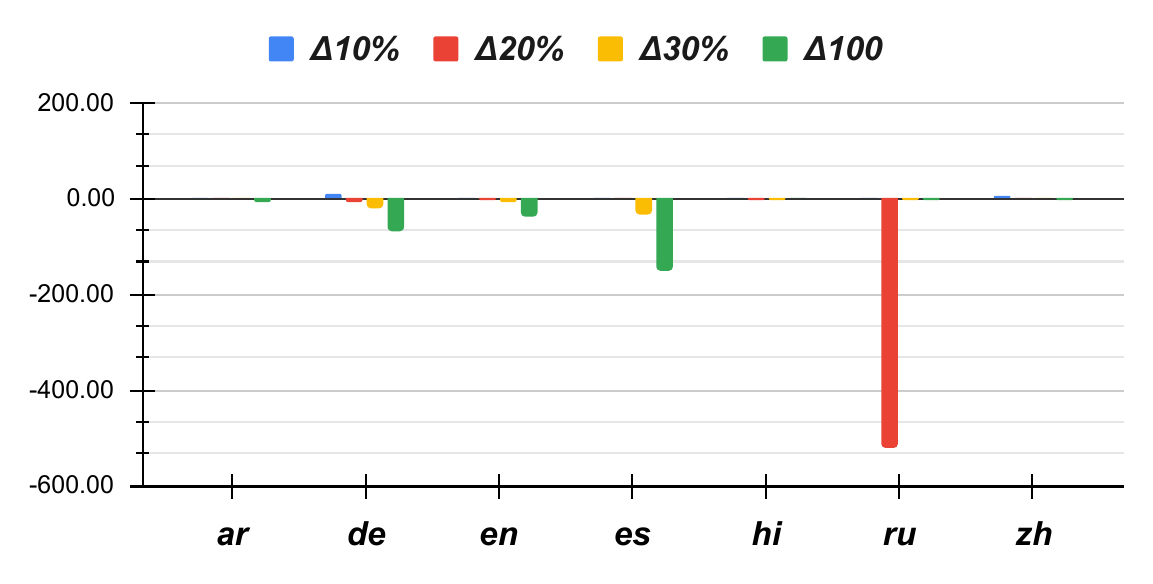}
\end{center}
\caption{Average$\Delta$-Perplexity scores for $P$-$FT$ \textit{vs} $M$-$FT$ for \texttt{aya-23-8B} over the \textit{toxic-train} evaluation set. \textbf{\textit{Takeaway}}: \textit{The $100\%$-FT showed adverse effects in ``en'' and ``es'' and $20\%$ in ``zh''.}}
\label{fig:delta-perp-aya-23-toxic-train}
\end{figure}

\begin{figure}[t!]
\begin{center}
\includegraphics[width=0.8\columnwidth]{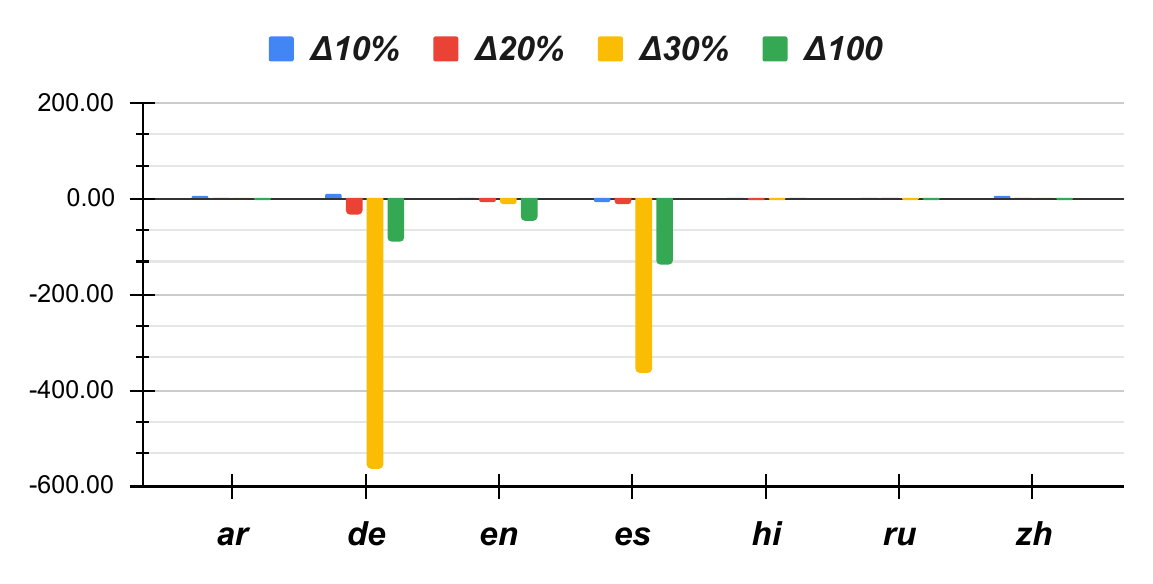}
\end{center}
\caption{Average$\Delta$-Perplexity scores for $P$-$FT$ \textit{vs} $M$-$FT$ for \texttt{aya-23-8B} over the \textit{toxic-test} evaluation set. \textbf{\textit{Takeaway}}: \textit{The $30\%$-FT showed adverse effects in ``de'' and ``es''.}}
\label{fig:delta-perp-aya-23-toxic-test}
\end{figure}
\begin{figure}[t!]
\begin{center}
\includegraphics[width=0.8\columnwidth]{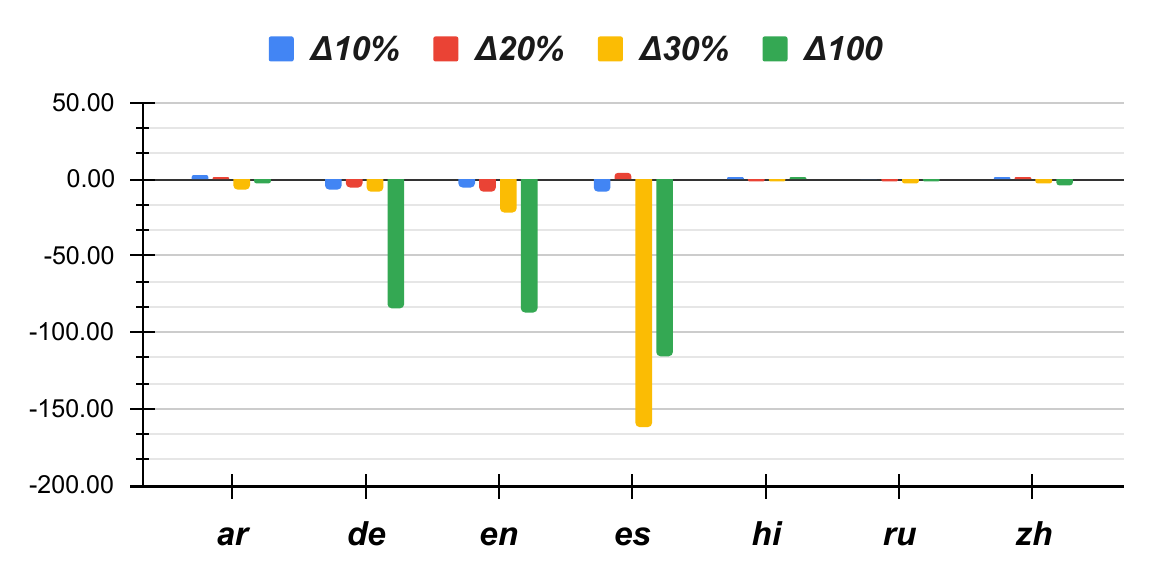}
\end{center}
\caption{Average$\Delta$-Perplexity scores for $P$-$FT$ \textit{vs} $M$-$FT$ for \texttt{aya-23-8B} over the \textit{neutral-test} evaluation set. \textbf{\textit{Takeaway}}: \textit{The $100\%$-FT showed adverse effects in ``en'' and ``es'', and $30\%$ in ``es''.}}
\label{fig:delta-perp-aya-23-neutral-test}
\end{figure}

\begin{figure}[t!]
\begin{center}
\includegraphics[width=0.8\columnwidth]{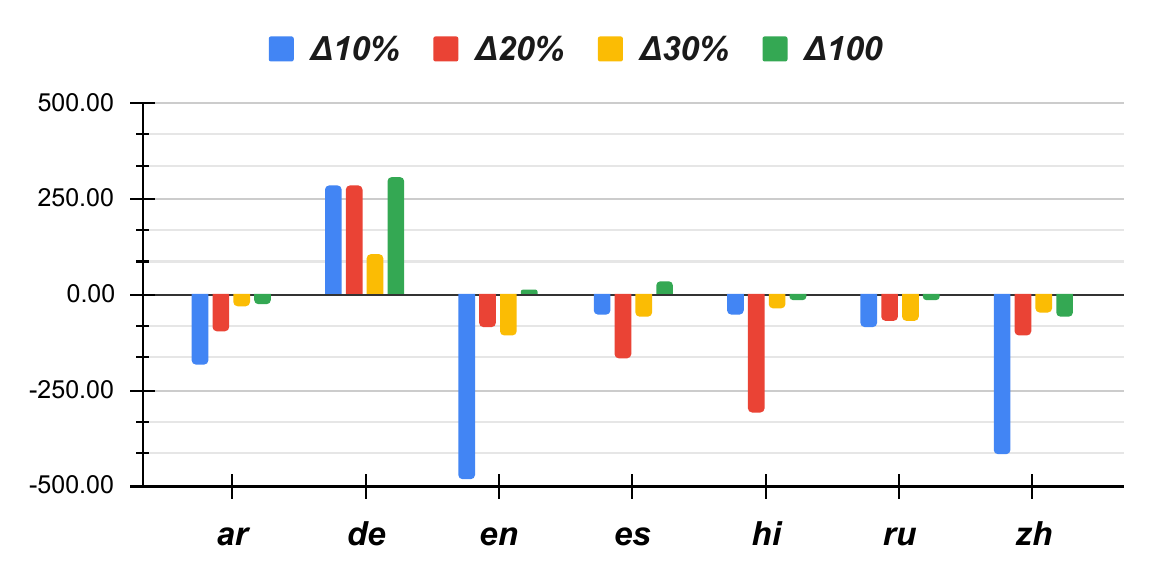}
\end{center}
\caption{Average$\Delta$-Perplexity scores for $P$-$FT$ \textit{vs} $M$-$FT$ for \texttt{mt5-large} over the \textit{toxic-train} evaluation set. \textbf{\textit{Takeaway}}: \textit{All the languages were adversely affected except ``de''.}}
\label{fig:delta-perp-mt5-toxic-train}
\end{figure}

\begin{figure}[t!]
\begin{center}
\includegraphics[width=0.8\columnwidth]{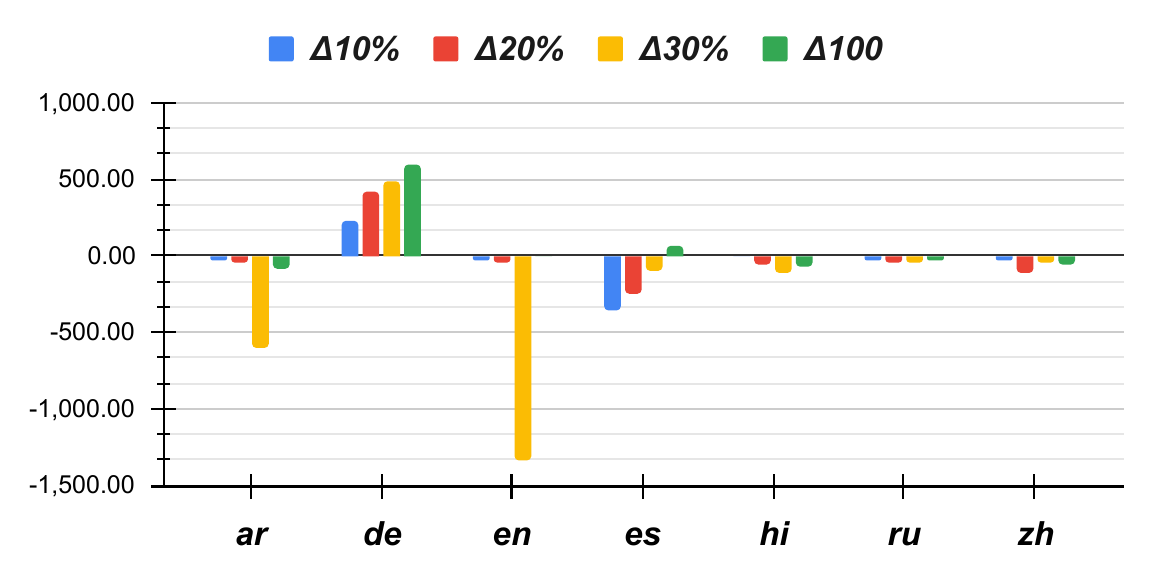}
\end{center}
\caption{Average$\Delta$-Perplexity scores for $P$-$FT$ \textit{vs} $M$-$FT$ for \texttt{mt5-large} over the \textit{toxic-test} evaluation set. \textbf{\textit{Takeaway}}: \textit{The $30\%$-FT showed adverse effects in ``en''.}}
\label{fig:delta-perp-mt5-toxic-test}
\end{figure}
\begin{figure}[t!]
\begin{center}
\includegraphics[width=0.8\columnwidth]{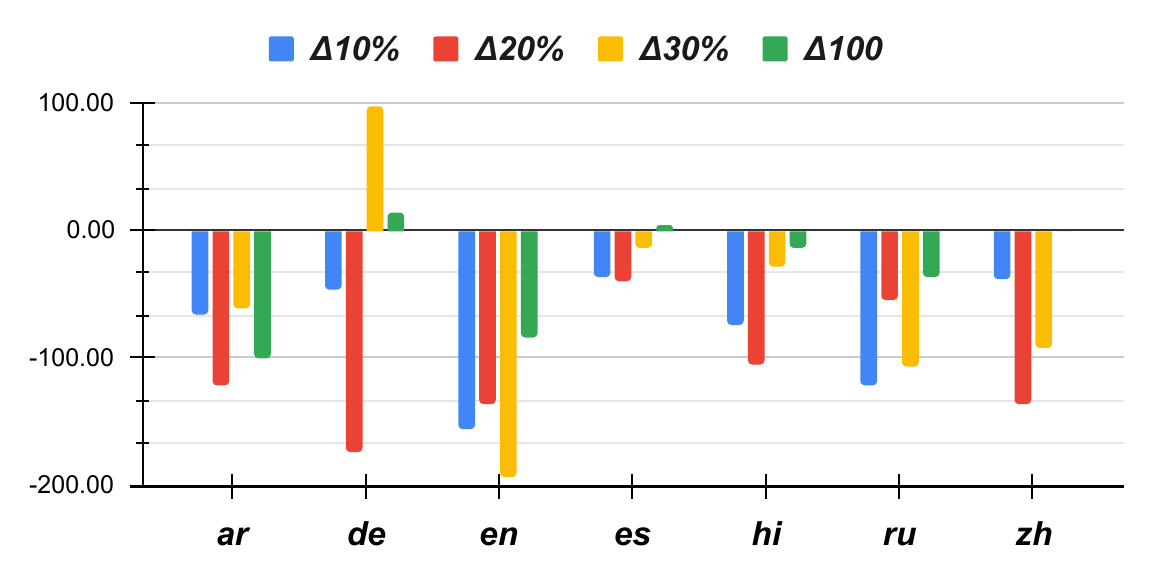}
\end{center}
\caption{Average$\Delta$-Perplexity scores for $P$-$FT$ \textit{vs} $M$-$FT$ for \texttt{mt5-large} over the \textit{neutral-test} evaluation set. \textbf{\textit{Takeaway}}: \textit{All the languages were adversely affected.}}
\label{fig:delta-perp-mt5-neutral-test}
\end{figure}

\begin{figure}[t!]
\begin{center}
\includegraphics[width=0.8\columnwidth]{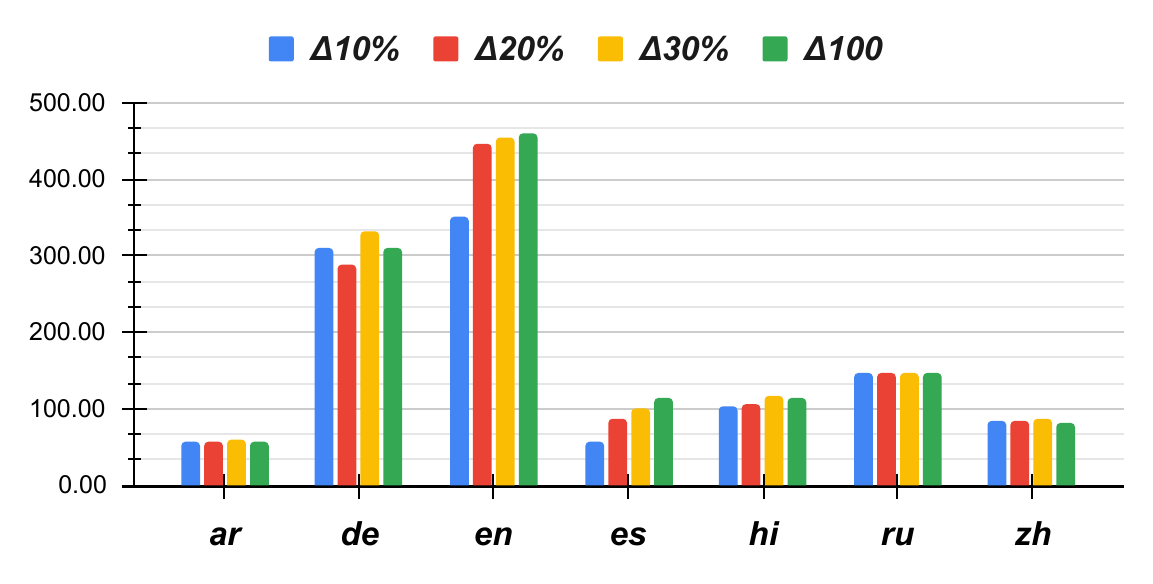}
\end{center}
\caption{Average$\Delta$-Perplexity scores for $P$-$FT$ \textit{vs} $M$-$FT$ for \texttt{bloom-7B1} over the \textit{toxic-train} evaluation set. \textbf{\textit{Takeaway}}: \textit{All the languages were not adversely affected except ``de'' in $10\%$.}}
\label{fig:delta-perp-bloom-toxic-train}
\end{figure}

\begin{figure}[t!]
\begin{center}
\includegraphics[width=0.8\columnwidth]{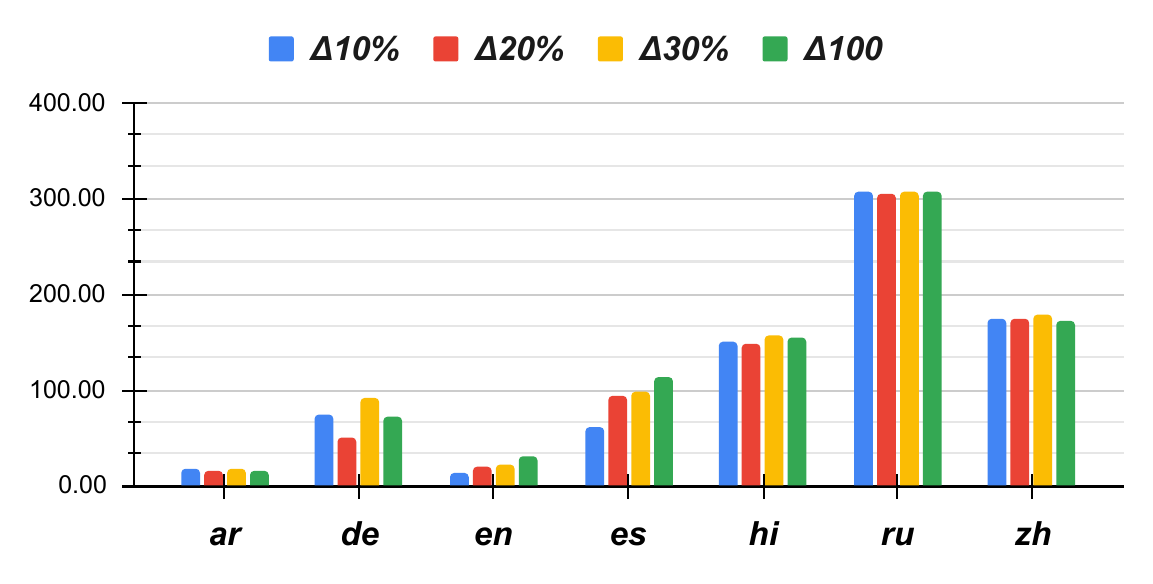}
\end{center}
\caption{Average$\Delta$-Perplexity scores for $P$-$FT$ \textit{vs} $M$-$FT$ for \texttt{bloom-7B1} over the \textit{toxic-test} evaluation set. \textbf{\textit{Takeaway}}: \textit{All the languages showed significant scores.}}
\label{fig:delta-perp-bloom-toxic-test}
\end{figure}

\begin{figure}[t!]
\begin{center}
\includegraphics[width=0.8\columnwidth]{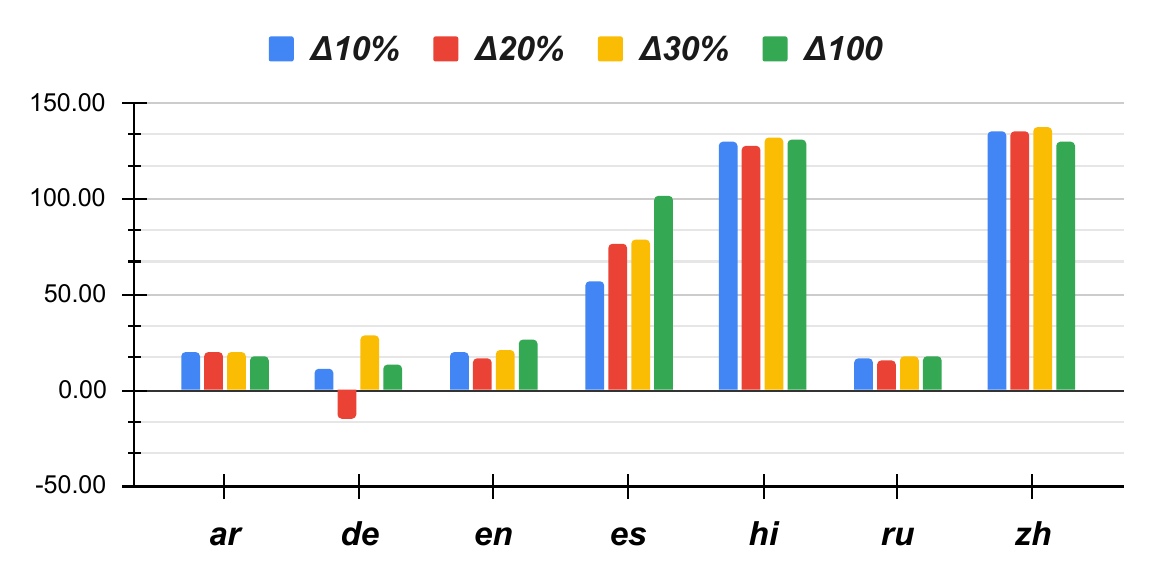}
\end{center}
\caption{Average$\Delta$-Perplexity scores for $P$-$FT$ \textit{vs} $M$-$FT$ for \texttt{bloom-7B1} over the \textit{neutral-test} evaluation set. \textbf{\textit{Takeaway}}: \textit{All the languages showed significant scores.}}
\label{fig:delta-perp-bloom-neutral-test}
\end{figure}

\end{document}